\documentclass[preprint]{vgtc}                 

\PassOptionsToPackage{svgnames}{xcolor}

\usepackage{xcolor} 
\usepackage{colortbl}
\usepackage{tabu}
\usepackage{booktabs}

\definecolor{red}{RGB}{230, 117, 101} 
\definecolor{red_darker}{RGB}{204, 64, 64}
\definecolor{green}{RGB}{152, 194, 130}
\definecolor{green_darker}{RGB}{112, 154, 90}
\definecolor{yellow}{RGB}{251, 210, 106}

\graphicspath{{figures/}{pictures/}{images/}{./}}

\usepackage{times}

\usepackage{tabu}
\usepackage{booktabs}
\usepackage{placeins}
\usepackage{balance}
\usepackage{lipsum}
\usepackage{mwe}

\usepackage{mathptmx}

\usepackage{pifont}
\usepackage{cite}
\usepackage[pagebackref=true,breaklinks=true,colorlinks,bookmarks=false,pagebackref,bookmarks]{hyperref}

\renewcommand{\vec}[1]{\boldsymbol{#1}}

\newcommand{\x}{\vec{x}}

\usepackage{longtable}

\usepackage{amsmath}
\usepackage{mathtools}
\usepackage[mathcal]{eucal}
\usepackage{orcidlink}

\vgtccategory{Research}
\vgtcinsertpkg

\preprinttext{}
\nocopyrightspace
\author{Yuqiao Xia\,\orcidlink{0009-0006-5913-2608}\thanks{e-mail: swuxyq@gmail.com}%
\and Yuchen Deng\,\orcidlink{0009-0006-3893-955X}\thanks{e-mail: swudyc714@email.swu.edu.cn}%
\and Ziheng Chang\,\orcidlink{0009-0001-6040-3165}\thanks{e-mail: chang1213x@gmail.com}%
\and Bingyao Huang\,\orcidlink{0000-0002-8647-5730}\thanks{e-mail: bhuang@swu.edu.cn. Corresponding author.}}
\affiliation{\scriptsize College of Computer and Information Science, Southwest University, Chongqing, China}

\crefname{figure}{Fig.}{Figs.}
\crefname{table}{Table}{Tables}
\crefname{equation}{Eq.}{Eqs.}

\title{SREdit: Scene Response Guided Appearance Editing for Projection Mapping}

\teaser{
    \centering
    \vspace{-0.35em}
    \makebox[\linewidth][c]{%
        \includegraphics[width=1.08\linewidth]{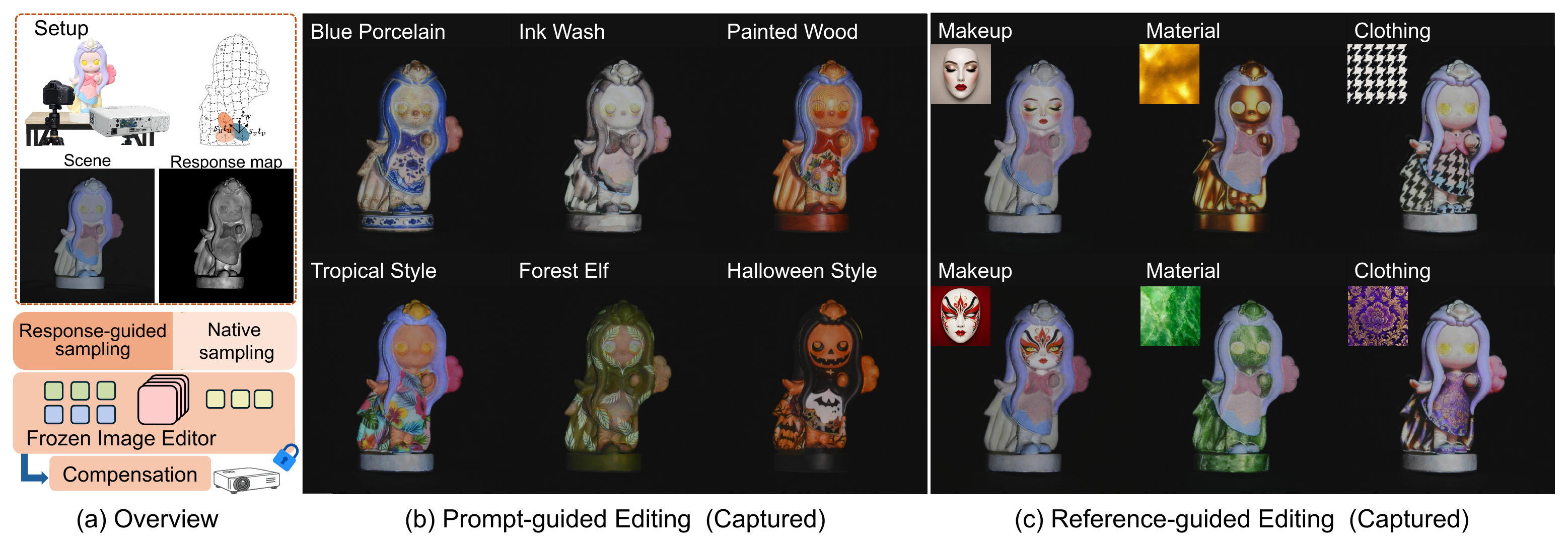}%
    }
    \vspace{-0.25em}
    \caption{\textbf{SREdit for response-guided appearance editing in projection mapping.}
    (a) Given a calibrated projector--camera setup, SREdit estimates a scene response map from the reconstructed scene.
    During early sampling, the source appearance preserves registration with the physical object, while the scene response map adapts the amount of editing across surface regions.
    The image editor then completes the remaining sampling steps.
    Downstream projector compensation converts the edited target into a compensation image for physical projection.
    (b) Camera-captured results of text-guided editing show diverse changes in style and appearance.
    (c) Camera-captured results of reference-guided editing demonstrate localized makeup and clothing edits, as well as material transfer.
    All results in (b) and (c) are captured from physical projections.
    The labels in (b) and (c) are abbreviated prompts; the full prompts are listed in \cref{tab:complete-prompts} in the supplementary material.}
    \label{fig:teaser}
    \vspace{-0.35em}
}

\abstract{
Projection-based appearance editing changes selected visual attributes of a physical object through projection mapping (PM).
To display a desired appearance, PM typically uses projector compensation, which computes a projector input that counteracts the geometric and photometric distortions of the scene.
Compensation can only approximate this appearance within the available projection range, and it cannot revise the appearance itself.
Pretrained LLM-based image editors provide a direct way to produce the desired appearance from a text prompt or reference image; projector compensation can then attempt to reproduce the edited result through projection.
However, these editors do not have an accurate measure of the scene geometry or scene response.
Therefore, their outputs may contain geometric misalignment, inconsistent shading, or changes outside the available projection range, which downstream projector compensation cannot correct.
To address this problem, we introduce SREdit, an efficient response-guided framework for projection-based appearance editing.
SREdit estimates a scene response map from the reconstructed scene that reflects how strongly projected illumination can affect different surface regions.
SREdit uses the source appearance to constrain the early sampling process and preserve registration with the fixed physical object, while the scene response map adapts the amount of editing across surface regions.
Experiments across two pretrained image editors show that SREdit improves projected appearance quality and reproduction fidelity.
It adds less than one second per image with negligible VRAM overhead.
Participants also consistently preferred the results produced with SREdit in our user study.
}


\begin{document}
\firstsection{Introduction}

\maketitle

\begingroup
Projection-based appearance editing changes selected visual attributes of a physical object through projection.
Most projection-mapping research instead assumes that a desired appearance is already available and focuses on reproducing it through projector compensation, which counteracts the geometric and photometric effects of the surface and projector--camera setup~\cite{bimber2008visualcomputing,grundhofer2018review,iwai2024review}.
Compensation can approximate this appearance only within the available projection range and does not create the appearance itself.
A user who wants to edit an existing object must therefore prepare this desired appearance separately.
Pretrained LLM-based image editors provide a direct way to produce the desired appearance from a text prompt or reference image; projector compensation can then attempt to reproduce the edited result through projection.
This pipeline reduces the need to prepare projection content manually, but it also exposes a mismatch between image editing and physical projection.
However, these editors do not have an accurate measure of the scene geometry or scene response.
Here, scene response describes how strongly projected illumination can affect different surface regions.
Without this information, the edited target may contain geometric misalignment, inconsistent shading, or changes outside the available projection range, all of which should be addressed during editing because downstream projector compensation cannot revise the target itself.

Existing approaches can incorporate scene-related constraints through test-time gradient-based guidance or optimization~\cite{Deng2025LAPIG,liu2026conphyg}.
They require gradients during generation, which adds computation and memory on top of the generator for each edit.
Other methods inject geometry through additional control modules~\cite{zhang2023controlnet,mou2024t2iadapter}, but each module must be paired with a compatible model architecture, and many pretrained image editors do not provide such modules.
These restrictions make it difficult to apply the same scene-related control across different editors.

To address this problem, we introduce SREdit, an efficient response-guided framework for projection-based appearance editing (\cref{fig:teaser}).
SREdit estimates a scene response map from the reconstructed scene that reflects how strongly projected illumination can affect different surface regions, and introduces this information into the editing process through response-guided sampling.
During sampling, SREdit uses the source appearance to constrain the early sampling process and preserve registration with the fixed physical object, and adapts the amount of change according to the scene response, preserving lower response regions more strongly while allowing greater changes in higher response regions.
It requires no additional training or control module for the image editor and does not backpropagate through the editor during generation.
Downstream projector compensation remains unchanged.

Experiments across two pretrained image editors show that SREdit improves projected appearance quality and reproduction fidelity while requiring little additional computation and memory.
Participants also consistently preferred the results produced with SREdit in our user study.
Our contributions are:
\begin{itemize}
    \item
    To the best of our knowledge, SREdit is the first framework to incorporate scene response into pretrained image editing for projection mapping, while separating appearance editing from downstream projector compensation.

    \item
    We introduce response-guided sampling, which uses a scene response map estimated from the reconstructed scene to adapt the amount of change across surface regions without backpropagating through the editor during generation.

    \item
    SREdit requires no additional training or control module for the image editor and is evaluated with two pretrained image editors across six physical setups.
    Our experiments demonstrate improved projected appearance quality and reproduction fidelity with little additional computation and memory.
\end{itemize}
\endgroup

\section{Related Work}

\subsection{Projection mapping and appearance editing}

Projection mapping (PM) alters the perceived appearance of physical surfaces by superimposing projected imagery, and prior surveys have reviewed its geometric, photometric, hardware, and application aspects~\cite{bimber2008visualcomputing,grundhofer2018review,iwai2024review}.
Projection has been used to augment pictorial artwork~\cite{Bimber2005artwork}, guide artists using traditional media and tools~\cite{Flagg2006painting}, manipulate perceived softness~\cite{Punpongsanon2015softar}, augment physical avatars with projected shading~\cite{bermano2013augmenting}, and dynamically alter the appearance of human faces~\cite{bermano2017makeuplamps}.
Dynamic PM further extended these effects to deforming non-rigid surfaces~\cite{narita2017dynamic}, markerless planar surfaces~\cite{Kagami2019animatedstickies}, human arms with skin deformation~\cite{peng2020arm}, and articulated hands~\cite{erel2024casperdpm}.
More recent systems address registration and display conditions in PM.
Peng \textit{et al.}~\cite{peng2025perceptually} combine high-speed facial tracking with a lens-shift coaxial projector--camera setup to reduce misalignment in dynamic facial projection.
Okamoto \textit{et al.}~\cite{okamoto2026shadowless} use a dense synthetic-aperture projector array to suppress interaction shadows in tabletop workspaces, and Kusuyama \textit{et al.}~\cite{Kusuyama2024multiaperture} combine a large-aperture projector with a coaxial residual projector to balance shadow suppression and deblurring.
In brightly illuminated environments, Yasui \textit{et al.}~\cite{yasui2024bright} support PM with mixed-light-field projection, and Fujimura \textit{et al.}~\cite{fujimura2026highcontrast} maintain projection contrast with target-excluding light-field illumination.
These systems demonstrate a broad range of appearance changes on physical targets.
However, the projected content or desired effect is usually pre-authored or produced by an application-specific authoring and tracking pipeline.

Once a desired appearance is specified, projector compensation computes a projector input that counteracts geometric and photometric distortions.
Practical and robust approaches simplified projector--camera calibration~\cite{chen2008practical} and improved compensation under nonlinear device responses~\cite{grundhofer2015robust}, steep surface-reflectance variations~\cite{mihara2014artifact}, and cooperative multi-projector configurations~\cite{tsukamoto2015radiometric}.
Perceptually motivated compensation exploits chromatic adaptation and perceptual anchoring~\cite{huang2017radiometriccompensation}, while related work on non-negative image synthesis accounts for the fact that projection can add light but cannot directly remove light from a physical scene~\cite{luo2021staypositive}.
Learning-based methods subsequently formulated photometric~\cite{huang2019compennet} or full~\cite{huang2022CompenNeSt++} projector compensation as end-to-end learning problems, with related work addressing projection deblurring~\cite{Kageyama2020prodebnet} and high-resolution compensation~\cite{wang2023compenhr}.
More recently, Li \textit{et al.}~\cite{li2026sicomp} generalize full compensation to unseen projection surfaces, illumination conditions, and projector--camera poses without fine-tuning or retraining.
Together, these methods formulate the inverse problem of computing a projector input for a prescribed camera-view appearance.

The complementary forward problem is commonly studied in projector--camera systems (ProCams) as relighting or project-and-capture simulation.
Given a projector input and a physical scene, it predicts the image observed by the camera.
Early methods represented this relationship through light transport matrices, with efficient approximations developed for complex light transport~\cite{wang2009kernel}.
Learned ProCams models subsequently combined geometric and photometric mappings for relighting, compensation, and shape reconstruction~\cite{huang2021DeProCams}, while neural reflectance fields enabled view-agnostic projection mapping~\cite{erel2023neuralprojectionmapping}.
Since these learned models are differentiable, they can also serve as simulators for optimizing projector inputs.
For example, Huang and Ling~\cite{huang2022spaa} optimize projector-based adversarial attacks through a learned project-and-capture network, PCNet.
More recently, Li \textit{et al.}~\cite{Li2025DPCS} formulate ProCams simulation through differentiable path tracing, while Deng \textit{et al.}~\cite{Deng2025GS-ProCams} represent surface geometry and material with 2D Gaussians and jointly estimate the ProCams mappings through differentiable physically based rendering.
These models make it possible to evaluate a projection result and optimize projector inputs in a computational space, but the semantic appearance to be displayed must still be supplied externally.

Vision--language models have also been applied to projected scenes.
Cao \textit{et al.}~\cite{Cao2026ProCap} distinguish the physical scene from projected content for projection-aware captioning.
This work addresses semantic understanding rather than projection content creation.
Complementary work has introduced generative models into the creation of projected appearances.
Erel \textit{et al.}~\cite{erel2023neuralprojectionmapping} use a 2D generative model to define consistent appearances across multiple views and optimize the projected texture through a neural reflectance field.
Deng \textit{et al.}~\cite{Deng2025GS-ProCams} use generated or edited images as desired targets for a Gaussian-based ProCams model.
Deng \textit{et al.}~\cite{Deng2025LAPIG} combine language-guided style transfer with learned compensation and project-and-capture simulation; projection-consistency, compensation-saturation, and projection-saturation losses are then used to update the style-transfer model.
However, LAPIG is not adapted to 3D scenes.
Liu \textit{et al.}~\cite{liu2026conphyg} inject gamut, depth, and edge priors into cooperative generation and use bounded numerical optimization for multi-projector compensation in the adversarial setting.
These methods reduce manual preparation of projection content, but mainly formulate the task as target generation, surface adaptation, or projector-input optimization.
Our problem begins with the observed scene and {seeks to edit its appearance under the current physical setup while preserving registration with the physical object.}

\subsection{Image editing and conditional control}

Image editing provides a direct interface for deriving a desired appearance from an image of the current scene.
Early diffusion-based methods support image guides, written instructions, or visual exemplars~\cite{meng2022sdedit,brooks2023instructpix2pix,yang2023paintbyexample}, while recent unified models cover broader editing tasks and interleaved image--text conditions~\cite{yu2025anyedit,xiao2025omnigen,he2026realign}.
The editing variants of Qwen-Image and LongCat-Image further provide the {pretrained image editors} used in our experiments~\cite{wu2025qwenimage,longcat2025longcatimage}.
A central requirement of these methods is to apply the requested change while preserving source content that should remain unchanged.
Existing approaches control cross-attention, infer edit masks, inject source features and self-attention maps, or retrieve source content through mutual attention~\cite{hertz2023prompttoprompt,couairon2023diffedit,tumanyan2023pnp,cao2023masactrl}.
Kulikov \textit{et al.}~\cite{kulikov2025flowedit} construct an ordinary differential equation (ODE) that directly maps between source and target distributions without source inversion or per-image optimization.
Zhang \textit{et al.}~\cite{zhang2025icedit} formulate image editing as in-context generation with parameter-efficient LoRA-MoE tuning.
These methods derive their controls from images, text, masks, or internal model features rather than the calibrated physical setup.
They therefore provide no direct mechanism for adapting the edit to spatial variations in scene response.

Another line of work edits a reconstructed NeRF or 3D Gaussian scene to maintain consistency across rendered views~\cite{haque2023instructnerf2nerf,wang2024gaussianeditor,wu2024gaussctrl,lee2025editsplat}.
These methods update and optimize the reconstruction for each edit, which can be time-consuming.
We instead keep the reconstructed physical scene fixed and use it to control editing of the target appearance in the camera view before projection.

Conditional control can be introduced through trained modules or by modifying the sampling process.
Zhang \textit{et al.}~\cite{zhang2023controlnet} and Mou \textit{et al.}~\cite{mou2024t2iadapter} train additional branches or adapters for spatial conditions such as edges, depth, segmentation, and pose while keeping the original generator frozen.
Yu \textit{et al.}~\cite{yu2023freedom} and Ye \textit{et al.}~\cite{ye2024tfg} instead guide sampling with energy functions or target predictors without training another conditional generator, but require target evaluation and gradients during inference.
Avrahami \textit{et al.}~\cite{avrahami2022blendeddiffusion,avrahami2023blendedlatent} and Lugmayr \textit{et al.}~\cite{lugmayr2022repaint} likewise require no additional training and apply the user's edit only within a specified image region.
Levin and Fried~\cite{levin2023differentialdiffusion} use a user-provided change map to assign different edit strengths to pixels or regions without external-loss optimization.
In these methods, the user specifies which regions should change and by how much, whereas our scene response map describes the relative effect of direct projector illumination at each surface location.

Unlike the physically guided PM pipelines discussed above~\cite{Deng2025LAPIG,liu2026conphyg}, SREdit applies the same response-guided sampling rule to two pretrained image editors without updating their parameters, adding {an additional control module}, or backpropagating a physical loss during editing.

\section{Method}
\subsection{Problem formulation}
Given a fixed physical scene, a calibrated projector--camera setup, a target camera view, and an editing condition \(c_{\mathrm{edit}}\), our goal is to edit the target object's appearance according to \(c_{\mathrm{edit}}\) and reproduce the resulting appearance as closely as possible through physical projection from the target view.

The condition contains a text prompt and may include a reference image when supported by the editor.
We separate the task into two stages.
First, the image editor produces a camera-space edited target from the observed object.
Second, projector compensation computes a compensation image whose captured result approaches that target.
SREdit changes how the edited target is produced and keeps downstream projector compensation separate.

We assume access to a trained and frozen Gaussian-based differentiable scene renderer~\cite{Deng2025GS-ProCams} \(\mathrm R_\phi\), which maps a projector input \(\mathbf I_p\) to its camera-space appearance \(\hat{\mathbf I}_c=\mathrm R_\phi(\mathbf I_p)\).
The parameters \(\phi\) collect the reconstructed scene and calibrated projector--camera setup.
Given a uniform gray projector pattern \(\mathbf I_p^{\mathrm{gray}}\), we define the corresponding camera-space source appearance as
\begin{equation}
    \mathbf I_{\mathrm{src}}
    \coloneqq
    \mathrm R_\phi
    \left(
        \mathbf I_p^{\mathrm{gray}}
    \right).
\end{equation}

\begin{figure*}[t]
    \centering
    \includegraphics[width=\textwidth]{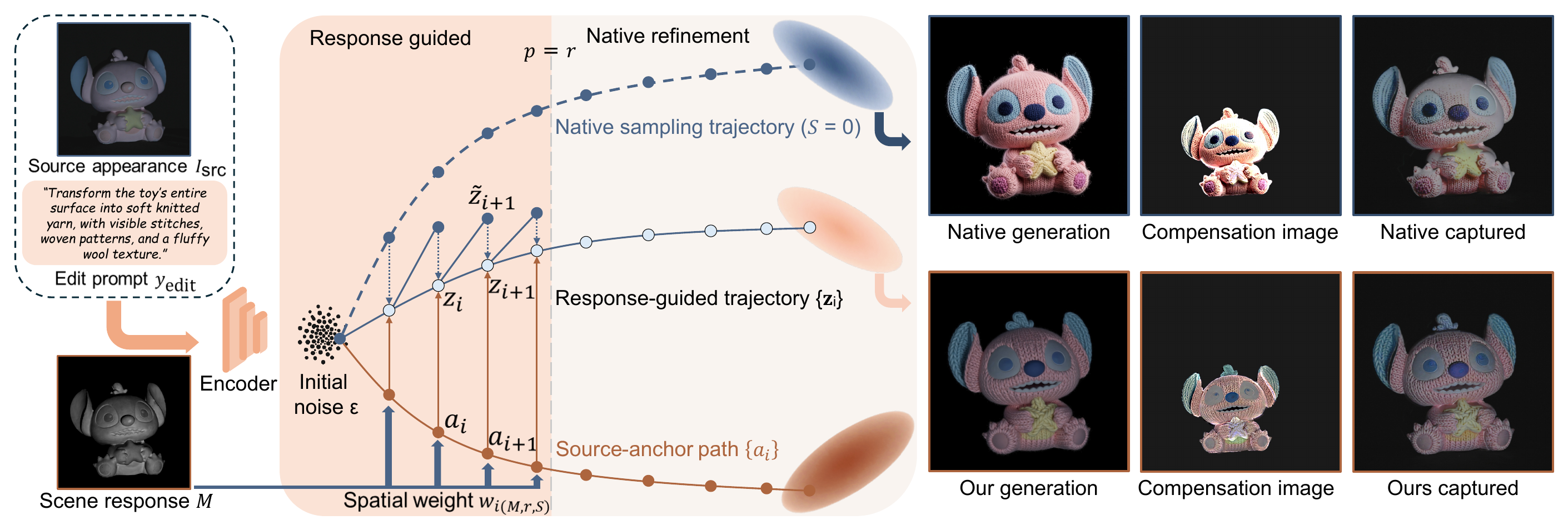}
    \caption{\textbf{Method overview.}
    The pretrained image editor does not have an accurate measure of the scene response of the physical setup.
    During the first \(r\) portion of sampling, SREdit uses the source appearance to preserve registration with the physical object, while the scene response map adapts the amount of editing across surface regions.
    Regions with lower response retain more of the source appearance, whereas regions with higher response allow larger changes.
    The image editor then completes the remaining sampling steps.
    Finally, projector compensation uses iterative refinement to obtain a compensation image for physical projection.}
    \label{fig:method-overview}
\end{figure*}

A pretrained and frozen image editor \(\mathrm E_\theta\) takes \(\mathbf I_{\mathrm{src}}\) and \(c_{\mathrm{edit}}\) as conditions, iteratively samples a latent state from initial noise, and decodes the final state into a camera-space target
\begin{equation}
    \mathbf I_d^{\mathrm{native}}
    =
    \mathrm E_\theta
    \left(
        \mathbf I_{\mathrm{src}},
        c_{\mathrm{edit}}
    \right).
\end{equation}
The pretrained image editor does not have an accurate measure of the scene response of the physical setup.
We therefore derive a scene response map \(M\) from the renderer and use the source appearance as a reference during early sampling, as illustrated in \cref{fig:method-overview}.
The source appearance limits spatial deviations from the observed object, while \(M\) adapts the strength of this constraint across surface regions.
Together, they preserve registration and establish coarse color and shading according to the local scene response, while leaving the remaining steps free to refine the details required by the edit.
With an early control ratio \(r\) and a control strength \(S\), the edited target is written as
\begin{equation}
    \mathbf I_d
    =
    \mathrm E_\theta^{\mathrm{ctrl}}
    \left(
        \mathbf I_{\mathrm{src}},
        c_{\mathrm{edit}};
        M,
        r,
        S
    \right).
\end{equation}
After obtaining the edited target, we perform projector compensation by iteratively refining a compensation image using the frozen renderer:
\begin{equation}
    \hat{\mathbf I}_p^*
    =
    \operatorname*{arg\,min}_{\hat{\mathbf I}_p}
    \mathcal{L}_{\mathrm{comp}}
    \left(
        \mathrm R_\phi(\hat{\mathbf I}_p),
        \mathbf I_d
    \right).
\end{equation}
Iterative refinement optimizes only the compensation image, which is then used as the projector input.
Gradients from \(\mathcal{L}_{\mathrm{comp}}\) are not passed to the image editor.
Thus, the complete task is summarized as
\begin{equation}
    \left(
        \mathbf I_{\mathrm{src}},
        M,
        c_{\mathrm{edit}}
    \right)
    \longrightarrow
    \mathbf I_d
    \longrightarrow
    \hat{\mathbf I}_p^*.
\end{equation}

\subsection{Scene response map}
\label{sec:scene-response-map}
The renderer explicitly models the direct contribution of projector illumination at each reconstructed surface point.
Suppressing the color-channel notation, the direct contribution at \(\mathbf{x}_s\) is
\begin{equation}
    C_p(
        \boldsymbol{\omega}_o,
        \mathbf{x}_s
    )
    =
    f(
        \boldsymbol{\omega}_o,
        \boldsymbol{\omega}_p,
        \mathbf{x}_s
    )
    L_p(
        \boldsymbol{\omega}_p,
        \mathbf{x}_s
    )
    \left(
        \boldsymbol{\omega}_p
        \cdot
        \mathbf{n}_s
    \right),
\end{equation}
where \(\boldsymbol{\omega}_p\) and \(\boldsymbol{\omega}_o\) denote the directions from \(\mathbf{x}_s\) toward the projector and camera, respectively; \(\mathbf{n}_s\) is the surface normal, \(f\) is the surface bidirectional reflectance distribution function (BRDF), and \(L_p\) is the incident projector illumination.
For a camera pixel \(\mathbf u\) observing \(\mathbf{x}_s\) and color channel \(c\), we write the corresponding direct contribution as
\begin{equation}
    C_{p,c}(\mathbf u)
    =
    B_c(\mathbf u)
    L_{p,c}(\mathbf u),
    \qquad
    B_c(\mathbf u)
    =
    f_c(
        \boldsymbol{\omega}_o,
        \boldsymbol{\omega}_p,
        \mathbf{x}_s
    )
    \left(
        \boldsymbol{\omega}_p
        \cdot
        \mathbf{n}_s
    \right).
\end{equation}
We refer to \(B_c(\mathbf u)\) as the channelwise direct-response factor.
It is evaluated from the surface appearance, normal, projector direction, and viewing direction, and describes how strongly a local change in projector illumination affects the direct camera-space contribution.

\begin{figure}[t]
    \centering
    \includegraphics[width=\columnwidth]{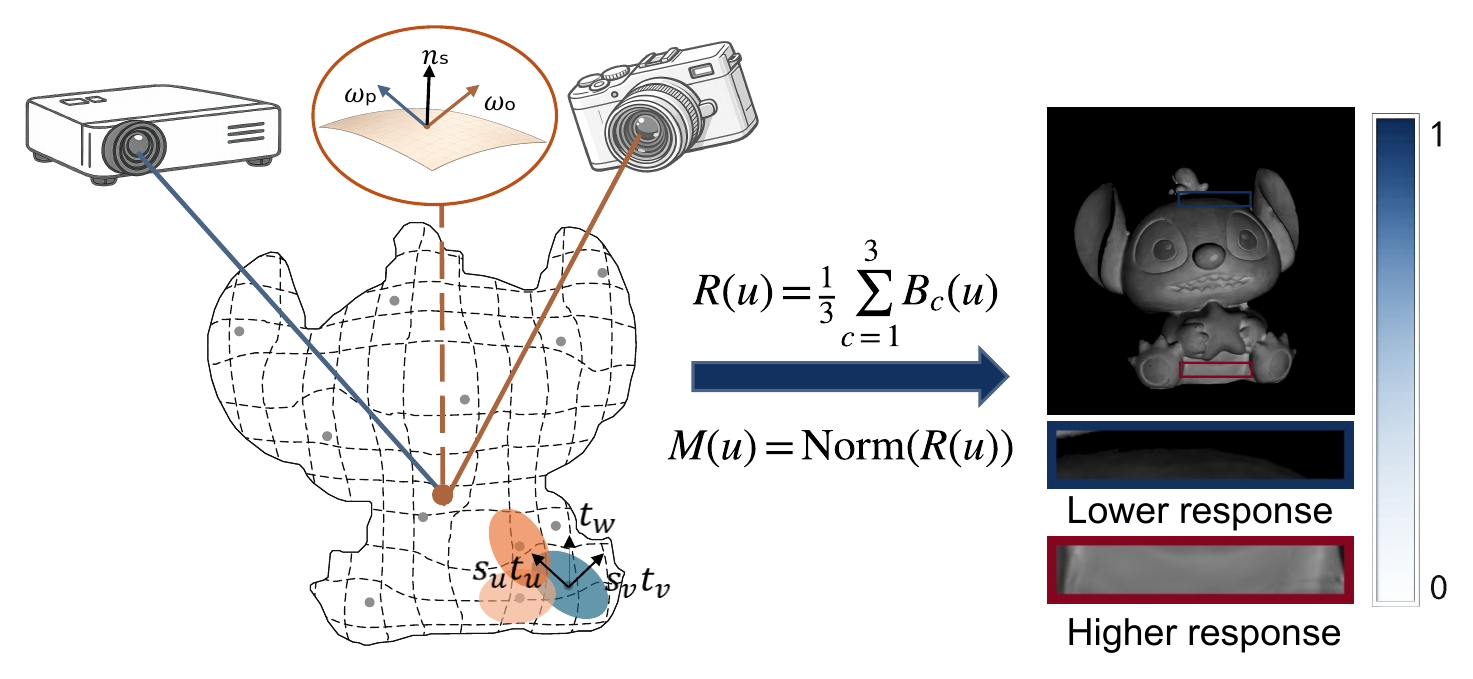}
    \caption{\textbf{Scene response map construction.}
    We evaluate the channelwise direct-response factor \(B_c(\mathbf u)\).
    Channel averaging gives \(R(\mathbf u)\), which is normalized over the reconstructed-object support \(\Omega\) to form \(M\).
    The insets show representative regions with lower and higher response.}
    \label{fig:scene-response-map}
\end{figure}

Response-guided sampling requires a scalar spatial weighting rather than a channel-dependent factor.
We therefore average \(B_c\) over the three color channels to obtain the scalar response \(R\):
\begin{equation}
    R(\mathbf u)
    =
    \frac{1}{3}
    \sum_{c=1}^{3}
    B_c(\mathbf u).
\end{equation}
We then normalize the scalar response within the reconstructed-object support \(\Omega\).
Let \(R_{\Omega}^{\min}\) and \(R_{\Omega}^{\max}\) denote the extrema of \(R\) over this region.
The scene response map is defined as
\begin{equation}
    M(\mathbf u)
    =
    \begin{cases}
        \displaystyle
        \frac{
            R(\mathbf u)-R_{\Omega}^{\min}
        }{
            R_{\Omega}^{\max}-R_{\Omega}^{\min}
        },
        & \mathbf u\in\Omega,\\[10pt]
        0,
        & \mathbf u\notin\Omega.
    \end{cases}
    \label{eq:response-map}
\end{equation}
The construction and spatial interpretation of the scene response map are illustrated in \cref{fig:scene-response-map}.

The map in \cref{eq:response-map} records the normalized spatial distribution of the scalar response within the target view.
Larger values indicate stronger response, whereas smaller values indicate weaker response.
We later use \(1-M\) to distribute the latent correction spatially and use \(S\) to control its overall strength.
The map is applied to the latent update rather than passed to the image editor as an additional image condition; outside \(\Omega\), it is set to zero.

\subsection{Response-guided sampling}
\label{sec:latent-correction}
We formulate response-guided sampling for the deterministic first-order Euler solver used by flow-matching editors~\cite{lipman2023flow}.
Let \(\{\sigma_i\}_{i=0}^{N}\) denote the decreasing noise schedule, where \(\sigma_0=1\) corresponds to the initial noise and \(\sigma_N=0\) to the final data state.
At each step, the frozen image editor predicts a conditional velocity from the current latent, source appearance, and editing condition.
The corresponding Euler update is
\begin{equation}
    \widetilde{\mathbf z}_{i+1}
    =
    \mathbf z_i
    +
    \left(\sigma_{i+1}-\sigma_i\right)
    \mathbf v_\theta
    \left(
        \mathbf z_i,
        \sigma_i;
        \mathbf I_{\mathrm{src}},
        c_{\mathrm{edit}}
    \right),
    \label{eq:native-euler-update}
\end{equation}
where \(\mathbf v_\theta\) is the velocity predicted by the frozen editing model and \(\widetilde{\mathbf z}_{i+1}\) is the latent produced by the Euler step.
SREdit leaves the velocity prediction and Euler update unchanged and applies its correction only afterward.

We reuse the latent representation \(\mathbf z_{\mathrm{src}}\) of the source appearance and the initial noise \(\boldsymbol{\epsilon}\) of the current sampling run.
Directly correcting an intermediate latent toward \(\mathbf z_{\mathrm{src}}\) would mix states at different noise levels.
We therefore construct a source anchor at each noise level by interpolating \(\mathbf z_{\mathrm{src}}\) with the initial noise:
\begin{equation}
    \mathbf a_i
    =
    \left(1-\sigma_i\right)
    \mathbf z_{\mathrm{src}}
    +
    \sigma_i
    \boldsymbol{\epsilon}.
    \label{eq:source-anchor}
\end{equation}
It follows that \(\mathbf a_0=\boldsymbol{\epsilon}\) and \(\mathbf a_N=\mathbf z_{\mathrm{src}}\).
The sequence \(\{\mathbf a_i\}_{i=0}^{N}\) therefore provides a source anchor at the noise level corresponding to each sampling state, without introducing an additional inversion trajectory.
During early sampling, the source anchor limits spatial deviations from the source appearance and helps preserve registration with the fixed physical object.

Applying the correction throughout the complete trajectory would repeatedly pull the latent toward the source anchor.
At high control strength, this can drive the final result back toward the original source appearance and suppress the requested edit.
We therefore restrict the correction to the early sampling stage and leave the remaining steps to the pretrained image editor.
To keep its total temporal weight consistent under different solver discretizations, we parameterize the control schedule by normalized denoising progress.
For the \(i\)-th scheduler state, we define
\begin{equation}
    p_i
    =
    \frac{
        \displaystyle
        \sum_{j=0}^{i-1}
        \left(
            \sigma_j-\sigma_{j+1}
        \right)
    }{
        \displaystyle
        \sum_{j=0}^{N-1}
        \left(
            \sigma_j-\sigma_{j+1}
        \right)
    }.
\end{equation}
Under the endpoint convention above, this reduces to \(p_i=1-\sigma_i\).
Given an early control ratio \(r\in(0,1]\), we use the normalized half-cosine kernel
\begin{equation}
    k_r(p)
    =
    \begin{cases}
        \displaystyle
        \frac{1}{r}
        \left[
            1+
            \cos
            \left(
                \frac{\pi p}{r}
            \right)
        \right],
        & 0\leq p\leq r,\\[10pt]
        0,
        & r<p\leq 1,
    \end{cases}
\end{equation}
and assign the \(i\)-th solver interval the integrated temporal weight
\begin{equation}
    m_i
    =
    \int_{p_i}^{p_{i+1}}
    k_r(p)\,\mathrm{d}p.
\end{equation}
Because the kernel integrates to one, \(\sum_{i=0}^{N-1}m_i=1\), irrespective of how the same sigma range is discretized.
The schedule concentrates the correction toward the source anchor in the early denoising stage, where registration, coarse color, and shading are stabilized according to the local scene response, and turns it off afterward to leave later steps more freedom to refine details required by the edit.

\begin{figure}[t]
    \centering
    \includegraphics[width=\columnwidth]{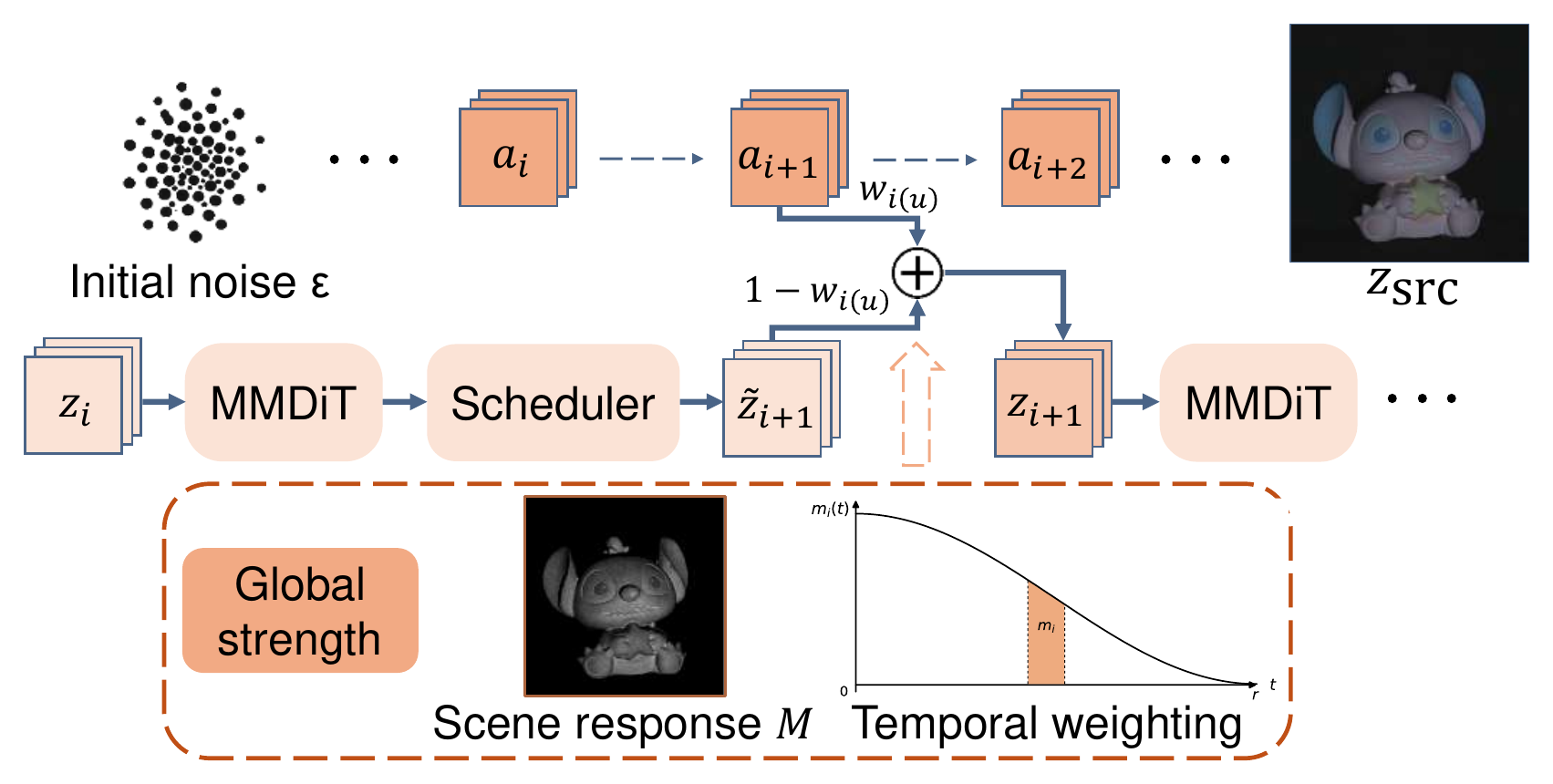}
    \caption{\textbf{Response-guided sampling.}
    At one sampling step, the Euler update produces \(\widetilde{\mathbf z}_{i+1}\), while \(\mathbf a_{i+1}\) provides the source anchor at the same noise level.
    The {scene }response map, control strength, and temporal weight determine \(w_i(\mathbf u)\), which moves the updated latent toward the source anchor.}
    \label{fig:post-step-correction}
\end{figure}

After the Euler update in \cref{eq:native-euler-update}, we resize \(M\) to the latent spatial resolution and combine it with the temporal weight \(m_i\) and the control strength \(S\) to obtain
\begin{equation}
    w_i(\mathbf u)
    =
    1-
    \exp
    \left[
        -S
        \left(
            1-M(\mathbf u)
        \right)
        m_i
    \right].
    \label{eq:response-weight}
\end{equation}
The post-step latent is then corrected toward the source anchor at the same next scheduler state:
\begin{equation}
    \mathbf z_{i+1}(\mathbf u)
    =
    \widetilde{\mathbf z}_{i+1}(\mathbf u)
    +
    w_i(\mathbf u)
    \left[
        \mathbf a_{i+1}(\mathbf u)
        -
        \widetilde{\mathbf z}_{i+1}(\mathbf u)
    \right].
    \label{eq:latent-correction}
\end{equation}
The complete update of one controlled solver step is illustrated in \cref{fig:post-step-correction}.

Using \(\mathbf a_{i+1}\) keeps the source anchor synchronized with the state reached by the Euler step.
The exponential form bounds \(w_i(\mathbf u)\) in \([0,1)\): regions with lower response receive stronger correction toward the source anchor, whereas regions with higher response retain more freedom to follow the editing request.
Once the progress exceeds \(r\), \(m_i=0\), and the pretrained image editor completes the remaining sampling steps.
The source appearance and editing condition remain unchanged throughout sampling.

\begin{figure*}[t]
    \centering
    \includegraphics[width=\textwidth]{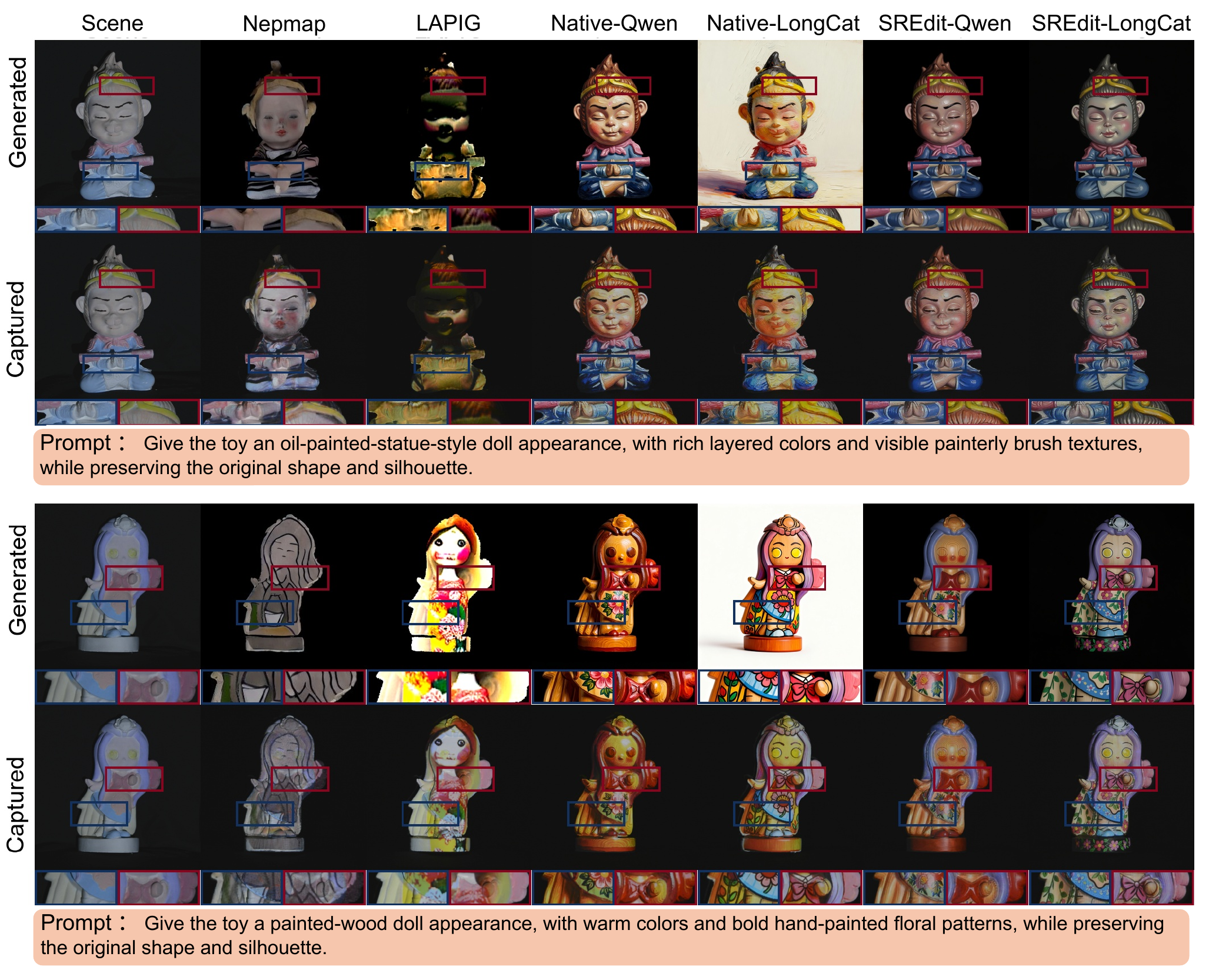}
    \caption{\textbf{Real-projection comparisons.}
    Results are shown for an oil-painted-statue edit (top) and a painted-wood edit (bottom).
    The upper and lower rows show generated targets and camera-captured results, respectively.
    Red and blue boxes indicate enlarged regions for comparing local texture and color reproduction.
    The prompts for these representative comparisons and additional qualitative results are provided in the supplementary material.}
    \label{fig:main-comparison}
\end{figure*}

\section{Experiments}
We organize the evaluation around two comparisons.
First, we compare the complete SREdit pipeline with the language-guided projection-mapping methods Nepmap~\cite{erel2023neuralprojectionmapping} and LAPIG~\cite{Deng2025LAPIG}.
Second, we evaluate Qwen-Image-Edit-2511~\cite{wu2025qwenimage} and LongCat-Image-Edit~\cite{longcat2025longcatimage}, hereafter Qwen and LongCat.
We refer to the original Qwen and LongCat sampling pipelines followed by the same downstream projector compensation as Native-Qwen and Native-LongCat.
SREdit-Qwen and SREdit-LongCat differ from their corresponding Native versions only by response-guided sampling.
Every projector input is physically projected and captured by the camera.
We measure target reproduction and agreement with the editing request, conduct a user study, and ablate the scene response map, early control ratio \(r\), and control strength \(S\).

\subsection{Experimental settings}

\subsubsection{Physical setup}
Our projector--camera system consists of an O.B.E C3 Air projector and a Nikon D3200 camera.
For each scene, the projector and camera poses, together with the camera settings, remain fixed throughout data acquisition.
All experiments are conducted using an NVIDIA A100 GPU with 64 GB of memory.

\subsubsection{Implementation details}
We implement Nepmap and LAPIG following their published pipelines.
Nepmap optimizes the projector input through its learned neural reflectance field, whereas LAPIG performs test-time optimization using InstructPix2Pix~\cite{brooks2023instructpix2pix}, CompenNeSt~\cite{huang2022CompenNeSt++}, and PCNet~\cite{huang2022spaa}.
For our object-centered scenes, we restrict the LAPIG losses to the segmented object region and leave all other parameters unchanged.
The public ConPhyG release~\cite{liu2026conphyg} provides the gamut-network component but not the complete generation and multi-projector pipeline.
We therefore exclude it from the quantitative comparison because the full method cannot be faithfully reproduced.

We use the Gaussian-based differentiable scene renderer~\cite{Deng2025GS-ProCams} and iterative refinement for projector compensation.
The renderer remains frozen during appearance editing, and iterative refinement optimizes only the compensation image used as the projector input.

Qwen and LongCat use 40 and 50 sampling steps, respectively, and receive the same editing request.
Within each editor, Native and SREdit use the same source appearance and iterative refinement.
Because Nepmap and LAPIG use different conditioning interfaces, we adapt their prompts while preserving the same target concept.
The prompts used for the representative qualitative results are provided in the supplementary material.

We fix \(r=0.2, S=2.5\) for Qwen and \(r=0.2, S=3.0\) for LongCat according to the result quality on a separate setup that is not used for evaluation.
These settings remain unchanged in the main comparison and scene response map ablation.

\subsubsection{Evaluation metrics}
We evaluate reproduction fidelity using peak signal-to-noise ratio (PSNR), structural similarity index measure (SSIM), learned perceptual image patch similarity (LPIPS)~\cite{zhang2018lpips}, and CIEDE2000 color difference (\(\Delta E\))~\cite{luo2001ciede2000} between the intended appearance of each method and its camera-captured result within the object region.
Since the intended appearances produced by different methods may vary in complexity, these image-based metrics measure how faithfully each method reproduces its own target rather than directly measuring editing quality.
Contrastive Language-Image Pre-training (CLIP)~\cite{radford2021clip} and the image-text matching score of Bootstrapping Language-Image Pre-training (BLIP-ITM)~\cite{li2022blip} measure agreement between the captured result and the requested appearance; all methods are evaluated using the same concise target description.
We also report wall-clock generation time and peak video random-access memory (VRAM).

\subsection{Comparison with baselines}

\begin{table*}[t]
    \centering
    \caption{\textbf{Performance and efficiency under real projection.}
    PSNR, SSIM, LPIPS, and
    \(\Delta E\) are computed between the intended
    appearance of each method and its camera-captured result. CLIP and
    BLIP-ITM measure agreement with the requested appearance. Results are
    averaged over 18 prompts from six setups. Time is the wall-clock time
    required to generate one image, and VRAM is the peak allocated GPU memory
    during image generation. We use \textbf{bold} and \underline{underline} to
    highlight the best and second-best results, respectively.}
    \label{tab:main-results}
    \renewcommand{\arraystretch}{1.10}
    \scriptsize
    \begin{tabular*}{\textwidth}{
        @{\extracolsep{\fill}}
        lcccccc@{\hspace{10pt}}cc
        @{}
    }
        \toprule
        \textbf{Method} & PSNR \(\uparrow\) & SSIM \(\uparrow\) &
        LPIPS \(\downarrow\) & \(\Delta E\) \(\downarrow\) &
        CLIP \(\uparrow\) & BLIP-ITM \(\uparrow\) &
        Time (s/image) \(\downarrow\) & VRAM (GiB) \(\downarrow\) \\
        \midrule
        Nepmap~\cite{erel2023neuralprojectionmapping} & 19.1512 & \textbf{0.5962} &
        0.3885 & 9.8198 & 29.79 & 59.72 & \textbf{8.70} & \underline{7.68} \\
        LAPIG~\cite{Deng2025LAPIG}\textsuperscript{\(\dagger\)} & 13.4033 & 0.3163 & 0.4189 & 17.4810 &
        28.41 & 48.06 & \underline{9.08} & \textbf{5.60} \\
        Native-LongCat~\cite{longcat2025longcatimage} & 14.7231 & 0.3721 &
        0.3369 & 14.5425 & 30.56 & 91.66 &
        58.50 & 29.73 \\
        SREdit-LongCat & \underline{20.6623} & 0.5432 & \textbf{0.2316} &
        \underline{8.2552} & \textbf{32.44} & \underline{92.62} & 59.10 &
        29.73 \\
        Native-Qwen~\cite{wu2025qwenimage} & 16.5086 & 0.3608 & 0.3571 &
        13.7814 & 31.61 & 88.61 &
        88.37 & 57.94 \\
        SREdit-Qwen & \textbf{21.3113} & \underline{0.5690} & \underline{0.2559} &
        \textbf{8.1197} & \underline{32.10} & \textbf{94.77} &
        89.10 & 57.95 \\
        \bottomrule
    \end{tabular*}
    \vspace{1pt}

    \parbox{\textwidth}{\scriptsize \(\dagger\) 2D surface stylization method.}
\end{table*}

\setcounter{table}{3}

\begin{table*}[!t]
    \centering
    \caption{{\textbf{Scene response map ablation.} For each image
    editor, the results are averaged over the same 18 cases from six setups.
    We use \(r=0.2, S=2.5\) for Qwen and \(r=0.2, S=3.0\) for LongCat.
    We use \textbf{bold} and \underline{underline} to highlight the best and
    second-best results, respectively.}}
    \label{tab:response-map-ablation}
    \renewcommand{\arraystretch}{1.12}
    \scriptsize
    \begin{tabular*}{\textwidth}{
        @{\extracolsep{\fill}}
        lcccccc
        @{}
    }
        \toprule
        \textbf{Method} & PSNR \(\uparrow\) & SSIM \(\uparrow\) &
        LPIPS \(\downarrow\) &
        \(\Delta E\) \(\downarrow\) & CLIP \(\uparrow\) &
        BLIP-ITM \(\uparrow\) \\
        \midrule
        SREdit-Qwen & \underline{21.3113} & \underline{0.5690} & 0.2559 & 8.1197 &
        \underline{32.10} & \textbf{94.77} \\
        SREdit-LongCat & 20.6623 & 0.5432 & \underline{0.2316} & 8.2552 &
        \textbf{32.44} & 92.62 \\
        Uniform-Qwen & \textbf{21.9647} &
        \textbf{0.6129} & 0.2398 &
        \textbf{7.4097} & 31.07 & \underline{93.63} \\
        Uniform-LongCat & 20.9145 &
        0.5487 & \textbf{0.2286} &
        \underline{7.9873} & 31.73 & 90.16 \\
        \bottomrule
    \end{tabular*}
\end{table*}

\setcounter{table}{1}

\cref{tab:main-results} reports the numerical comparison, while \cref{fig:main-comparison} shows representative intended appearances and their camera-captured results.
Nepmap and LAPIG are prior projection-mapping methods, while each Native/SREdit pair uses the same image editor and downstream projector compensation.
The image-based metrics characterize the reproduction of each method's own target and are therefore interpreted together with the qualitative results.
CLIP and BLIP-ITM provide complementary measures of agreement with the requested appearance.

Among the prior projection-mapping baselines, Nepmap achieves a relatively high SSIM of 0.5962 and PSNR of 19.1512\,dB.
As shown in \cref{fig:main-comparison}, however, its generated appearances mainly contain smooth and low-frequency variations, which are comparatively easy to reproduce under projection.
The resulting SSIM therefore does not necessarily indicate a more convincing edit: fine-grained material details are limited, and the captured appearance remains visually less consistent with the requested effect.
LAPIG, which was originally developed for 2D projection surfaces, exhibits noticeable geometric misalignment when applied to a three-dimensional object, including misaligned contours and displaced appearance elements.

The Native variants produce recognizable edits, but their intended appearances often contain bright highlights, saturated colors, and fine changes that are attenuated or clipped after projection.
Native-LongCat also introduces clear geometric misalignment in several cases.
Because projector compensation cannot revise an already misaligned target, this error persists in the camera-captured result.
In contrast, the corresponding SREdit-LongCat results remain registered with the physical object and do not exhibit the same mismatch in the cases shown.

Compared with the corresponding Native variant, SREdit improves all six quality metrics for both image editors.
Relative to Native-LongCat, it increases PSNR by 5.9392\,dB, reduces LPIPS by 0.1053, and raises CLIP by 1.88.
For Qwen, it increases PSNR by 4.8027\,dB, reduces \(\Delta E\) by 5.6617, and raises BLIP-ITM by 6.16.
The source anchor helps preserve registration with the fixed physical object during early sampling, while the scene response map retains more of the observed surface color and shading in regions with lower response.
The captured results still preserve local texture required by the editing request.
This observation agrees with the improvements in both reproduction and semantic metrics.

The absolute runtime and memory usage of Qwen and LongCat are higher than those of Nepmap and LAPIG because the image editors are substantially larger.
Within each comparison using the same image editor, SREdit changes the measured runtime by less than one second and adds negligible VRAM overhead.

\subsubsection{User study}
The study was approved by the institutional ethics review board prior to data collection.
We conducted a pairwise user study with 20 participants (10 female and 10 male, aged 18--24), nine of whom reported a background in image processing, computer graphics, or computer vision.
Each participant evaluated 36 pairs of camera-captured results, comprising 18 Qwen and 18 LongCat comparisons, and compared the SREdit and Native versions of the same image editor.
Each pair was shown with the editing request, and the left--right order varied across cases.
Participants used a four-level forced-choice scale to judge text alignment, visible artifacts, lighting and surface realism, and overall preference.
They also assessed each result separately for visible misalignment.
\cref{tab:user-study} reports the preference and rates of visible misalignment.
The complete questionnaire and additional study details are provided in the supplementary material.

\begin{table}[t]
    \centering
    \caption{\textbf{User study.} Pairwise perceptual evaluation with 20 participants over 36
    cases. Values for text alignment, artifacts, realism, and overall
    preference are the percentages of judgments favoring SREdit over
    the corresponding Native version.
    For visible misalignment, entries report the percentages
    judged free of misalignment for SREdit/Native, respectively.}
    \label{tab:user-study}
    \renewcommand{\arraystretch}{1.10}
    \footnotesize
    \setlength{\tabcolsep}{0.9pt}
    \begin{tabular*}{\columnwidth}{
        @{\extracolsep{\fill}}
        lccc
        @{}
    }
        \toprule
        \textbf{Metric} & All cases & Qwen & LongCat \\
        \midrule
        Text alignment & 65.4\% & 61.4\% & 69.4\% \\
        Fewer artifacts & 72.8\% & 63.9\% & 81.7\% \\
        Realism & 75.0\% & 68.1\% & 81.9\% \\
        No misalignment & 96.5\%/80.4\% &
        96.4\%/90.3\% & 96.7\%/70.6\% \\
        \textbf{Overall preference} & \textbf{72.4\%} &
        \textbf{65.3\%} & \textbf{79.4\%} \\
        \bottomrule
    \end{tabular*}
\end{table}

\begin{table}[t]
    \centering
    \caption{\textbf{Reference-guided editing.} Quantitative comparison over 12 cases
    from four physical setups. Metrics are computed between each intended
    appearance and its camera-captured result within the foreground object mask
    and averaged over all cases. The better result for each metric is shown in
    \textbf{bold}.}
    \label{tab:reference-editing}
    \renewcommand{\arraystretch}{1.10}
    \footnotesize
    \begin{tabular*}{\columnwidth}{
        @{\extracolsep{\fill}}
        lcccc
        @{}
    }
        \toprule
        \textbf{Method} & PSNR \(\uparrow\) & SSIM \(\uparrow\) &
        LPIPS \(\downarrow\) & \(\Delta E\) \(\downarrow\) \\
        \midrule
        Native-Qwen & 18.5257 & 0.6675 & 0.1159 & 10.6841 \\
        SREdit-Qwen & \textbf{21.7335} & \textbf{0.7358} &
        \textbf{0.0965} & \textbf{7.6941} \\
        \bottomrule
    \end{tabular*}
\end{table}

\begin{figure*}[t]
    \centering
    \includegraphics[width=\textwidth]{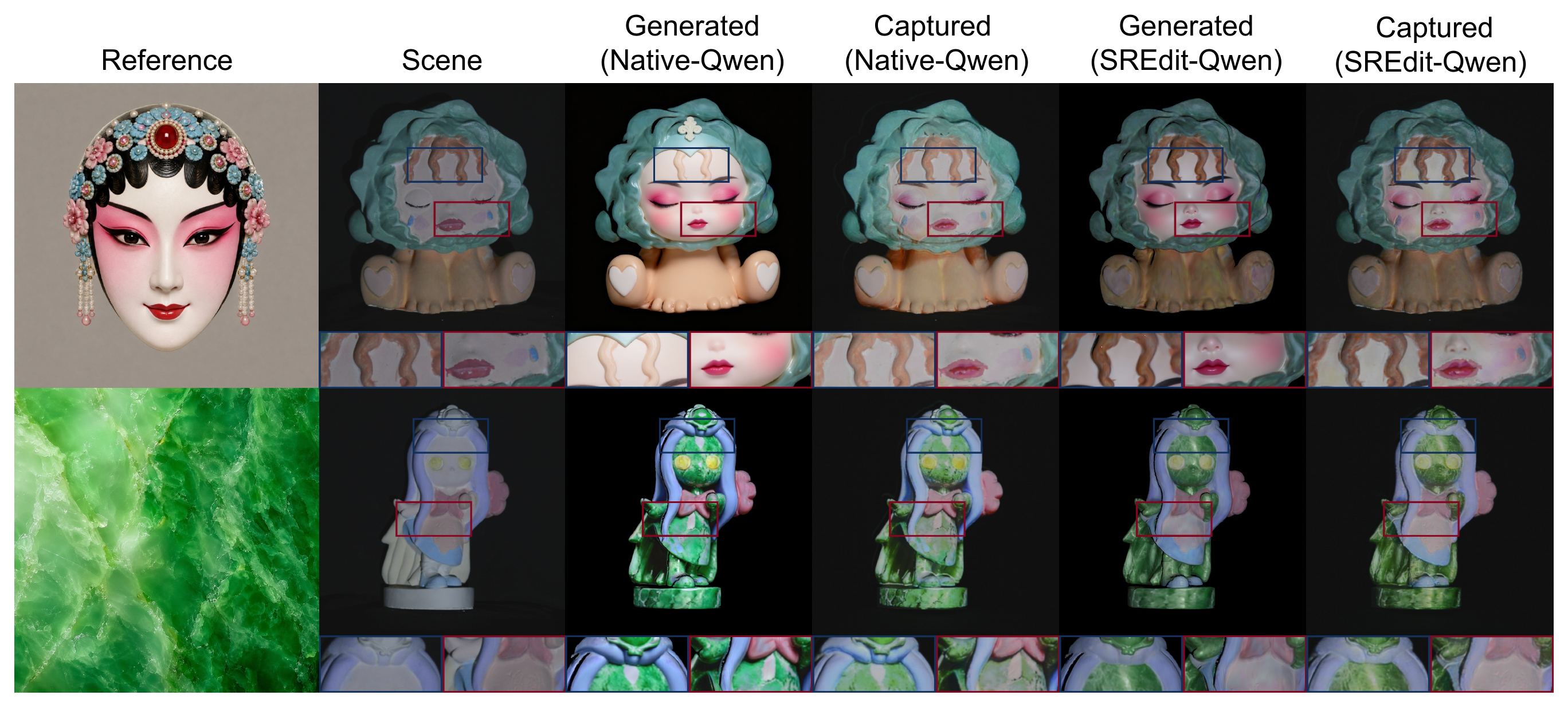}
    \caption{\textbf{Reference-guided qualitative results.}
    Appearance editing using Qwen.
    The first two columns show the reference and source scene.
    The remaining columns show the intended and camera-captured appearances produced by Native-Qwen and SREdit-Qwen.
    The top row uses Chinese opera face painting as the reference for makeup transfer, whereas the bottom row uses emerald as the reference material for material replacement.
    Marked regions are enlarged for local comparison.}
    \label{fig:reference-material-editing}
\end{figure*}

\begin{figure*}[!t]
    \centering
    \includegraphics[width=0.98\textwidth]{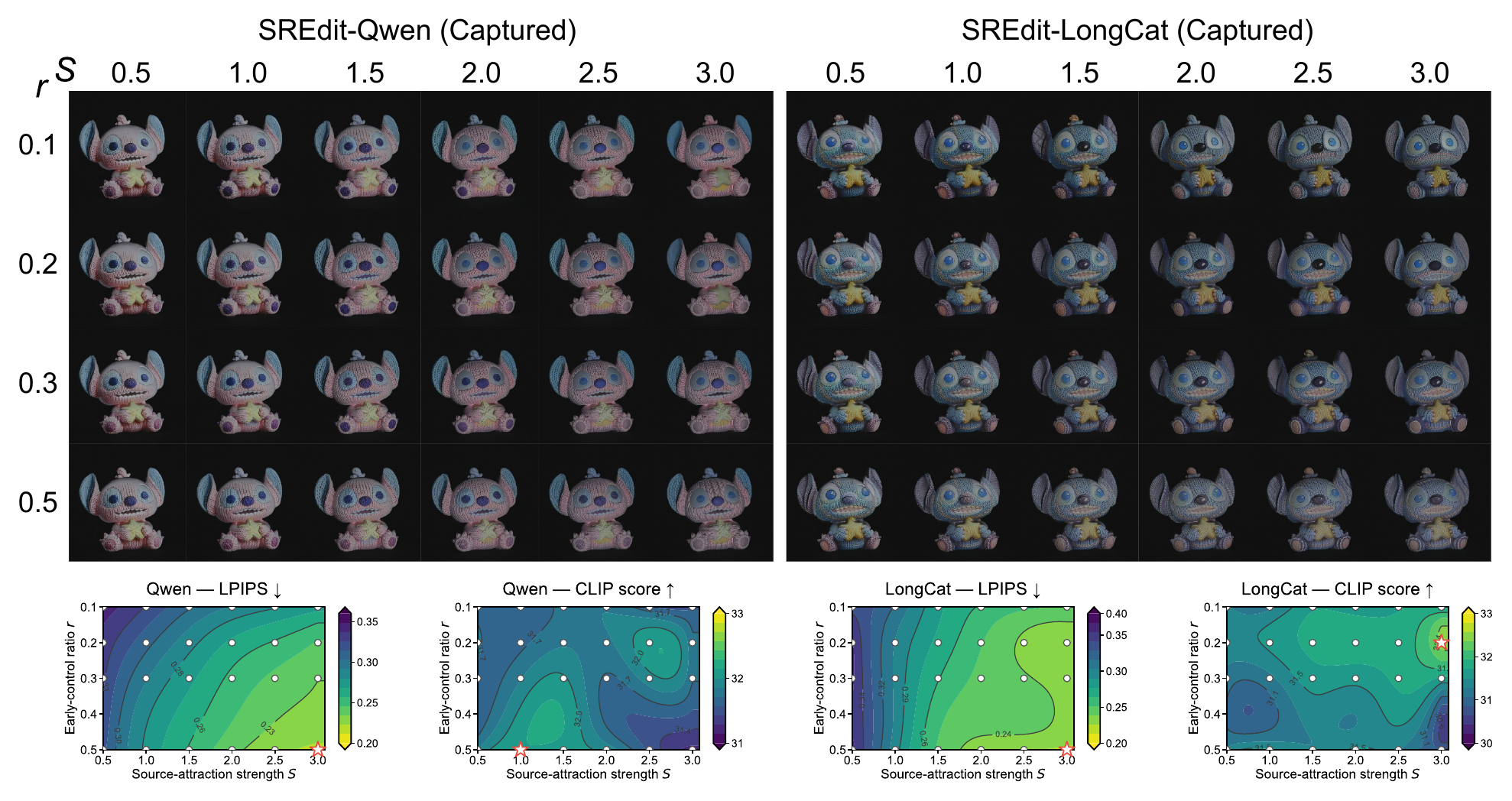}
    \caption{\textbf{Control-parameter ablation.}
    All results are physical projector--camera captures.
    Top: complete \(r\)-by-\(S\) parameter grids for both image editors.
    Bottom: mean CLIP and LPIPS for SREdit-Qwen and SREdit-LongCat over 18 cases (six setups, three prompts each).
    Dots mark measured settings, contours show interpolated trends, and red stars mark the best measured setting in each plot.}
    \label{fig:parameter-ablation}
\end{figure*}

Across 720 judgments, SREdit achieved an overall preference rate of 72.4\%.
After aggregating repeated judgments by participant, this preference remained significantly above chance (two-sided sign test with Holm correction, \(p<0.001\)).
In addition, 96.5\% of SREdit results were judged free of visible misalignment, compared with 80.4\% for the Native versions.

\subsection{Reference-guided editing}

We further evaluate SREdit for reference-guided appearance editing, where an additional image specifies the desired object appearance.
The LongCat implementation used in our experiments does not accept an additional reference image, whereas Qwen supports multi-image conditioning directly.
We therefore conduct this experiment only with Qwen.
We test three reference-guided edits on each of four physical setups, yielding 12 cases.

\cref{fig:reference-material-editing} compares Native-Qwen with SREdit-Qwen.
Both methods use the same source and reference images.

Native-Qwen generally follows the reference but sometimes alters regions unrelated to the requested edit.
It also tends to introduce saturated colors, strong highlights, and synthesized shading.
These changes may conflict with the fixed object geometry or exceed the available projection range at the corresponding surface region.
After projection, this appears as geometric misalignment, clipped highlights, or color shifts.
SREdit retains more of the observed surface color and shading in regions with lower response while allowing stronger changes in regions with higher response.
The captured results therefore preserve registration with the physical object more consistently and exhibit fewer color and shading artifacts.

\cref{tab:reference-editing} reports the quantitative results.
SREdit improves all four reproduction metrics; for example, PSNR increases from 18.5257\,dB to 21.7335\,dB.
This indicates that the targets produced with response-guided sampling can be reproduced more faithfully by the physical projection system.

\subsection{Ablation studies}

\begin{figure}[!t]
    \centering
    \includegraphics[width=\columnwidth]{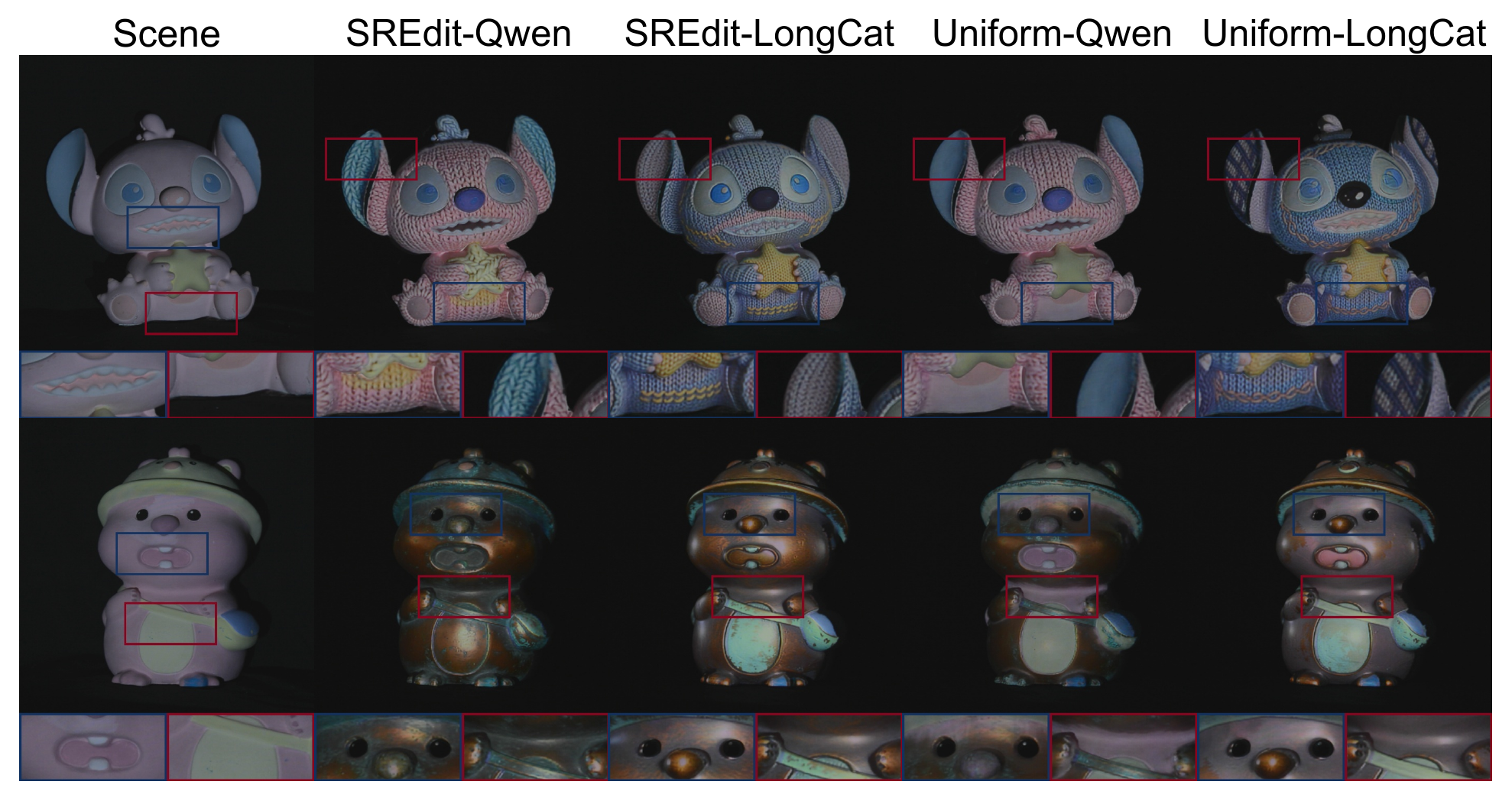}
    \caption{{\textbf{Scene response map ablation.}
    Camera-captured results under real projection.
    Uniform replaces the scene response map with its mean while retaining the paired \(r\) and \(S\) values.
    Insets show unedited patches that retain the source color under Uniform; SREdit shows fewer such patches.}}
    \label{fig:uniform-ablation}
\end{figure}

We compare the scene response map with a Uniform variant.
Uniform replaces the complete map with the mean over the complete image, applying the same response value throughout the image.
All other settings are identical to those used by SREdit.
The aggregate quantitative comparison is shown in \cref{tab:response-map-ablation}, with qualitative results in \cref{fig:uniform-ablation}.

Uniform favors more conservative edits and therefore obtains higher reproduction scores, whereas the scene response map improves CLIP and BLIP-ITM for both image editors.

As shown in \cref{fig:uniform-ablation}, however, a uniform control weight does not produce a spatially uniform perceptual edit because the residual between the Euler-updated latent and the source anchor varies across spatial locations.
In the camera-captured results, Uniform therefore leaves unedited patches that retain the source color while suppressing editable content elsewhere.
The scene response map preserves regions with lower response more coherently and leaves regions with higher response greater freedom to realize the requested appearance.
This spatial weighting by the response map produces fewer unedited patches in these examples, consistent with the higher semantic agreement measured by CLIP and BLIP-ITM.

We next examine the influence of the early control ratio \(r\) and the control strength \(S\).
\cref{fig:parameter-ablation} shows the complete parameter grid for both image editors and summarizes the corresponding quantitative trends over all 18 cases from six setups.

The two parameters control complementary aspects of response-guided sampling.
The control strength \(S\) determines the overall magnitude of correction toward the source anchor, whereas \(r\) determines how its normalized temporal weight is distributed over the early sampling stage.
A smaller \(r\) concentrates the control in the earliest steps, while a larger \(r\) distributes it over a longer portion of the sampling trajectory.
Thus, changing \(r\) redistributes the control without changing its total control weight.

As shown in \cref{fig:parameter-ablation}, a small \(S\) provides limited guidance regardless of \(r\), and the results remain closer to those of the corresponding Native version.
Increasing \(S\) strengthens source preservation, particularly in regions with lower response, and generally reduces geometric misalignment and appearance variations that are difficult to reproduce under projection.
With a relatively small \(r\), the control is concentrated at the beginning of sampling.
This stabilizes registration, coarse color, and shading early, while leaving more of the sampling trajectory for texture refinement.

The influence of \(r\) becomes more visible as \(S\) increases.
Extending the controlled stage keeps the source-anchor correction active over a longer portion of the sampling trajectory, but an overly large \(r\) combined with a strong \(S\) may suppress local texture and shading details.
For both editors, the four reproduction metrics improve monotonically with \(S\) at every \(r\) and are best at \(r=0.5, S=3.0\), which also gives the lowest CLIP score in the grid.
In practice, we recommend keeping \(r\) around 0.2 and adjusting \(S\): increase it when the captured result shows misalignment or clipped highlights, and decrease it when the requested texture is weakened.

Additional qualitative results for the parameter sweep and per-setup metric heatmaps are provided in the supplementary material.

\FloatBarrier
\section{Discussion}

\subsection{Limitations and future work}

{\noindent\textbf{Editing and reproduction trade-off.}
The early control ratio \(r\) and control strength \(S\) determine the balance between physical reproduction and the semantic and textural richness of the edit.
Increasing \(S\) or extending the controlled portion of sampling tends to preserve registration and the observed surface appearance, but may weaken the requested texture.
Weaker or shorter control leaves more freedom for the image editor, but may produce changes that are difficult to reproduce through projection.
Because the scene response differs across physical setups, a single parameter pair cannot provide the preferred balance in every case.}

\begin{figure}[!h]
    \centering
    \includegraphics[width=\columnwidth]{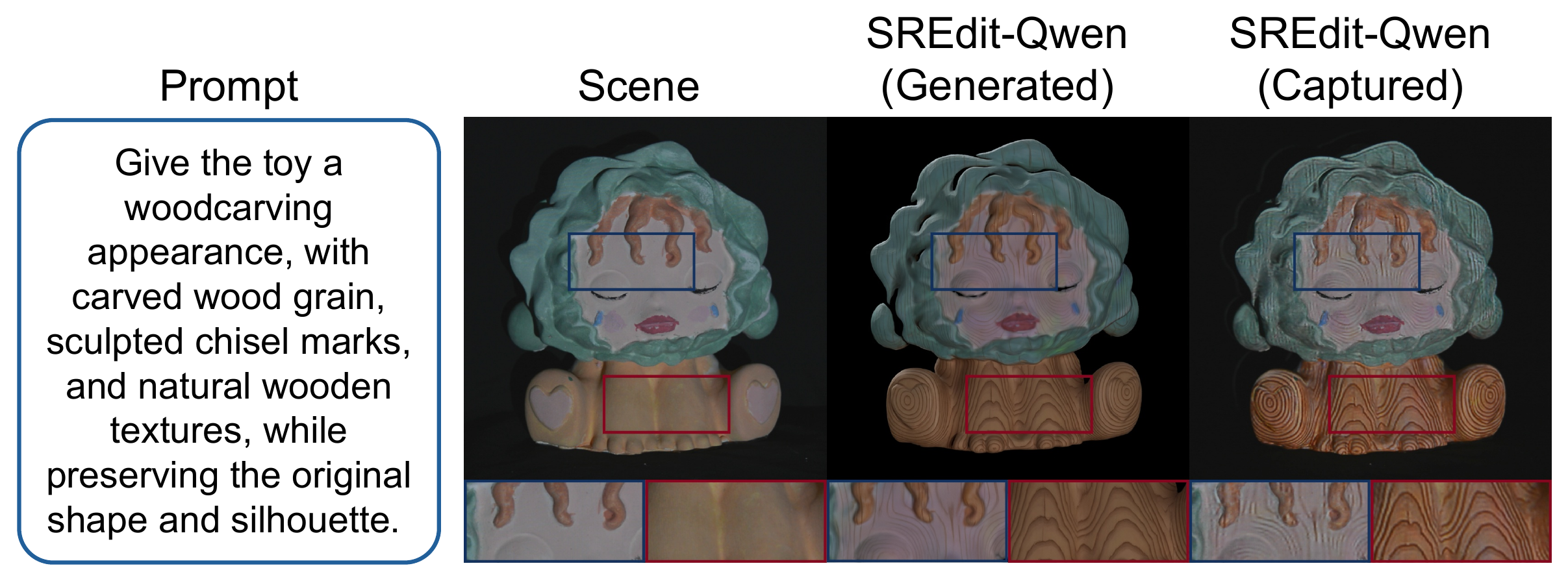}
    \caption{{\textbf{Failure case.}
    This result uses \(r=0.2\) and \(S=2.5\).
    Although it remains registered with the physical object, parts of the requested wood texture are weakened or spatially uneven in the intended and camera-captured appearances.}}
    \label{fig:failure-case}
    \vspace{-0.7em}
\end{figure}

{\cref{fig:failure-case} shows this trade-off under a fixed setting: registration is retained, but part of the requested texture is weakened and spatially uneven.
The preferred parameters therefore depend on whether the user prioritizes closer physical reproduction or a semantically richer edit.
Future work may select \(r\) and \(S\) automatically for each setup and editing request.}

\noindent\textbf{Computational requirements.}
SREdit does not train or fine-tune the image editor, attach an additional control module, or backpropagate gradients through the editor during generation.
Nevertheless, the pretrained image editors are large and still dominate the runtime and memory requirements of the complete pipeline.
Future work should examine how response-guided sampling can be used with lighter image editors.

\noindent\textbf{{Soft constraint.}}
Response-guided sampling acts as a soft constraint: it biases the edit according to the relative scene response but does not explicitly exclude all appearances that are difficult to reproduce.
Under extreme surface colors, a large difference between the source and requested appearances, or overly strong control, sampling may approach a degenerate solution in which the requested texture is completely suppressed.
Future work should seek an efficient hard constraint that prevents this failure without gradient computation or an additional model during generation.

\section{Conclusion}

In this paper, we introduced SREdit, an efficient response-guided framework that incorporates scene response into pretrained image editing for projection mapping while separating appearance editing from downstream projector compensation.
SREdit estimates a scene response map from the reconstructed scene that reflects how strongly projected illumination can affect different surface regions.
During early sampling, it uses the source appearance to preserve registration with the physical object and adapts the amount of editing according to the scene response.
SREdit requires no additional training or control module for the image editor, does not backpropagate through the editor during generation, and leaves downstream projector compensation unchanged.
Experiments with two pretrained image editors across six physical setups show consistent improvements in projected appearance quality and reproduction fidelity.
Response-guided sampling adds less than one second to image generation with negligible VRAM overhead, and participants preferred SREdit in 72.4\% of the user-study comparisons.
These results show that scene response can guide pretrained image editing before projector compensation with little additional cost.

\bibliographystyle{abbrv-doi}
\bibliography{references}

\clearpage
\pagestyle{plain}

\appendix

\setcounter{figure}{0}
\setcounter{table}{0}
\renewcommand{\thefigure}{S\arabic{figure}}
\renewcommand{\thetable}{S\arabic{table}}
\renewcommand{\theHfigure}{supp.\arabic{figure}}
\renewcommand{\theHtable}{supp.\arabic{table}}

\begin{figure*}[p]
    \centering
    \setlength{\abovecaptionskip}{1pt}
    \setlength{\belowcaptionskip}{1pt}
    \includegraphics[width=0.95\textwidth,trim=1pt 1pt 2pt 6pt,clip]{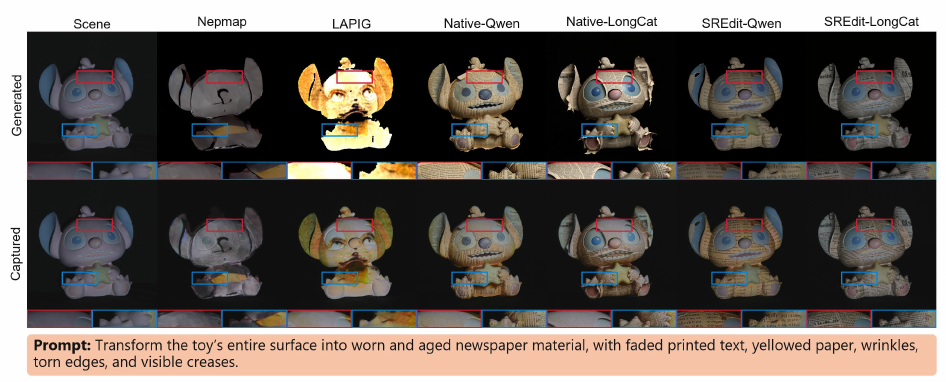}
    \caption{\textbf{Qualitative comparison for Case~1.} SREdit uses
    \((r,S)=(0.2,2.5)\) for Qwen and \((0.2,3.0)\) for LongCat. The upper
    and lower rows show generated appearances and camera-captured projection
    results, respectively.}
    \label{fig:case1-qualitative}
\end{figure*}

\begin{figure*}[p]
    \centering
    \setlength{\abovecaptionskip}{1pt}
    \setlength{\belowcaptionskip}{1pt}
    \includegraphics[width=0.95\textwidth]{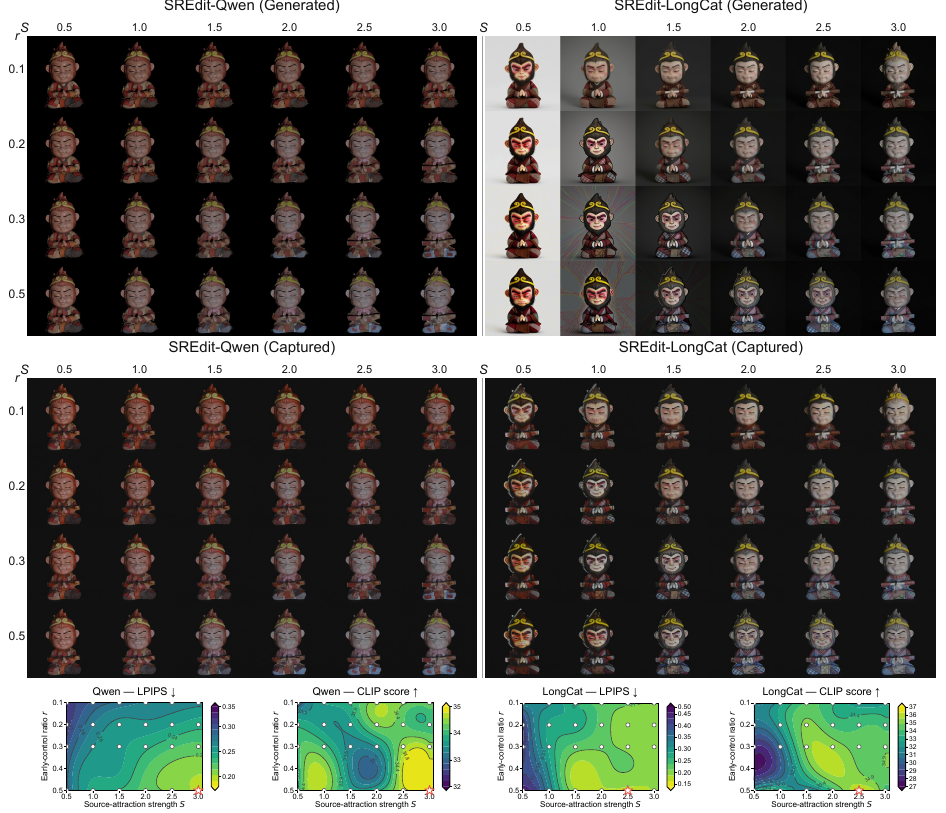}
    \caption{\textbf{Parameter ablation for Case~10.} Columns vary
    \(S\), while rows vary \(r\). The upper and lower matrices show generated
    and camera-captured results, respectively, and the bottom plots report
    LPIPS and CLIP over the same parameter grid.}
    \label{fig:case10-parameter-ablation}
\end{figure*}

\twocolumn[
\begin{center}
    {\fontsize{15}{18}\fontencoding{OT1}\fontfamily{phv}\fontseries{b}\selectfont
    SREdit: Scene Response Guided Appearance Editing for Projection\par
    Mapping}

    \vspace{0.65em}
    {\large\sffamily\textemdash\quad\textit{Supplementary Materials}
    \quad\textemdash}
\end{center}
\vspace{0.7em}
\begin{minipage}{\textwidth}
\section{Introduction}
In this supplementary material, we present additional qualitative comparisons
and parameter ablations across six physical setups and diverse text prompts.
Figures~\ref{fig:case1-qualitative}--%
\ref{fig:case10-parameter-ablation} include both generated
appearances and camera-captured projection results, providing a more complete
evaluation of SREdit under real projection. We further provide the complete
user study protocol and all 36 comparison trials.

\section{Experimental configuration}
{
\normalsize
\renewcommand{\arraystretch}{1.24}

\begin{tabular*}{\textwidth}{@{\extracolsep{\fill}}p{0.20\textwidth}p{0.75\textwidth}@{}}
\toprule
\textbf{Configuration} & \textbf{Setting} \\
\midrule
Physical system & O.B.E C3 Air projector and Nikon D3200 camera. The projector
and camera poses and the camera settings remain fixed within each setup. \\
Compute & AMD EPYC 7642 CPU and NVIDIA A100 GPU with 64~GB of
available memory. \\
Image editors & Qwen-Image-Edit-2511 (Qwen) with 40 sampling steps and
LongCat-Image-Edit (LongCat) with 50 sampling steps. \\
Main parameters & Qwen uses \(r=0.2\) and \(S=2.5\); LongCat uses \(r=0.2\)
and \(S=3.0\). \\
Parameter sweep & \(r\in\{0.1,0.2,0.3,0.5\}\) and
\(S\in\{0.5,1.0,1.5,2.0,2.5,3.0\}\). \\
Projection pipeline &
The Gaussian-based differentiable scene renderer remains frozen,
while iterative refinement updates only the compensation image. \\
Evaluation & Six physical setups and 18 text prompts. PSNR, SSIM, LPIPS, and
\(\Delta E\) are computed within the object region; CLIP and BLIP-ITM
measure agreement with the target description. \\
\bottomrule
\end{tabular*}

\vspace{0.45em}
\noindent\textbf{Comparison protocol.}
All reported projection results are obtained by physically projecting the
compensation image and capturing the result with the camera. For each
image editor, the Native and SREdit
versions use the same source appearance, editing request, number of sampling steps,
and compensation procedure. Qwen and LongCat receive
the same editing request for each case. For Nepmap and LAPIG, we adapt the
request to the input format required by each method while retaining the same
target appearance.

\vspace{0.45em}
\noindent\textbf{Scene response map construction.}
For each reconstructed object, we evaluate the channelwise
direct-response factor, average it across the three color channels, and
normalize it over the reconstructed-object support to obtain the scene response
map used during editing. The map is computed once for each setup and remains
fixed across its editing requests.

\vspace{0.45em}
\noindent\textbf{Ablation protocol.}
The parameter study evaluates every combination of the listed \(r\) and \(S\)
values using both generated appearances and camera-captured results.
In the scene response map ablation,
the uniform map takes the mean over the complete scene response map
and assigns it to every image location. The
image editor, sampling schedule,
control parameters, and
compensation procedure remain unchanged.
}

\section{User study configuration}
{
\normalsize
\renewcommand{\arraystretch}{1.20}

\begin{tabular*}{\textwidth}{@{\extracolsep{\fill}}p{0.20\textwidth}p{0.75\textwidth}@{}}
\toprule
\textbf{Configuration} & \textbf{Setting} \\
\midrule
Participants & 20 participants (10 female and 10 male), aged 18--24
(mean 20.25, SD 1.37); nine reported a background in image processing, computer
graphics, or computer vision. \\
Study design & Anonymous, within-subject pairwise comparison conducted online. \\
Trials & 36 camera-captured comparison pairs: 18 Qwen cases and 18 LongCat
cases. \\
Compared results & the Native and SREdit versions, displayed anonymously as A and B;
their positions were randomized and recorded for every trial. \\
Evaluation criteria & Text alignment, artifacts, lighting and surface realism,
visible misalignment, and overall preference. \\
Responses & Four-level forced choice for comparative questions; visible
misalignment was judged separately for A and B as present or absent. \\
\bottomrule
\end{tabular*}

\vspace{0.4em}
Participation was voluntary and anonymous. Before beginning the study,
participants reviewed the study information and provided informed consent.
Comparative responses were mapped back to the method shown on each side before
aggregation. The complete instructions and all 36 formal trials are provided
at the end of this supplementary material.
}

\section{Representative main and ablation results}
\noindent\textbf{Main comparisons.}
The following pages present ten representative cases. Each comparison includes
the generated appearances and the corresponding camera-captured projection
results.

\vspace{0.35em}
\noindent\textbf{Parameter ablations.}
Each main comparison is paired with the generated and captured \(r\)-by-\(S\)
grids for Qwen and LongCat, together with the corresponding metric heat maps.
\end{minipage}
\vspace{0.5em}
]

\begin{figure*}[p]
    \centering
    \setlength{\abovecaptionskip}{1pt}
    \setlength{\belowcaptionskip}{1pt}
    \includegraphics[width=0.95\textwidth]{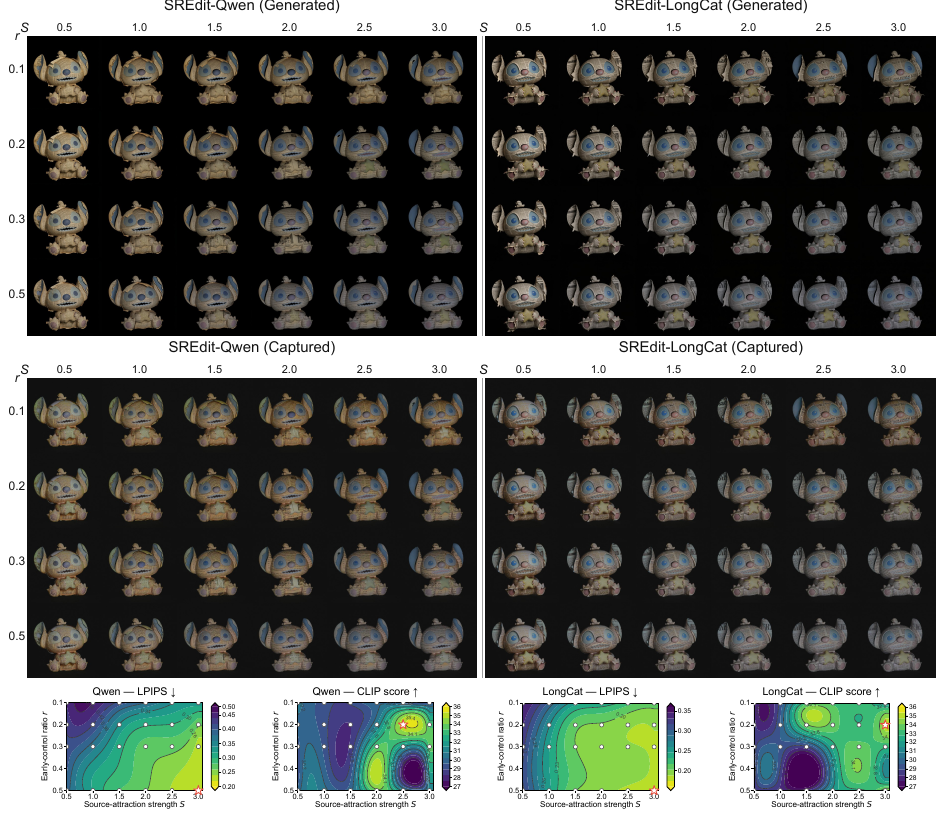}
    \caption{\textbf{Parameter ablation for Case~1.} Columns vary
    \(S\), while rows vary \(r\). The upper and lower matrices show generated
    and camera-captured results, respectively, and the bottom plots report
    LPIPS and CLIP over the same parameter grid.}
    \label{fig:case1-parameter-ablation}
\end{figure*}%
\clearpage

\begin{figure*}[p]
    \centering
    \setlength{\abovecaptionskip}{1pt}
    \setlength{\belowcaptionskip}{1pt}
    \includegraphics[width=0.95\textwidth,trim=1pt 1pt 2pt 6pt,clip]{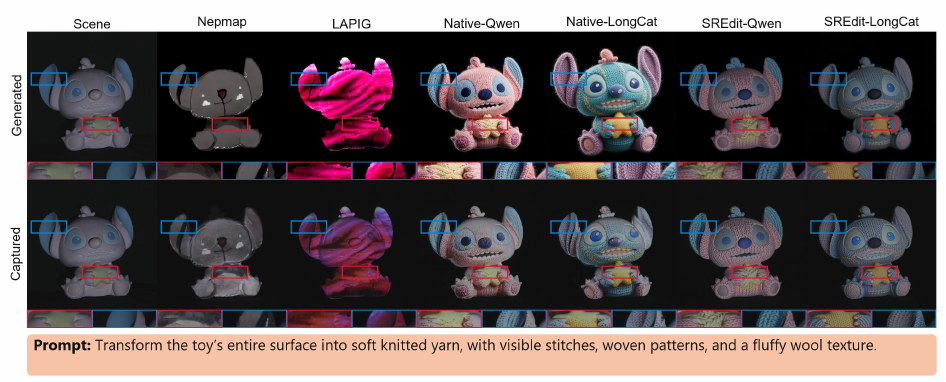}
    \caption{\textbf{Qualitative comparison for Case~2.} SREdit uses
    \((r,S)=(0.2,2.5)\) for Qwen and \((0.2,3.0)\) for LongCat. The upper
    and lower rows show generated appearances and camera-captured projection
    results, respectively.}
    \label{fig:case2-qualitative}
\end{figure*}

\begin{figure*}[p]
    \centering
    \setlength{\abovecaptionskip}{1pt}
    \setlength{\belowcaptionskip}{1pt}
    \includegraphics[width=0.95\textwidth]{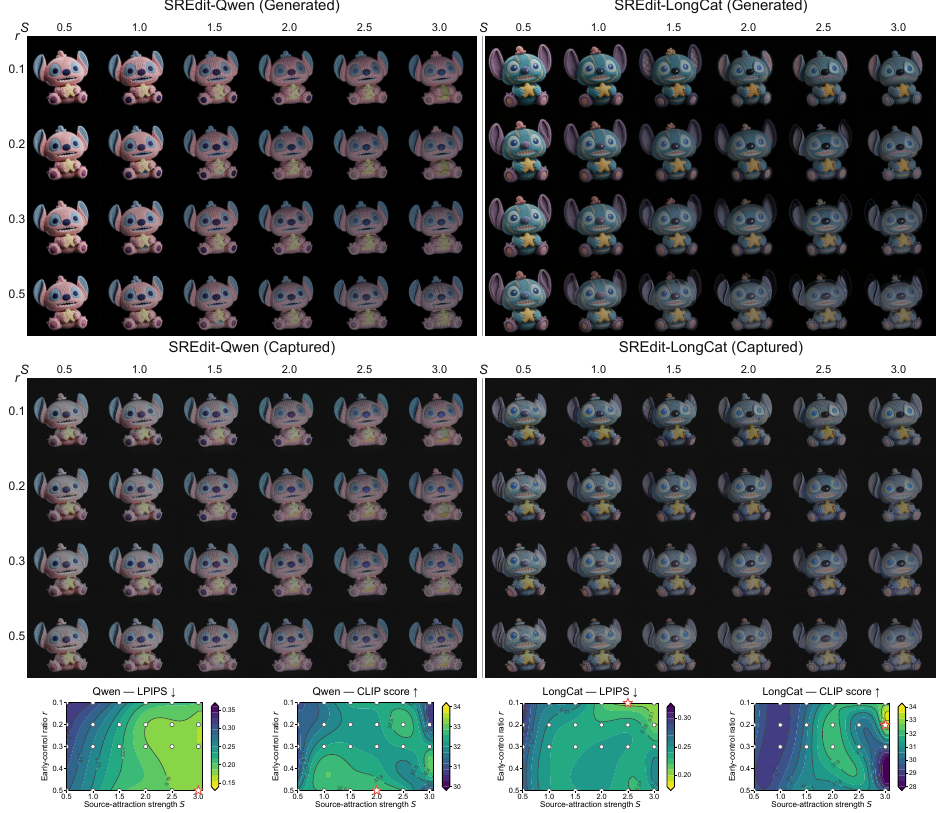}
    \caption{\textbf{Parameter ablation for Case~2.} Columns vary
    \(S\), while rows vary \(r\). The upper and lower matrices show generated
    and camera-captured results, respectively, and the bottom plots report
    LPIPS and CLIP over the same parameter grid.}
    \label{fig:case2-parameter-ablation}
\end{figure*}%
\clearpage

\begin{figure*}[p]
    \centering
    \setlength{\abovecaptionskip}{1pt}
    \setlength{\belowcaptionskip}{1pt}
    \includegraphics[width=0.95\textwidth,trim=1pt 1pt 2pt 6pt,clip]{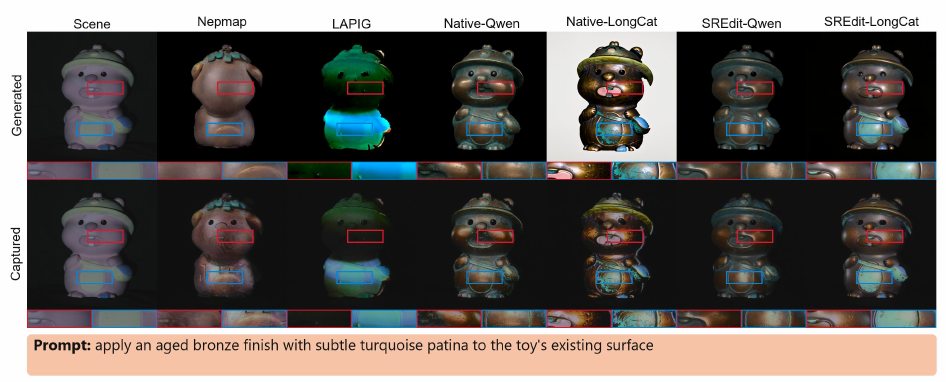}
    \caption{\textbf{Qualitative comparison for Case~3.} SREdit uses
    \((r,S)=(0.2,2.5)\) for Qwen and \((0.2,3.0)\) for LongCat. The upper
    and lower rows show generated appearances and camera-captured projection
    results, respectively.}
    \label{fig:case3-qualitative}
\end{figure*}

\begin{figure*}[p]
    \centering
    \setlength{\abovecaptionskip}{1pt}
    \setlength{\belowcaptionskip}{1pt}
    \includegraphics[width=0.95\textwidth]{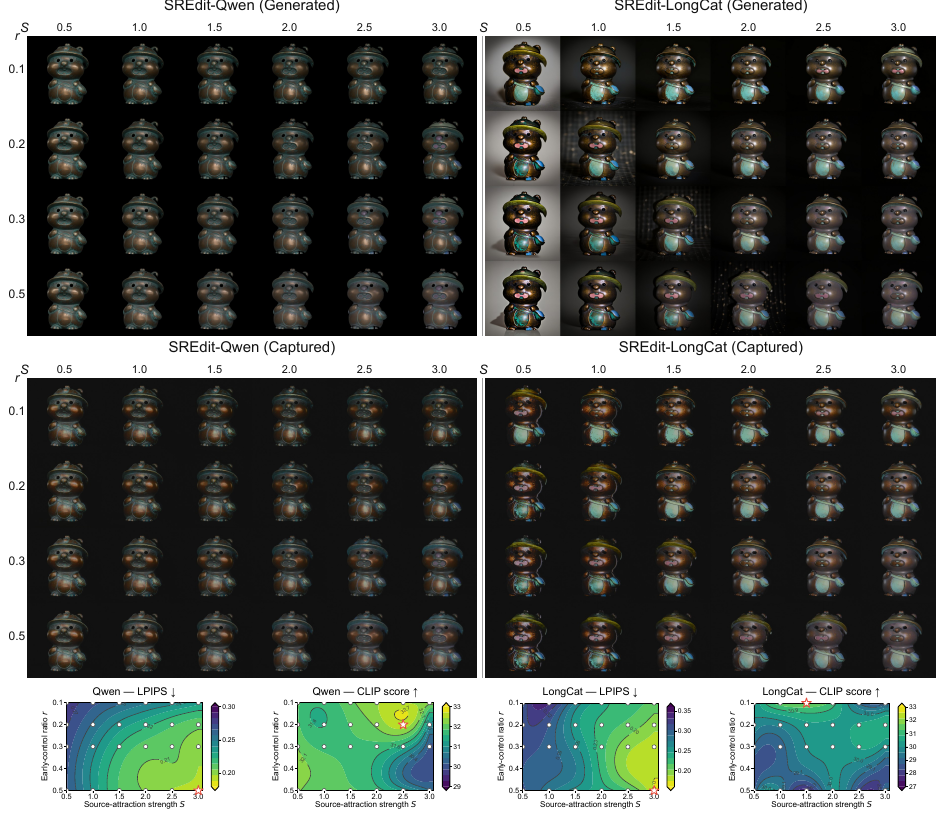}
    \caption{\textbf{Parameter ablation for Case~3.} Columns vary
    \(S\), while rows vary \(r\). The upper and lower matrices show generated
    and camera-captured results, respectively, and the bottom plots report
    LPIPS and CLIP over the same parameter grid.}
    \label{fig:case3-parameter-ablation}
\end{figure*}%
\clearpage

\begin{figure*}[p]
    \centering
    \setlength{\abovecaptionskip}{1pt}
    \setlength{\belowcaptionskip}{1pt}
    \includegraphics[width=0.95\textwidth,trim=1pt 1pt 2pt 6pt,clip]{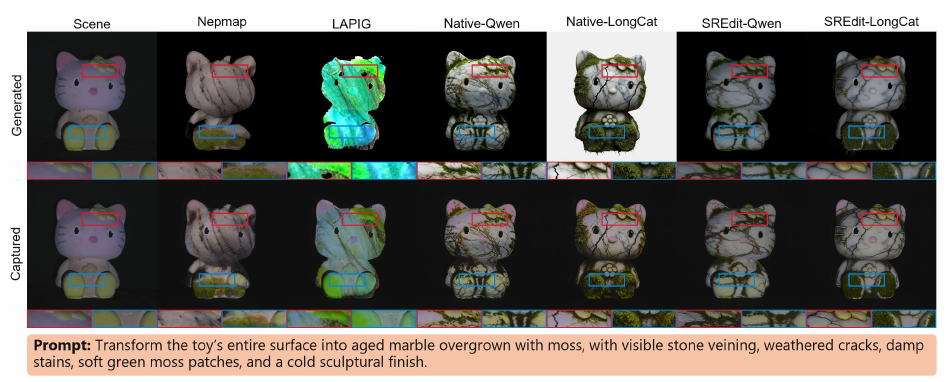}
    \caption{\textbf{Qualitative comparison for Case~4.} SREdit uses
    \((r,S)=(0.2,2.5)\) for Qwen and \((0.2,3.0)\) for LongCat. The upper
    and lower rows show generated appearances and camera-captured projection
    results, respectively.}
    \label{fig:case4-qualitative}
\end{figure*}

\begin{figure*}[p]
    \centering
    \setlength{\abovecaptionskip}{1pt}
    \setlength{\belowcaptionskip}{1pt}
    \includegraphics[width=0.95\textwidth]{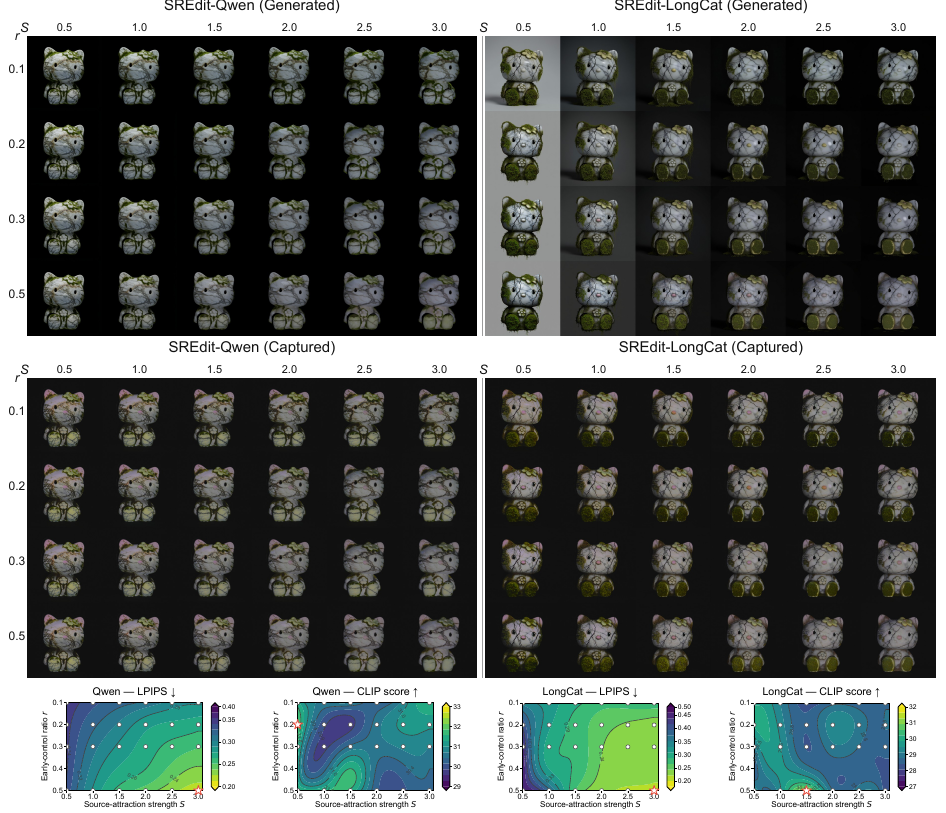}
    \caption{\textbf{Parameter ablation for Case~4.} Columns vary
    \(S\), while rows vary \(r\). The upper and lower matrices show generated
    and camera-captured results, respectively, and the bottom plots report
    LPIPS and CLIP over the same parameter grid.}
    \label{fig:case4-parameter-ablation}
\end{figure*}%
\clearpage

\begin{figure*}[p]
    \centering
    \setlength{\abovecaptionskip}{1pt}
    \setlength{\belowcaptionskip}{1pt}
    \includegraphics[width=0.95\textwidth,trim=1pt 1pt 2pt 6pt,clip]{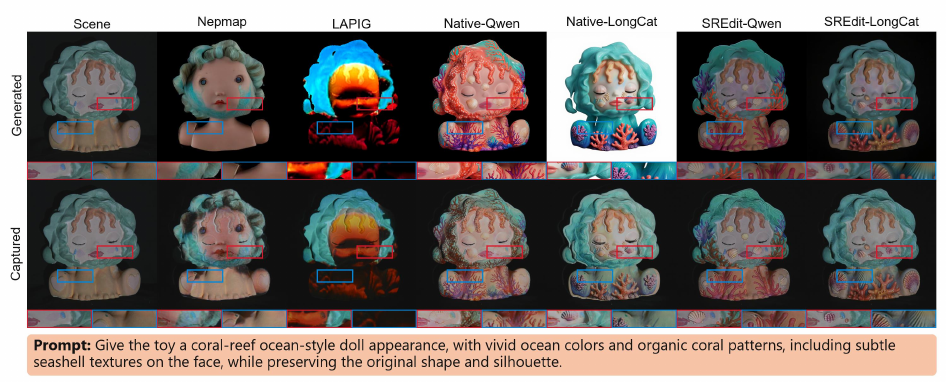}
    \caption{\textbf{Qualitative comparison for Case~5.} SREdit uses
    \((r,S)=(0.2,2.5)\) for Qwen and \((0.2,3.0)\) for LongCat. The upper
    and lower rows show generated appearances and camera-captured projection
    results, respectively.}
    \label{fig:case5-qualitative}
\end{figure*}

\begin{figure*}[p]
    \centering
    \setlength{\abovecaptionskip}{1pt}
    \setlength{\belowcaptionskip}{1pt}
    \includegraphics[width=0.95\textwidth]{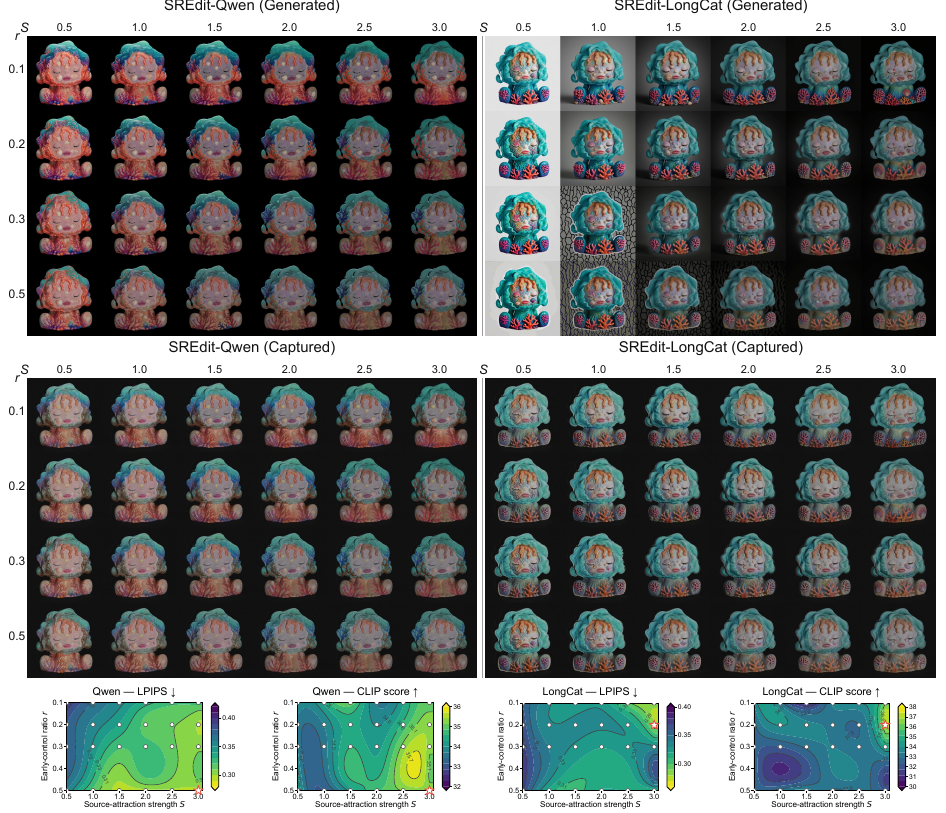}
    \caption{\textbf{Parameter ablation for Case~5.} Columns vary
    \(S\), while rows vary \(r\). The upper and lower matrices show generated
    and camera-captured results, respectively, and the bottom plots report
    LPIPS and CLIP over the same parameter grid.}
    \label{fig:case5-parameter-ablation}
\end{figure*}%
\clearpage

\begin{figure*}[p]
    \centering
    \setlength{\abovecaptionskip}{1pt}
    \setlength{\belowcaptionskip}{1pt}
    \includegraphics[width=0.95\textwidth,trim=1pt 1pt 2pt 6pt,clip]{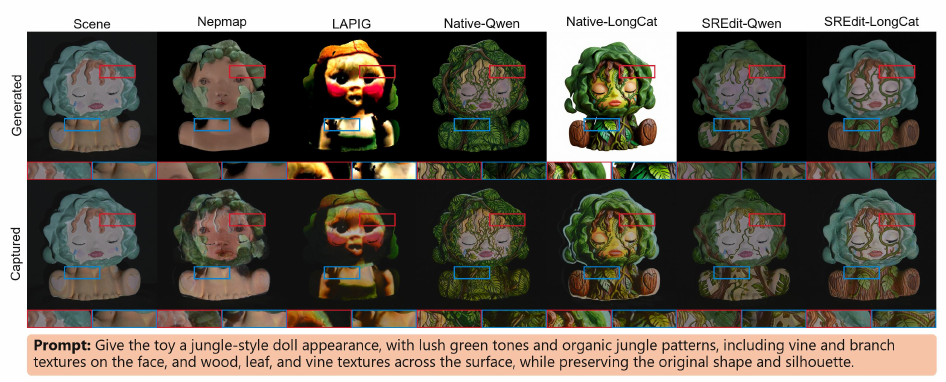}
    \caption{\textbf{Qualitative comparison for Case~6.} SREdit uses
    \((r,S)=(0.2,2.5)\) for Qwen and \((0.2,3.0)\) for LongCat. The upper
    and lower rows show generated appearances and camera-captured projection
    results, respectively.}
    \label{fig:case6-qualitative}
\end{figure*}

\begin{figure*}[p]
    \centering
    \setlength{\abovecaptionskip}{1pt}
    \setlength{\belowcaptionskip}{1pt}
    \includegraphics[width=0.95\textwidth]{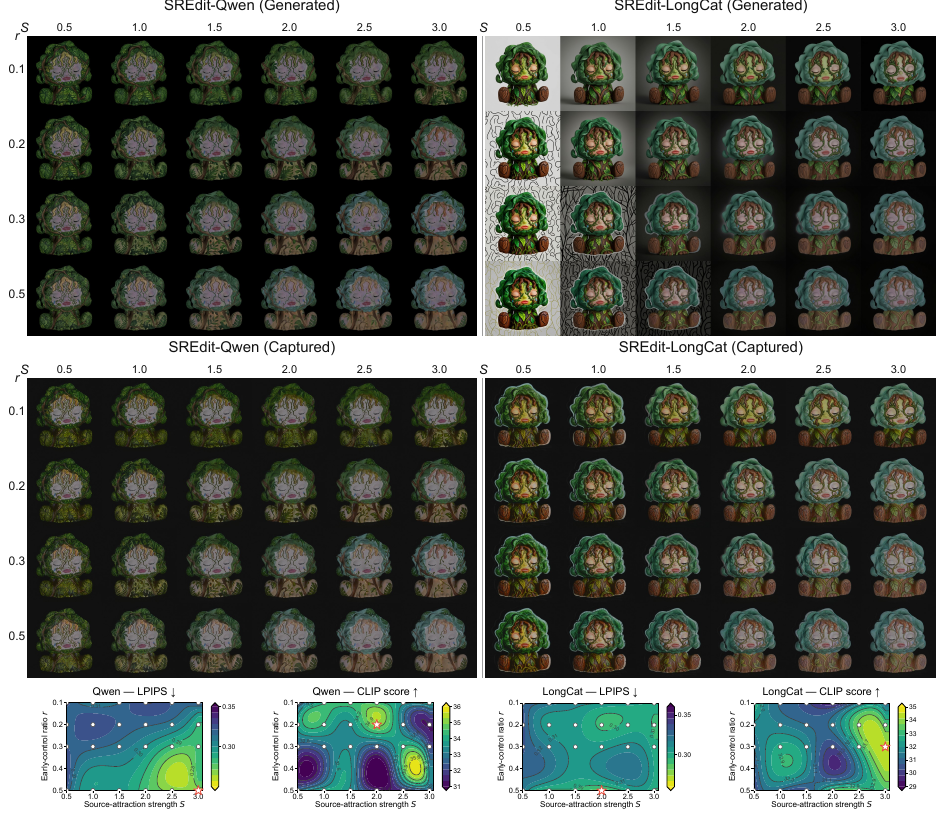}
    \caption{\textbf{Parameter ablation for Case~6.} Columns vary
    \(S\), while rows vary \(r\). The upper and lower matrices show generated
    and camera-captured results, respectively, and the bottom plots report
    LPIPS and CLIP over the same parameter grid.}
    \label{fig:case6-parameter-ablation}
\end{figure*}%
\clearpage

\begin{figure*}[p]
    \centering
    \setlength{\abovecaptionskip}{1pt}
    \setlength{\belowcaptionskip}{1pt}
    \includegraphics[width=0.95\textwidth,trim=1pt 1pt 2pt 6pt,clip]{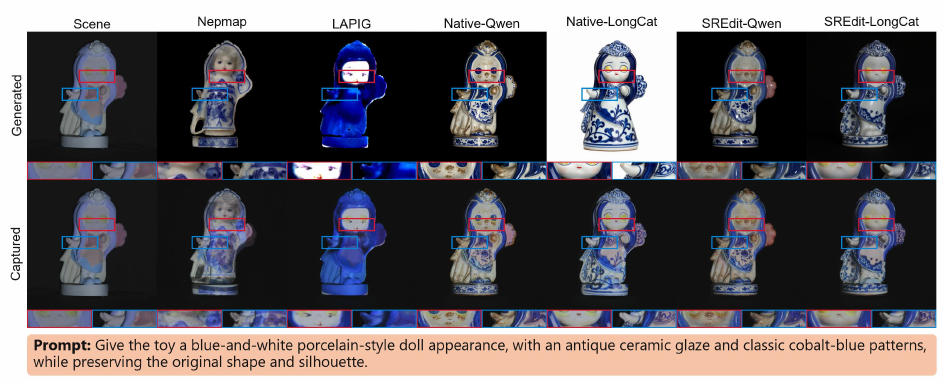}
    \caption{\textbf{Qualitative comparison for Case~7.} SREdit uses
    \((r,S)=(0.2,2.5)\) for Qwen and \((0.2,3.0)\) for LongCat. The upper
    and lower rows show generated appearances and camera-captured projection
    results, respectively.}
    \label{fig:case7-qualitative}
\end{figure*}

\begin{figure*}[p]
    \centering
    \setlength{\abovecaptionskip}{1pt}
    \setlength{\belowcaptionskip}{1pt}
    \includegraphics[width=0.95\textwidth]{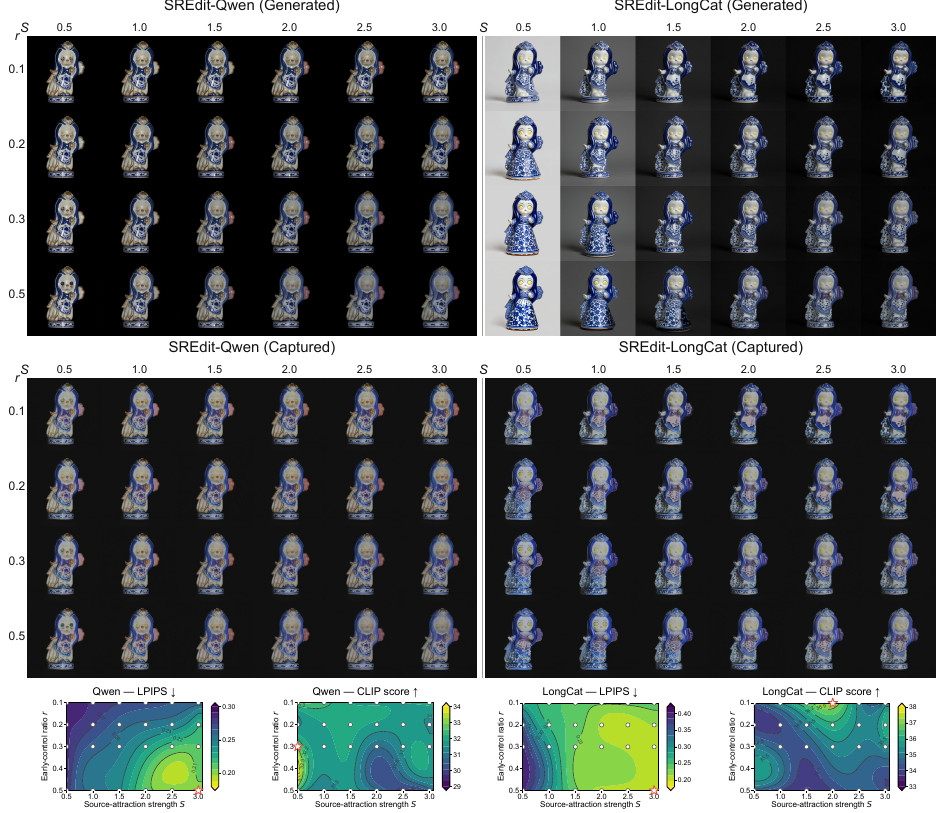}
    \caption{\textbf{Parameter ablation for Case~7.} Columns vary
    \(S\), while rows vary \(r\). The upper and lower matrices show generated
    and camera-captured results, respectively, and the bottom plots report
    LPIPS and CLIP over the same parameter grid.}
    \label{fig:case7-parameter-ablation}
\end{figure*}%
\clearpage

\begin{figure*}[p]
    \centering
    \setlength{\abovecaptionskip}{1pt}
    \setlength{\belowcaptionskip}{1pt}
    \includegraphics[width=0.95\textwidth,trim=1pt 1pt 2pt 6pt,clip]{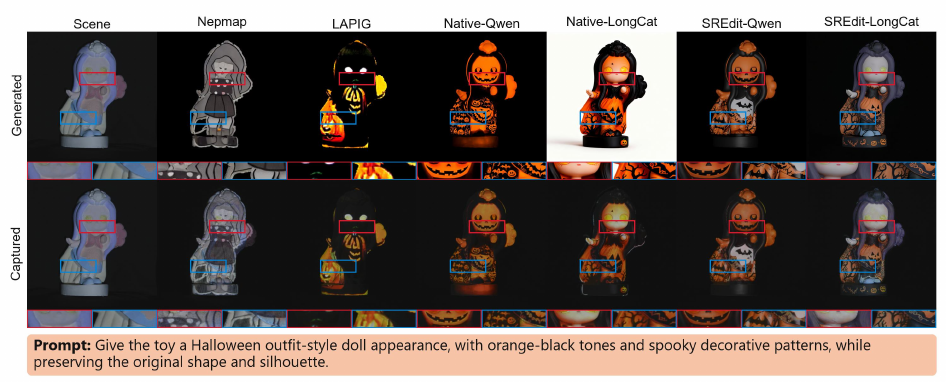}
    \caption{\textbf{Qualitative comparison for Case~8.} SREdit uses
    \((r,S)=(0.2,2.5)\) for Qwen and \((0.2,3.0)\) for LongCat. The upper
    and lower rows show generated appearances and camera-captured projection
    results, respectively.}
    \label{fig:case8-qualitative}
\end{figure*}

\begin{figure*}[p]
    \centering
    \setlength{\abovecaptionskip}{1pt}
    \setlength{\belowcaptionskip}{1pt}
    \includegraphics[width=0.95\textwidth]{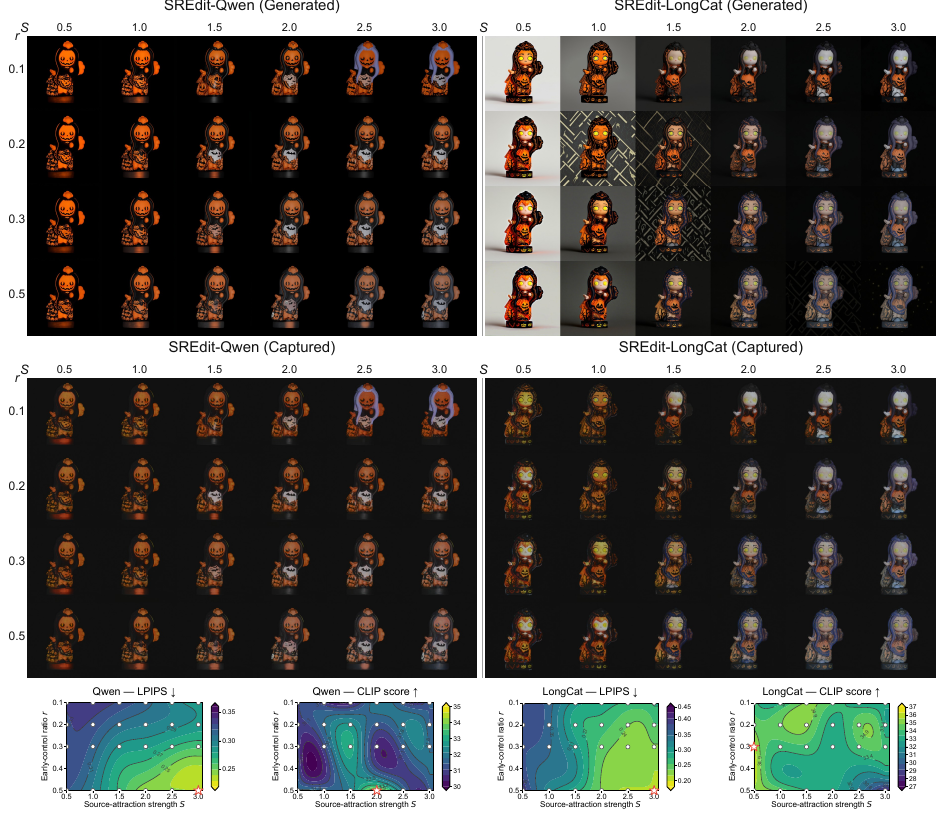}
    \caption{\textbf{Parameter ablation for Case~8.} Columns vary
    \(S\), while rows vary \(r\). The upper and lower matrices show generated
    and camera-captured results, respectively, and the bottom plots report
    LPIPS and CLIP over the same parameter grid.}
    \label{fig:case8-parameter-ablation}
\end{figure*}%
\clearpage

\begin{figure*}[p]
    \centering
    \setlength{\abovecaptionskip}{1pt}
    \setlength{\belowcaptionskip}{1pt}
    \includegraphics[width=0.95\textwidth,trim=1pt 1pt 2pt 6pt,clip]{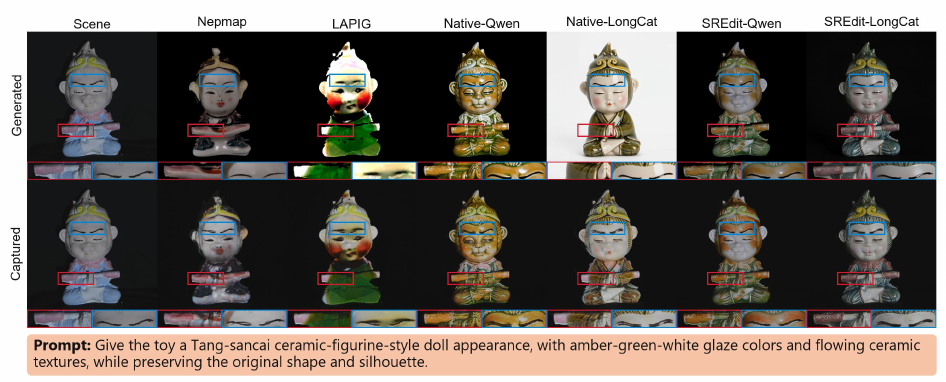}
    \caption{\textbf{Qualitative comparison for Case~9.} SREdit uses
    \((r,S)=(0.2,2.5)\) for Qwen and \((0.2,3.0)\) for LongCat. The upper
    and lower rows show generated appearances and camera-captured projection
    results, respectively.}
    \label{fig:case9-qualitative}
\end{figure*}

\begin{figure*}[p]
    \centering
    \setlength{\abovecaptionskip}{1pt}
    \setlength{\belowcaptionskip}{1pt}
    \includegraphics[width=0.95\textwidth]{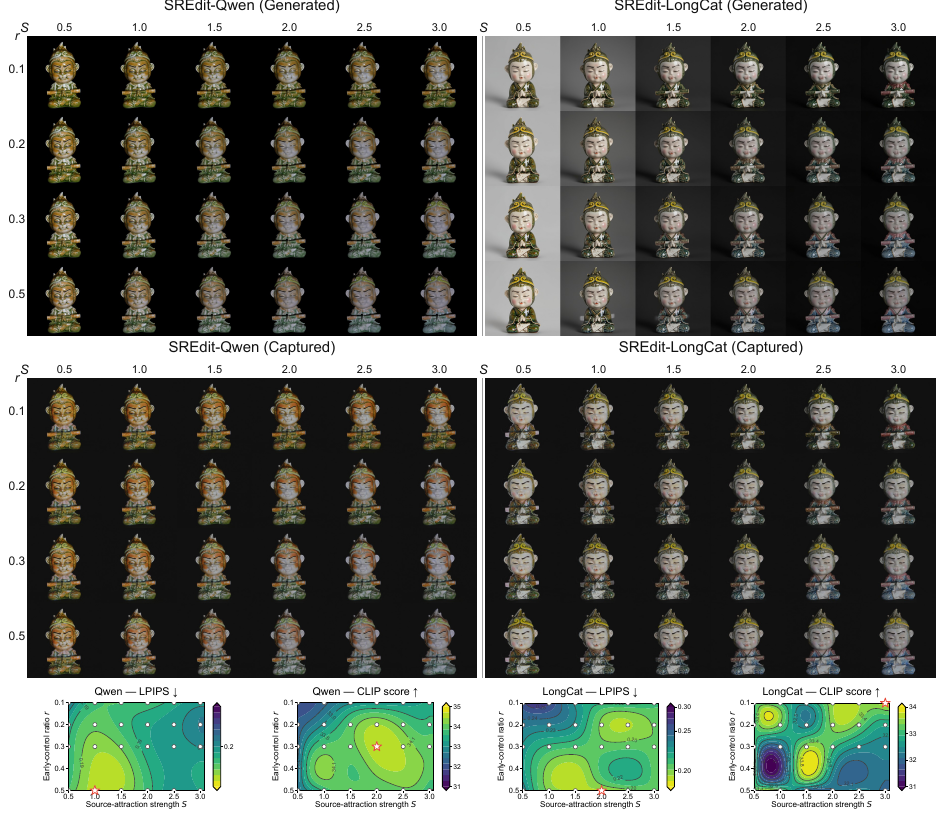}
    \caption{\textbf{Parameter ablation for Case~9.} Columns vary
    \(S\), while rows vary \(r\). The upper and lower matrices show generated
    and camera-captured results, respectively, and the bottom plots report
    LPIPS and CLIP over the same parameter grid.}
    \label{fig:case9-parameter-ablation}
\end{figure*}%
\clearpage

\begin{figure*}[p]
    \centering
    \setlength{\abovecaptionskip}{1pt}
    \setlength{\belowcaptionskip}{1pt}
    \includegraphics[width=0.95\textwidth,trim=1pt 1pt 2pt 6pt,clip]{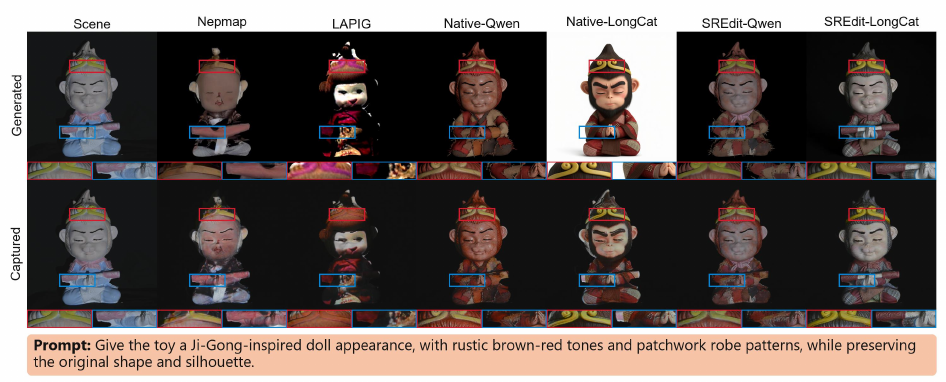}
    \caption{\textbf{Qualitative comparison for Case~10.} SREdit uses
    \((r,S)=(0.2,2.5)\) for Qwen and \((0.2,3.0)\) for LongCat. The upper
    and lower rows show generated appearances and camera-captured projection
    results, respectively.}
    \label{fig:case10-qualitative}
\end{figure*}

\clearpage

\clearpage
\onecolumn

\section{Prompts for representative comparisons}

Table~\ref{tab:complete-prompts} first lists the prompts used in the main paper,
namely the full prompts for the abbreviated labels in \cref{fig:teaser} and the
prompts for the two comparisons in \cref{fig:main-comparison}, followed by those
used for Cases~1--10 in the supplementary results. The reference-guided results
in \cref{fig:reference-material-editing} use the same prompts as
\cref{fig:teaser}(c), where Picture~1 is the source appearance and Picture~2 is
the reference image.
The wording, capitalization, and punctuation are reported exactly as used for
each method.

{\normalsize
\renewcommand{\arraystretch}{1.0}
\setlength{\tabcolsep}{5pt}

\begin{longtable}{@{}p{0.13\textwidth}p{0.08\textwidth}p{0.74\textwidth}@{}}
\caption{\textbf{Editing prompts.} Prompts used in \cref{fig:teaser} and in the
representative main-paper and supplementary comparisons.}\label{tab:complete-prompts}\\
\toprule
Comparison & Methods & Prompt \\
\midrule
\endfirsthead
\multicolumn{3}{l}{\textit{Table~\thetable{} continued from the previous page.}}\\
\toprule
Comparison & Methods & Prompt \\
\midrule
\endhead
\midrule
\multicolumn{3}{r}{\textit{Continued on the next page.}}\\
\endfoot
\bottomrule
\endlastfoot

\multicolumn{3}{@{}l}{\textbf{Prompts used in the main paper}}\\
\addlinespace[2pt]
\multicolumn{3}{@{}l}{\textit{\cref{fig:teaser}(b), text-guided editing}}\\
Blue Porcelain & SREdit & Give the toy a blue-and-white porcelain-style doll appearance, with an antique ceramic glaze and classic cobalt-blue patterns, while preserving the original shape and silhouette. \\
Ink Wash & SREdit & Give the toy an ink-wash-style doll appearance, with pale monochrome tones and flowing ink gradients, while preserving the original shape and silhouette. \\
Painted Wood & SREdit & Give the toy a painted-wood doll appearance, with warm colors and bold hand-painted floral patterns, while preserving the original shape and silhouette. \\
Tropical Style & SREdit & Give the toy a tropical outfit-style doll appearance, with vibrant island colors and bold floral patterns, while preserving the original shape and silhouette. \\
Forest Elf & SREdit & Give the toy a forest-elf-style doll appearance, with earthy green tones and softly glowing leaf patterns, while preserving the original shape and silhouette. \\
Halloween Style & SREdit & Give the toy a Halloween outfit-style doll appearance, with orange-black tones and spooky decorative patterns, while preserving the original shape and silhouette. \\
\addlinespace[4pt]
\multicolumn{3}{@{}l}{\textit{\cref{fig:teaser}(c) and \cref{fig:reference-material-editing}, reference-guided editing}}\\
Makeup & SREdit & Apply the facial makeup from Picture 2 onto the doll in Picture 1. \\
Material & SREdit & Change the entire body of the doll in Picture 1 to the texture and material from Picture 2. \\
Clothing & SREdit & Change the clothing material and texture of the doll in Picture 1 to the material and texture from Picture 2. \\
\addlinespace[4pt]
Main comparison (top) & SREdit & Give the toy an oil-painted-statue-style doll appearance, with rich layered colors and visible painterly brush textures, while preserving the original shape and silhouette. \\
 & Nepmap & an oil-painted-statue-style doll with rich layered colors and visible painterly brush textures \\
 & LAPIG & give the toy an oil-painted-statue-style doll appearance, with rich layered colors and visible painterly brush textures, while preserving the original shape and silhouette \\
\addlinespace[4pt]
Main comparison (bottom) & SREdit & Give the toy a painted-wood doll appearance, with warm colors and bold hand-painted floral patterns, while preserving the original shape and silhouette. \\
 & Nepmap & a painted-wood doll with warm colors and bold hand-painted floral patterns \\
 & LAPIG & give the toy a painted-wood doll appearance, with warm colors and bold hand-painted floral patterns, while preserving the original shape and silhouette \\

\midrule
\multicolumn{3}{@{}l}{\textbf{Prompts used in the supplementary results}}\\
\addlinespace[2pt]
Case 1 & SREdit & Transform the toy's entire surface into worn and aged newspaper material, with faded printed text, yellowed paper, wrinkles, torn edges, and visible creases. \\
 & Nepmap & a toy covered in worn old newspaper, with faded printed text, yellowed paper, wrinkles, creases, and torn edges \\
 & LAPIG & Turn the toy into worn old newspaper material with faded printed text, yellowed paper, wrinkles, creases, and torn edges. \\
\addlinespace[4pt]
Case 2 & SREdit & Transform the toy's entire surface into soft knitted yarn, with visible stitches, woven patterns, and a fluffy wool texture. \\
 & Nepmap & a toy made of soft knitted yarn, with visible stitches, woven patterns, and a fluffy wool texture \\
 & LAPIG & Turn the toy into soft knitted yarn with visible stitches, woven patterns, and a fluffy wool texture. \\
\addlinespace[4pt]
Case 3 & SREdit & apply an aged bronze finish with subtle turquoise patina to the toy's existing surface \\
 & Nepmap & a toy with an aged bronze surface and subtle turquoise patina \\
 & LAPIG & change the toy surface into an aged bronze finish with subtle turquoise patina \\
\addlinespace[4pt]
Case 4 & SREdit & Transform the toy's entire surface into aged marble overgrown with moss, with visible stone veining, weathered cracks, damp stains, soft green moss patches, and a cold sculptural finish. \\
 & Nepmap & a toy with an aged marble surface overgrown with moss, visible stone veining, weathered cracks, damp stains, soft green moss patches, and a cold sculptural finish \\
 & LAPIG & turn the toy surface into aged marble overgrown with moss, with visible stone veining, weathered cracks, damp stains, soft green moss patches, and a cold sculptural finish \\
\addlinespace[4pt]
Case 5 & SREdit & Give the toy a coral-reef ocean-style doll appearance, with vivid ocean colors and organic coral patterns, including subtle seashell textures on the face, while preserving the original shape and silhouette. \\
 & Nepmap & a coral-reef ocean-style doll with vivid ocean colors and organic coral patterns, including subtle seashell textures on the face \\
 & LAPIG & give the toy a coral-reef ocean-style doll appearance, with vivid ocean colors and organic coral patterns, including subtle seashell textures on the face, while preserving the original shape and silhouette \\
\addlinespace[4pt]
Case 6 & SREdit & Give the toy a jungle-style doll appearance, with lush green tones and organic jungle patterns, including vine and branch textures on the face, and wood, leaf, and vine textures across the surface, while preserving the original shape and silhouette. \\
 & Nepmap & a jungle-style doll with lush green tones and organic jungle patterns, including vine and branch textures on the face, and wood, leaf, and vine textures across the surface \\
 & LAPIG & give the toy a jungle-style doll appearance, with lush green tones and organic jungle patterns, including vine and branch textures on the face, and wood, leaf, and vine textures across the surface, while preserving the original shape and silhouette \\
\addlinespace[4pt]
Case 7 & SREdit & Give the toy a blue-and-white porcelain-style doll appearance, with an antique ceramic glaze and classic cobalt-blue patterns, while preserving the original shape and silhouette. \\
 & Nepmap & a blue-and-white porcelain-style doll with an antique ceramic glaze and classic cobalt-blue patterns \\
 & LAPIG & give the toy a blue-and-white porcelain-style doll appearance, with an antique ceramic glaze and classic cobalt-blue patterns, while preserving the original shape and silhouette \\
\addlinespace[4pt]
Case 8 & SREdit & Give the toy a Halloween outfit-style doll appearance, with orange-black tones and spooky decorative patterns, while preserving the original shape and silhouette. \\
 & Nepmap & a Halloween outfit-style doll with orange-black tones and spooky decorative patterns \\
 & LAPIG & give the toy a Halloween outfit-style doll appearance, with orange-black tones and spooky decorative patterns, while preserving the original shape and silhouette \\
\addlinespace[4pt]
Case 9 & SREdit & Give the toy a Tang-sancai ceramic-figurine-style doll appearance, with amber-green-white glaze colors and flowing ceramic textures, while preserving the original shape and silhouette. \\
 & Nepmap & a Tang-sancai ceramic-figurine-style doll with amber-green-white glaze colors and flowing ceramic textures \\
 & LAPIG & give the toy a Tang-sancai ceramic-figurine-style doll appearance, with amber-green-white glaze colors and flowing ceramic textures, while preserving the original shape and silhouette \\
\addlinespace[4pt]
Case 10 & SREdit & Give the toy a Ji-Gong-inspired doll appearance, with rustic brown-red tones and patchwork robe patterns, while preserving the original shape and silhouette. \\
 & Nepmap & a Ji-Gong-inspired doll with rustic brown-red tones and patchwork robe patterns \\
 & LAPIG & give the toy a Ji-Gong-inspired doll appearance, with rustic brown-red tones and patchwork robe patterns, while preserving the original shape and silhouette \\
\end{longtable}

}

\section{User study questionnaire}

\subsection{Instructions}

Participants were told that each comparison concerned an edited appearance
physically projected onto a real object and photographed by a camera. For each
trial, they were asked to consider the text instruction, the source object,
and the two captured results as a whole. They were also informed that the two
methods were anonymous, that the A/B positions carried no meaning, and that
there were no correct answers.

\subsection{Interface and questions}

Each trial showed the source object, the text instruction, and two anonymous
projection results labeled A and B. The assignment of
the SREdit and corresponding Native versions to A or B was randomized and recorded for each trial. Participants
answered the following five questions:

\begin{enumerate}
  \item \textbf{Text alignment:} which result better matched the requested
  appearance?
  \item \textbf{Artifacts:} which result showed fewer visible artifacts, such
  as abnormal color leakage or bleed-through inside the object region?
  \item \textbf{Lighting and surface realism:} which result showed more
  natural illumination and surface appearance, with fewer abnormal highlights
  or discontinuous shading?
  \item \textbf{Visible misalignment:} did each result show visible misalignment,
  such as texture crossing the object boundary or object parts appearing at
  implausible locations?
  \item \textbf{Overall preference:} which result was preferable when all
  of the above aspects were considered together?
\end{enumerate}

Text alignment, artifacts, realism, and
overall preference used a four-level
forced-choice scale: ``A is much better,'' ``A is slightly better,'' ``B is
slightly better,'' and ``B is much better.'' Visible misalignment was judged
separately for A and B as either present or absent.

Comparative responses were converted to the method shown on each side before
aggregation, so the reported preference rate always denotes the proportion of
judgments favoring SREdit. The 36 formal trials are reproduced below in their
presentation order; practice trials are not included in the analysis.

\newcommand{\sreditcaptionof}[2]{%
  \refstepcounter{figure}%
  \par\vspace{-1pt}%
  {\footnotesize Figure~\thefigure: #2\par}%
}

\clearpage
\twocolumn

\subsection{All main trials}

The 36 main comparisons are shown in their presentation order. Each panel
contains the source object and the two anonymous results A and B, with the A/B
order preserved from the study. The corresponding text instruction is given in
the caption.

\begin{center}
\includegraphics[width=\columnwidth]{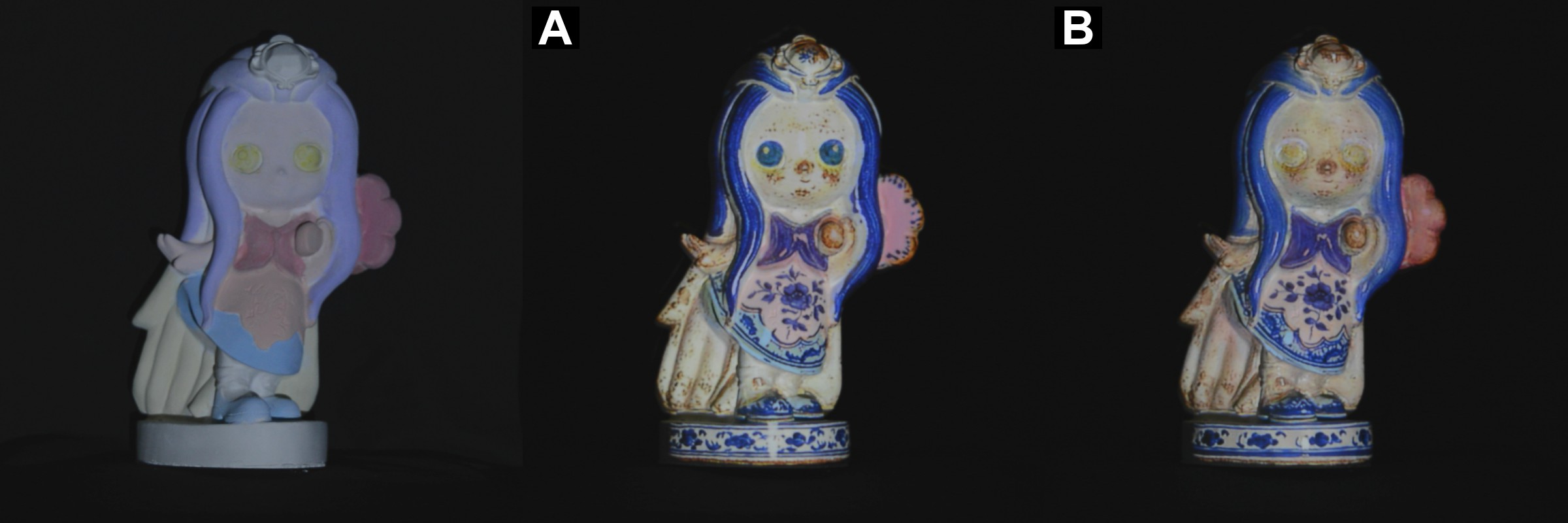}
\sreditcaptionof{figure}{Trial 1 (Qwen): ``Give the toy a blue-and-white porcelain-style doll appearance, with an antique ceramic glaze and classic cobalt-blue patterns, while preserving the original shape and silhouette.''}
\end{center}\vspace{2pt}

\begin{center}
\includegraphics[width=\columnwidth]{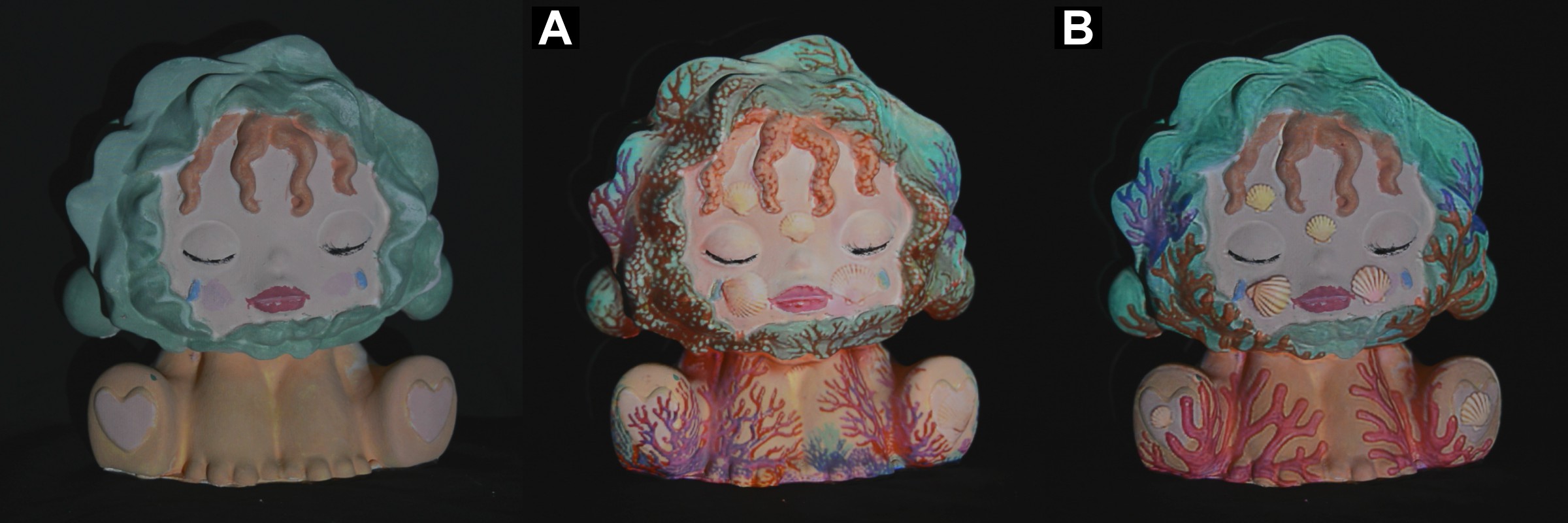}
\sreditcaptionof{figure}{Trial 2 (Qwen): ``Give the toy a coral-reef ocean-style doll appearance, with vivid ocean colors and organic coral patterns, including subtle seashell textures on the face, while preserving the original shape and silhouette.''}
\end{center}\vspace{2pt}

\begin{center}
\includegraphics[width=\columnwidth]{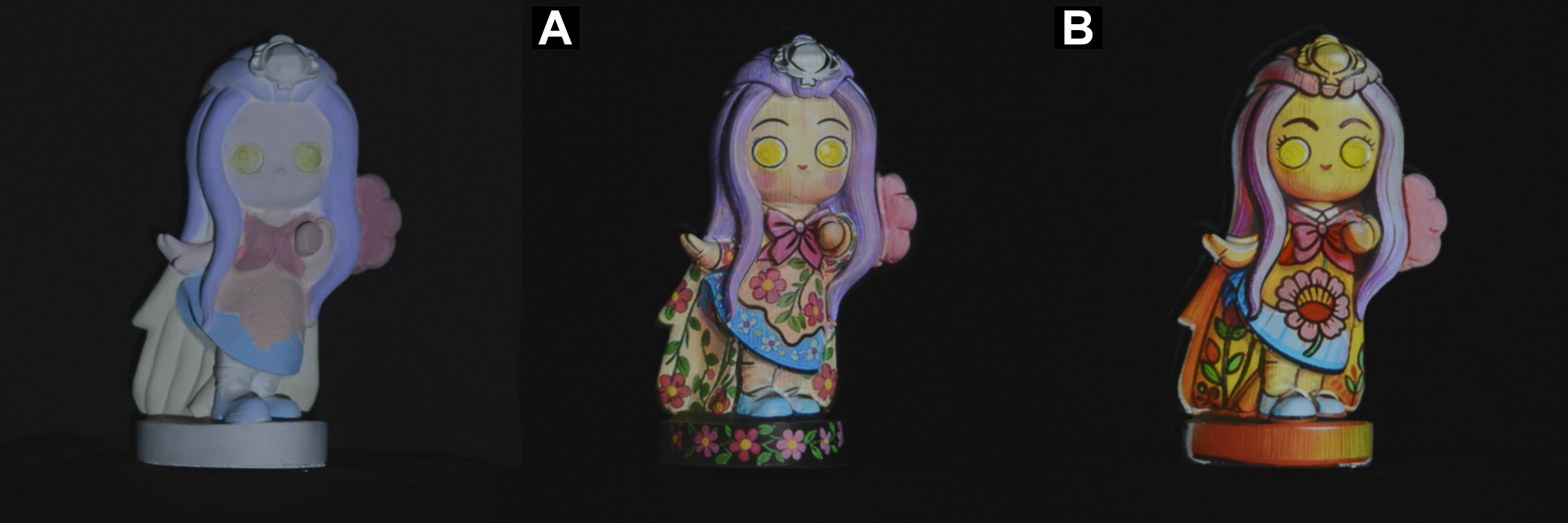}
\sreditcaptionof{figure}{Trial 3 (LongCat): ``Give the toy a painted-wood doll appearance, with warm colors and bold hand-painted floral patterns, while preserving the original shape and silhouette.''}
\end{center}\vspace{2pt}

\begin{center}
\includegraphics[width=\columnwidth]{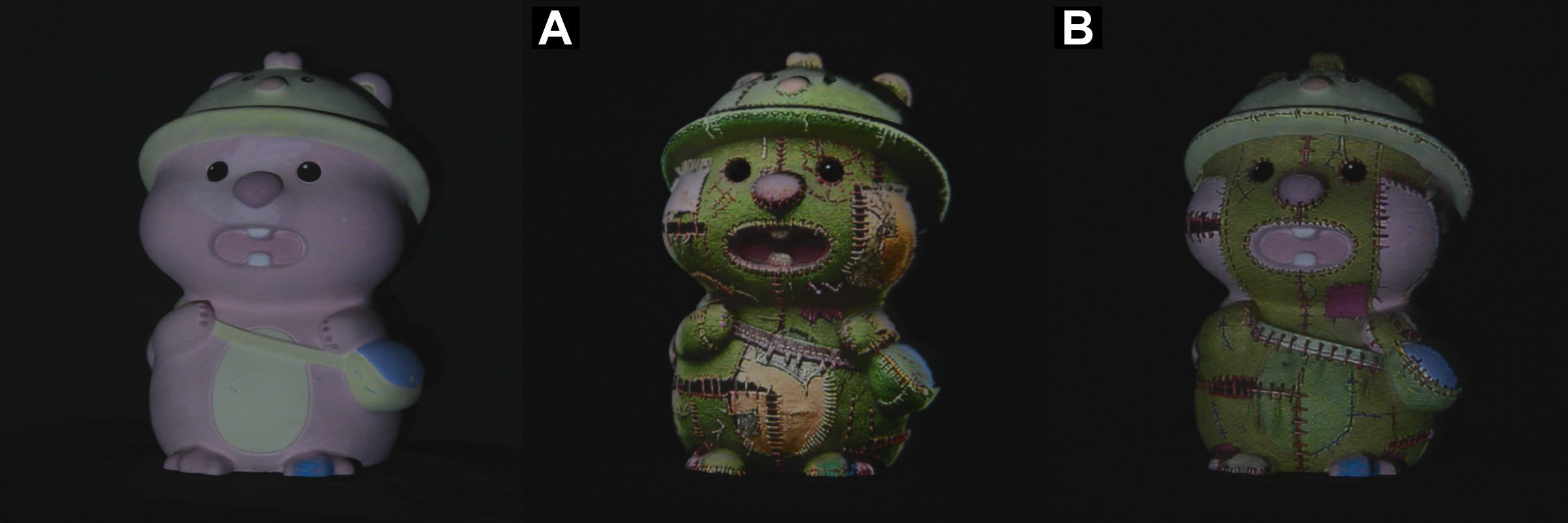}
\sreditcaptionof{figure}{Trial 4 (Qwen): ``Transform the toy’s entire surface into a Frankenstein-style stitched patchwork material, with mismatched green fabric pieces, visible seams, rough stitches, uneven patches, and a handmade monster-like look.''}
\end{center}\vspace{2pt}

\begin{center}
\includegraphics[width=\columnwidth]{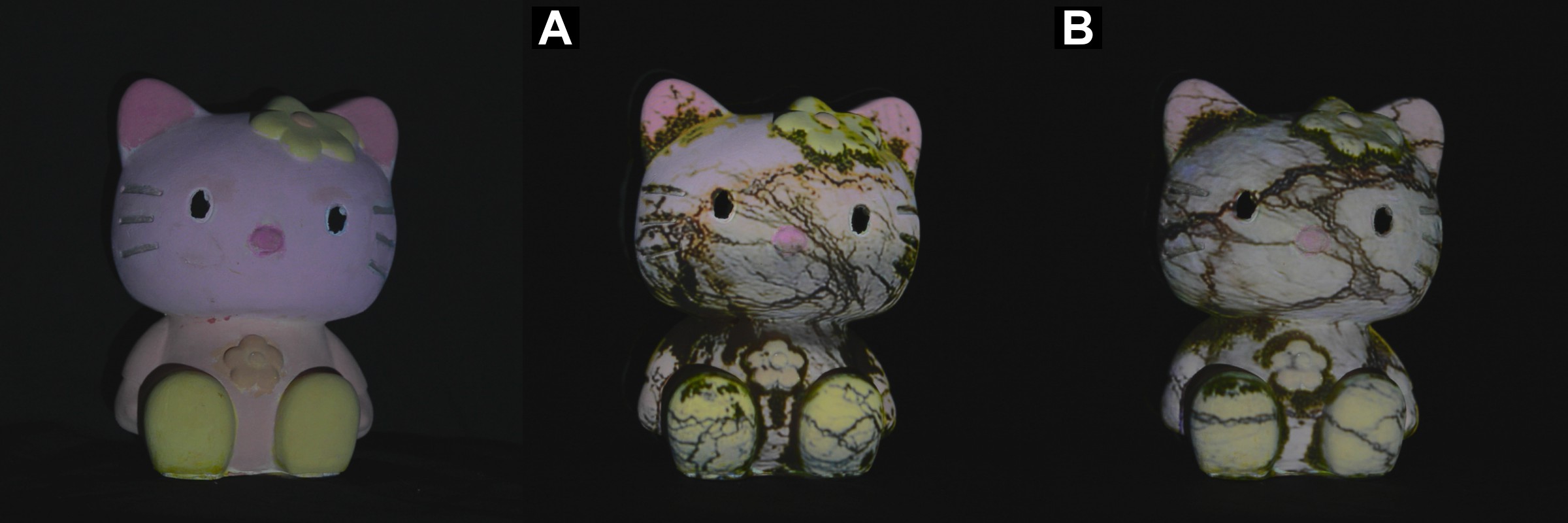}
\sreditcaptionof{figure}{Trial 5 (Qwen): ``Transform the toy’s entire surface into aged marble overgrown with moss, with visible stone veining, weathered cracks, damp stains, soft green moss patches, and a cold sculptural finish.''}
\end{center}\vspace{2pt}

\begin{center}
\includegraphics[width=\columnwidth]{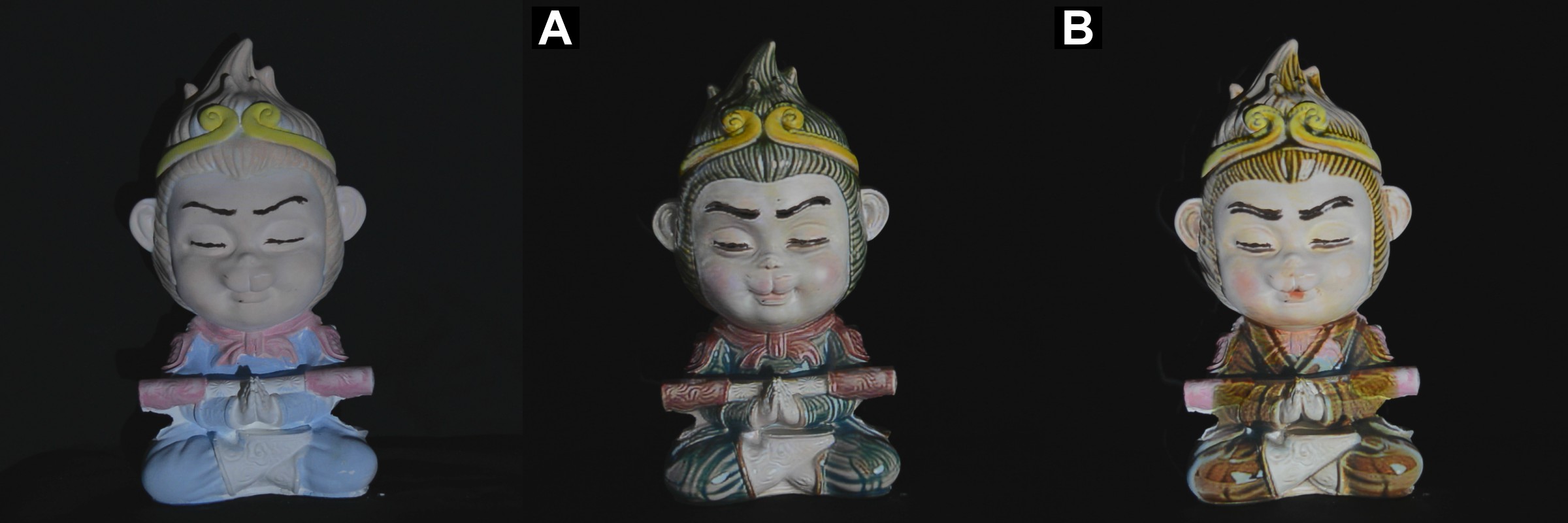}
\sreditcaptionof{figure}{Trial 6 (LongCat): ``Give the toy a Tang-sancai ceramic-figurine-style doll appearance, with amber-green-white glaze colors and flowing ceramic textures, while preserving the original shape and silhouette.''}
\end{center}\vspace{2pt}

\begin{center}
\includegraphics[width=\columnwidth]{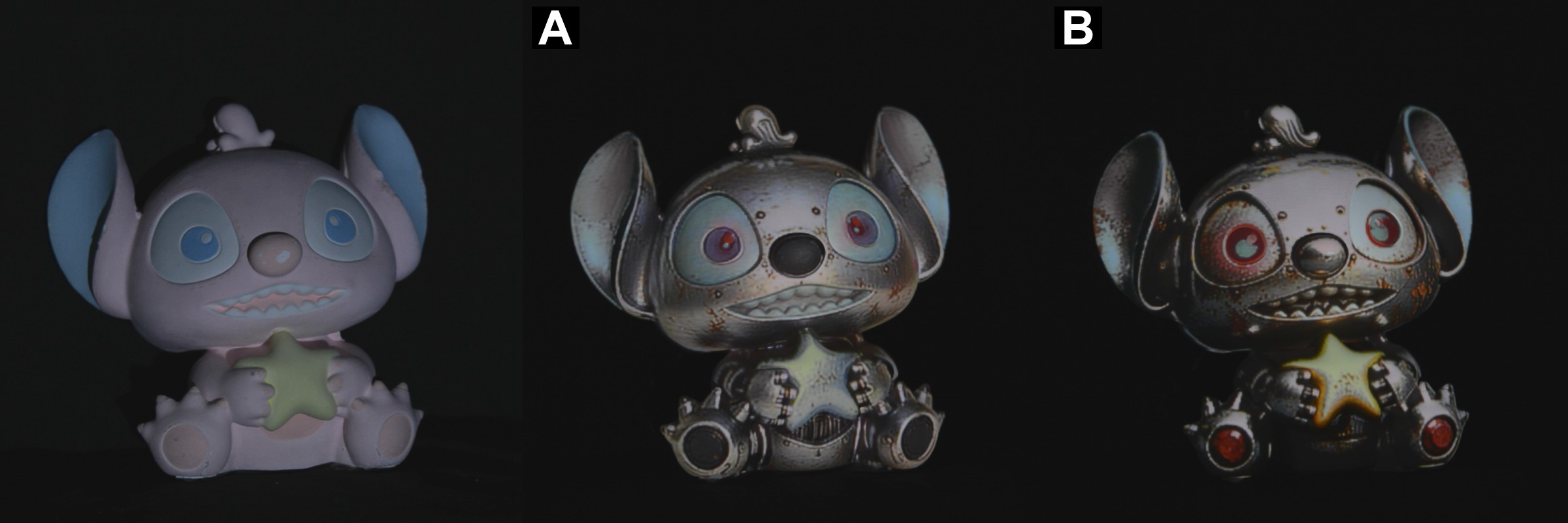}
\sreditcaptionof{figure}{Trial 7 (LongCat): ``Transform the toy’s entire surface into a battle-worn metallic robot material, with exposed chrome, dark steel plates, mechanical joints, and glowing red cybernetic eyes.''}
\end{center}\vspace{2pt}

\begin{center}
\includegraphics[width=\columnwidth]{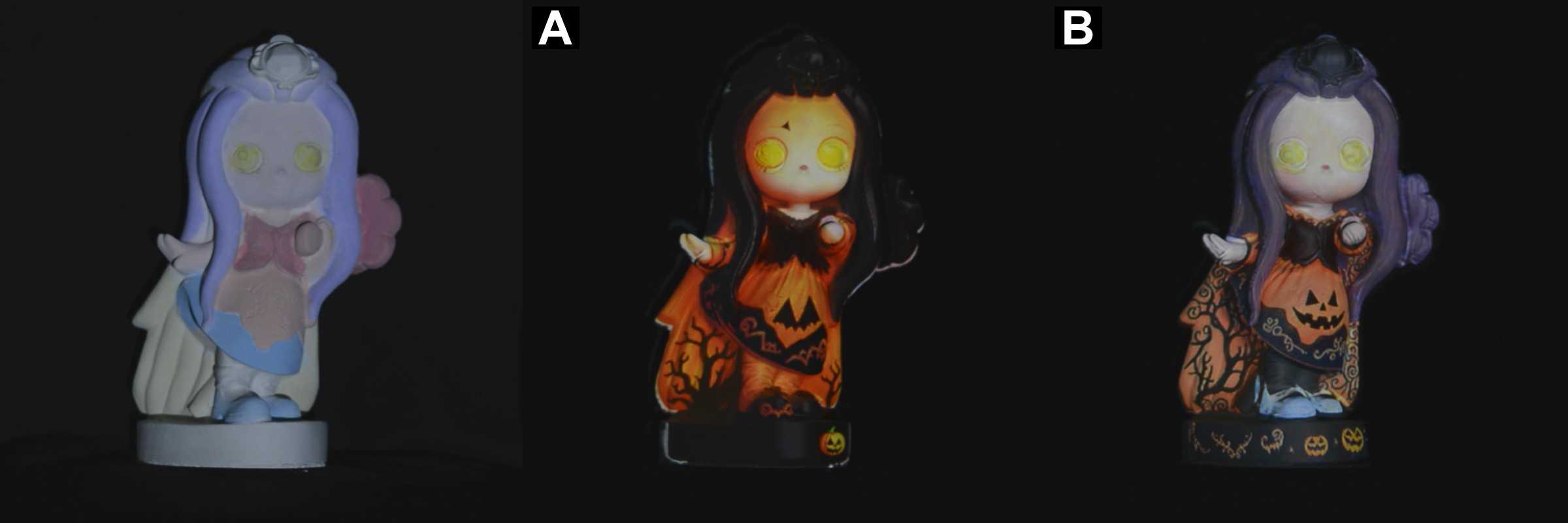}
\sreditcaptionof{figure}{Trial 8 (LongCat): ``Give the toy a Halloween outfit-style doll appearance, with orange-black tones and spooky decorative patterns, while preserving the original shape and silhouette.''}
\end{center}\vspace{2pt}

\begin{center}
\includegraphics[width=\columnwidth]{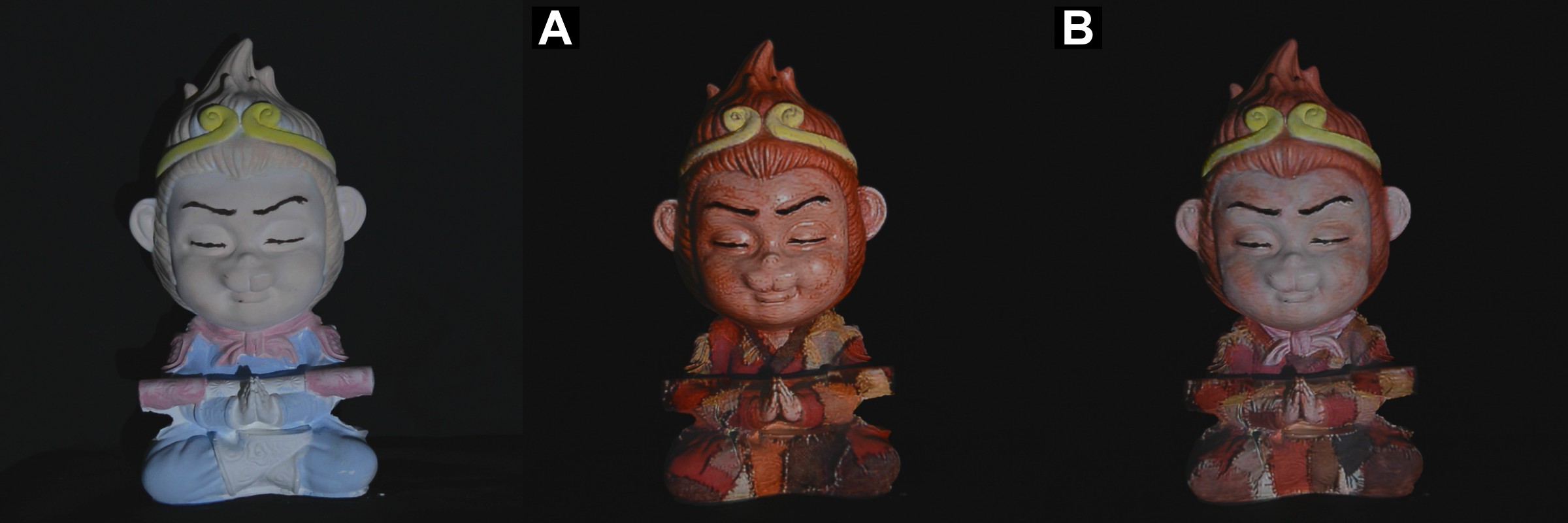}
\sreditcaptionof{figure}{Trial 9 (Qwen): ``Give the toy a Ji-Gong-inspired doll appearance, with rustic brown-red tones and patchwork robe patterns, while preserving the original shape and silhouette.''}
\end{center}\vspace{2pt}

\begin{center}
\includegraphics[width=\columnwidth]{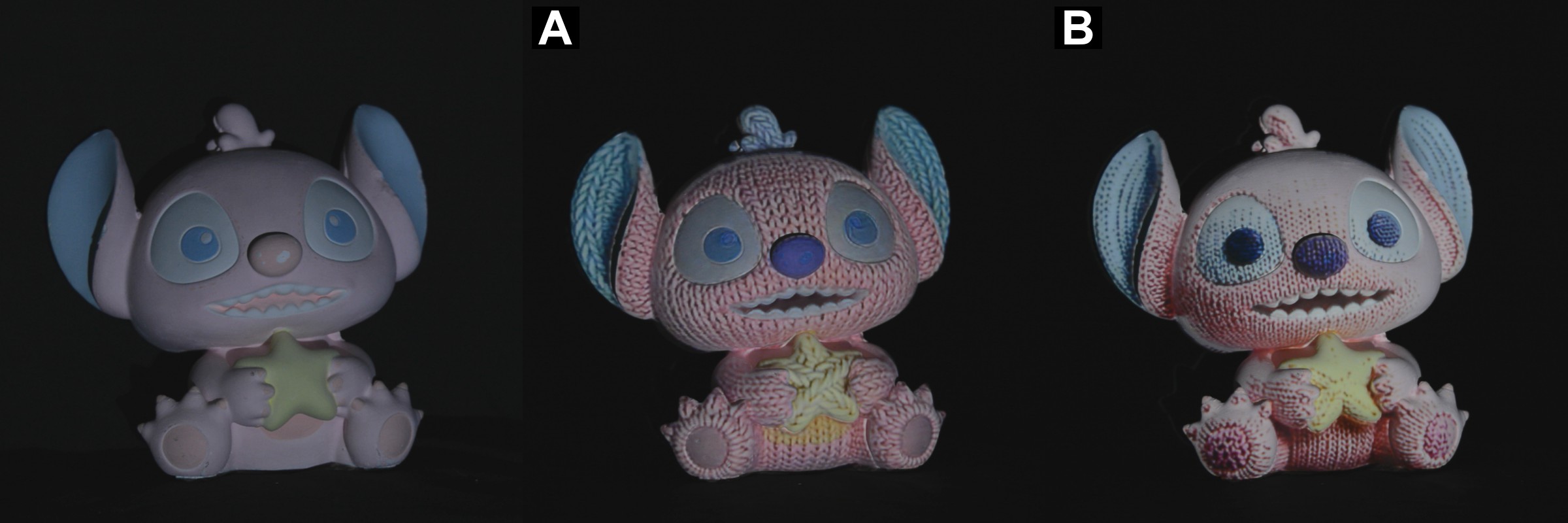}
\sreditcaptionof{figure}{Trial 10 (Qwen): ``Transform the toy’s entire surface into soft knitted yarn, with visible stitches, woven patterns, and a fluffy wool texture.''}
\end{center}\vspace{2pt}

\begin{center}
\includegraphics[width=\columnwidth]{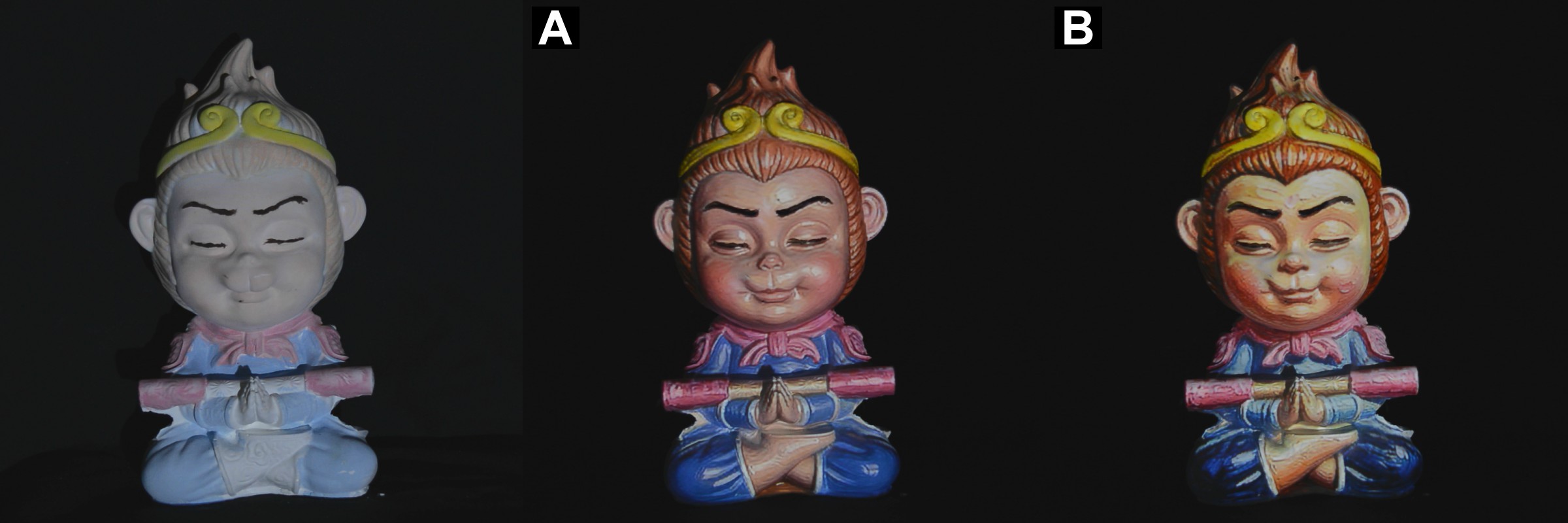}
\sreditcaptionof{figure}{Trial 11 (Qwen): ``Give the toy an oil-painted-statue-style doll appearance, with rich layered colors and visible painterly brush textures, while preserving the original shape and silhouette.''}
\end{center}\vspace{2pt}

\begin{center}
\includegraphics[width=\columnwidth]{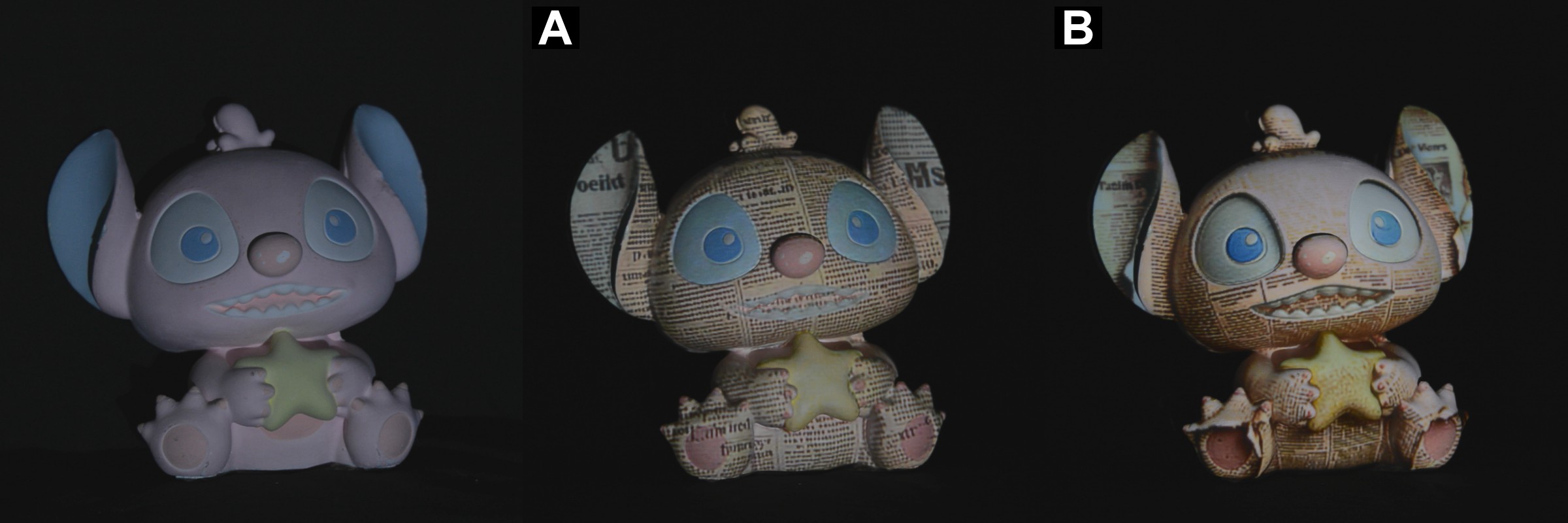}
\sreditcaptionof{figure}{Trial 12 (LongCat): ``Transform the toy’s entire surface into worn and aged newspaper material, with faded printed text, yellowed paper, wrinkles, torn edges, and visible creases.''}
\end{center}\vspace{2pt}

\begin{center}
\includegraphics[width=\columnwidth]{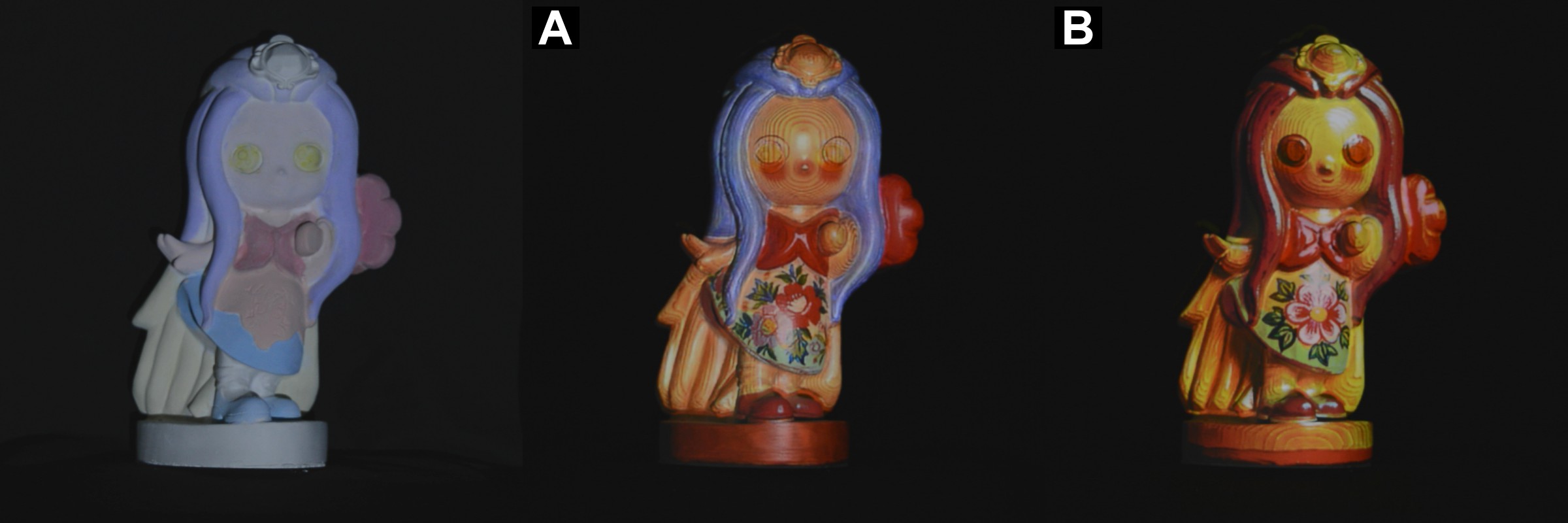}
\sreditcaptionof{figure}{Trial 13 (Qwen): ``Give the toy a painted-wood doll appearance, with warm colors and bold hand-painted floral patterns, while preserving the original shape and silhouette.''}
\end{center}\vspace{2pt}

\begin{center}
\includegraphics[width=\columnwidth]{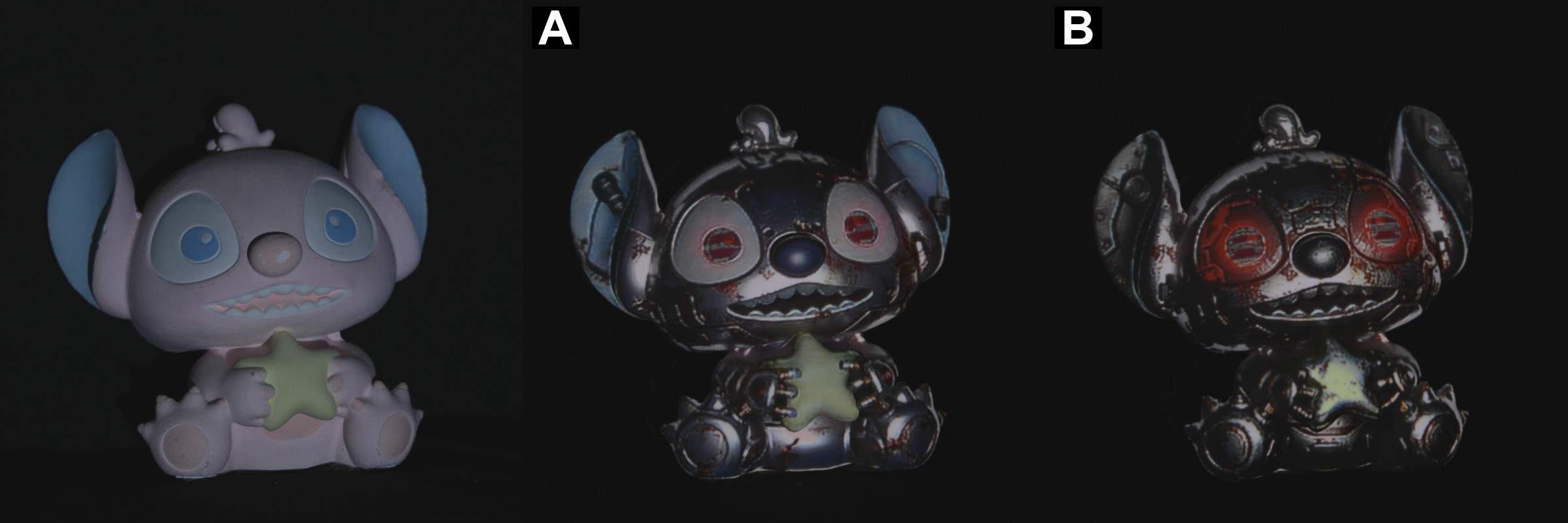}
\sreditcaptionof{figure}{Trial 14 (Qwen): ``Transform the toy’s entire surface into a battle-worn metallic robot material, with exposed chrome, dark steel plates, mechanical joints, and glowing red cybernetic eyes.''}
\end{center}\vspace{2pt}

\begin{center}
\includegraphics[width=\columnwidth]{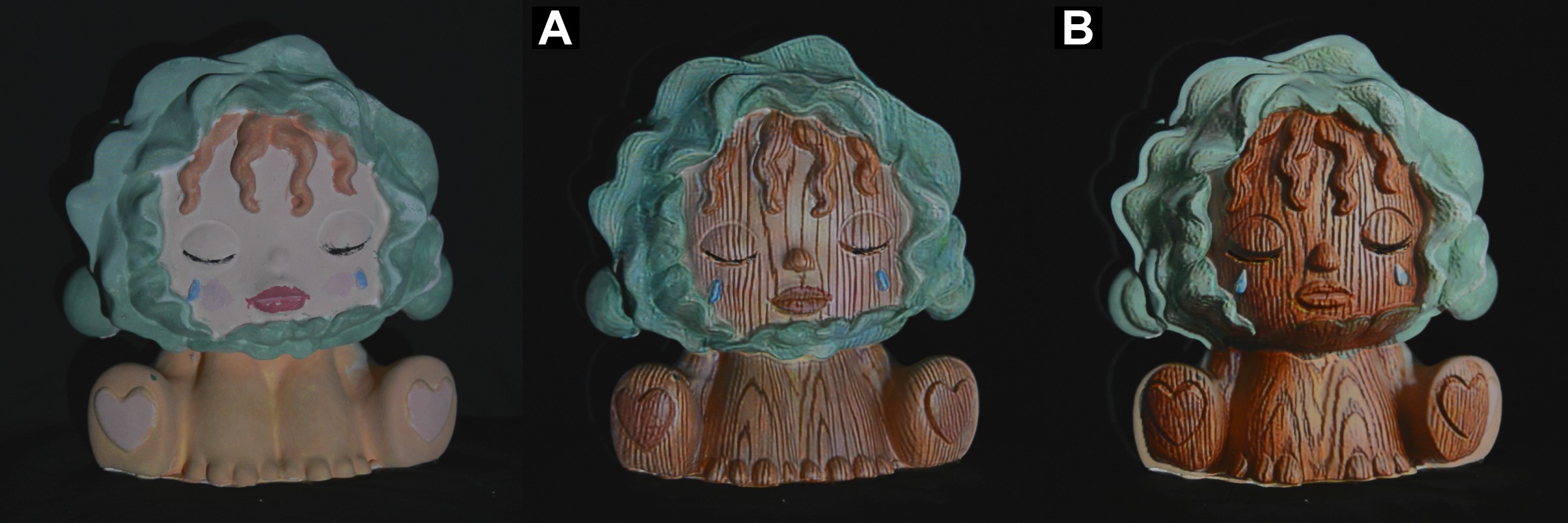}
\sreditcaptionof{figure}{Trial 15 (LongCat): ``Give the toy a woodcarving appearance, with carved wood grain, sculpted chisel marks, and natural wooden textures, while preserving the original shape and silhouette.''}
\end{center}\vspace{2pt}

\begin{center}
\includegraphics[width=\columnwidth]{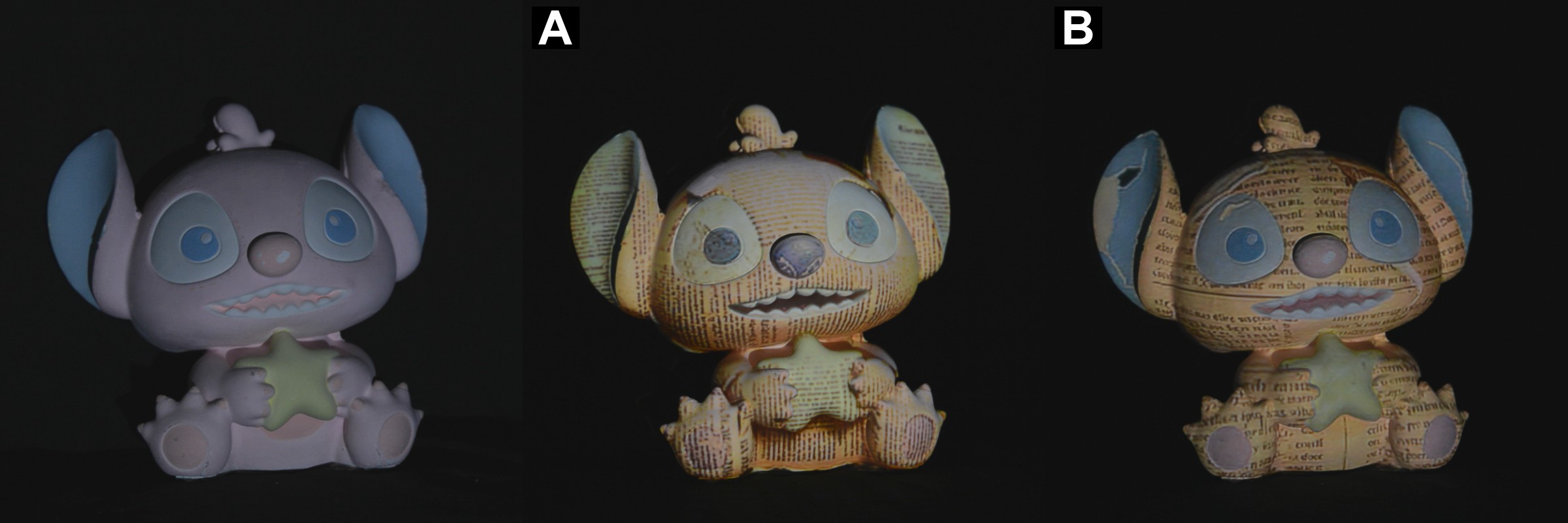}
\sreditcaptionof{figure}{Trial 16 (Qwen): ``Transform the toy’s entire surface into worn and aged newspaper material, with faded printed text, yellowed paper, wrinkles, torn edges, and visible creases.''}
\end{center}\vspace{2pt}

\begin{center}
\includegraphics[width=\columnwidth]{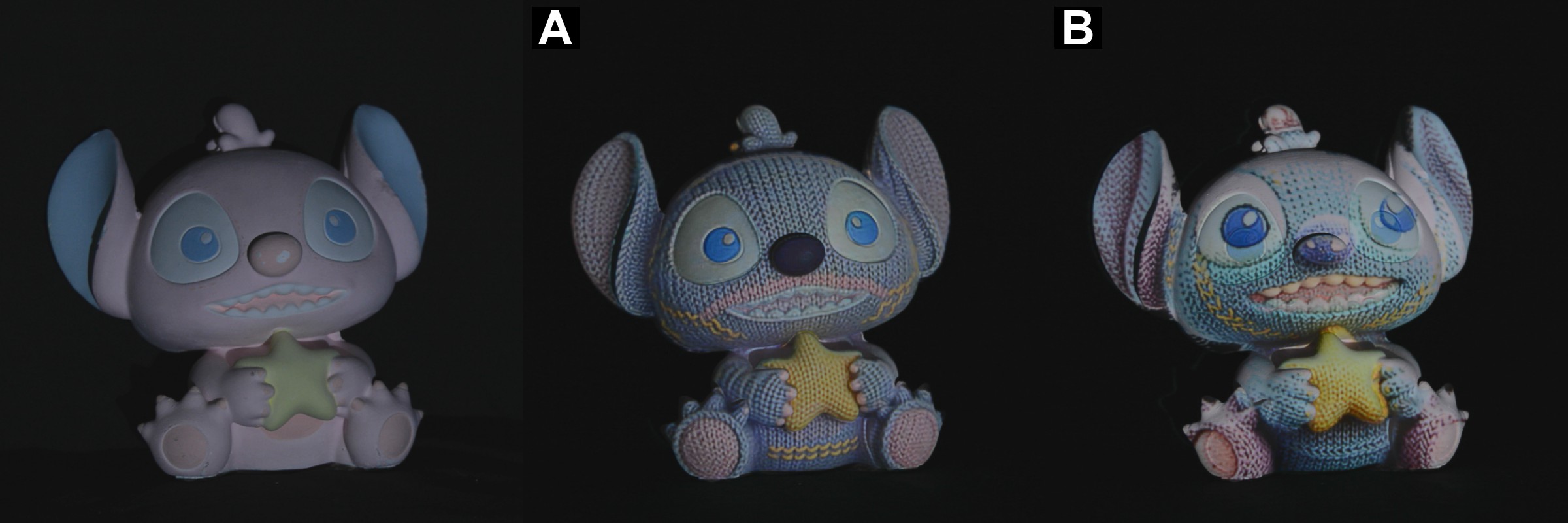}
\sreditcaptionof{figure}{Trial 17 (LongCat): ``Transform the toy’s entire surface into soft knitted yarn, with visible stitches, woven patterns, and a fluffy wool texture.''}
\end{center}\vspace{2pt}

\begin{center}
\includegraphics[width=\columnwidth]{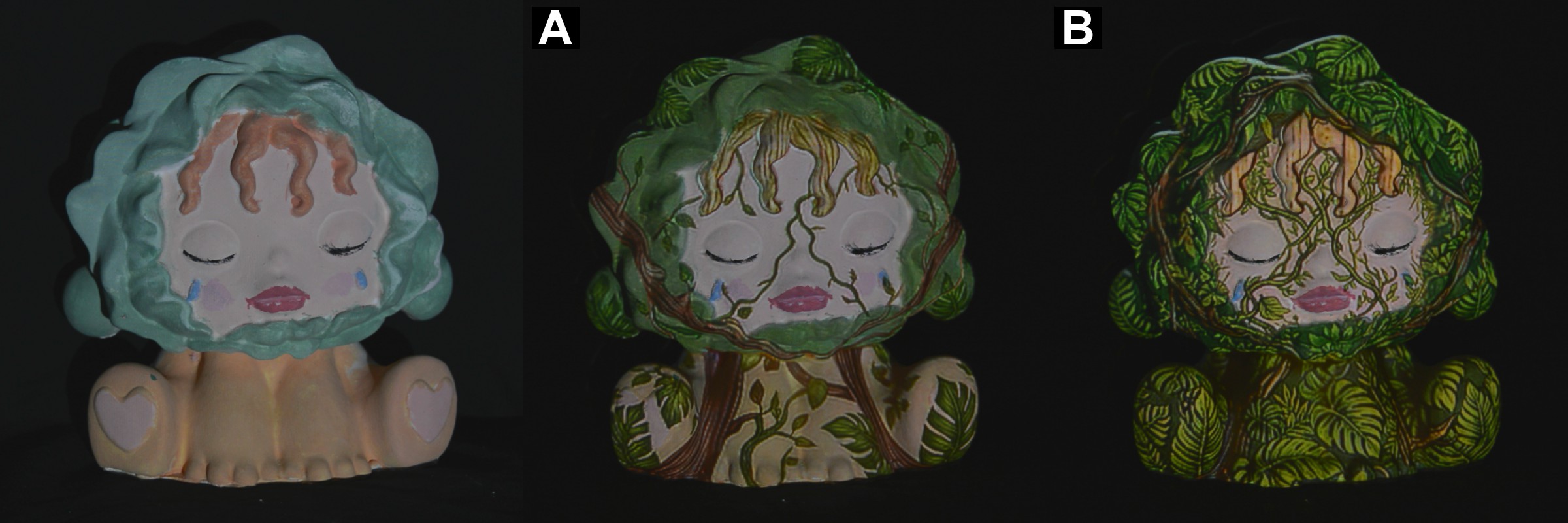}
\sreditcaptionof{figure}{Trial 18 (Qwen): ``Give the toy a jungle-style doll appearance, with lush green tones and organic jungle patterns, including vine and branch textures on the face, and wood, leaf, and vine textures across the surface, while preserving the original shape and silhouette.''}
\end{center}\vspace{2pt}

\begin{center}
\includegraphics[width=\columnwidth]{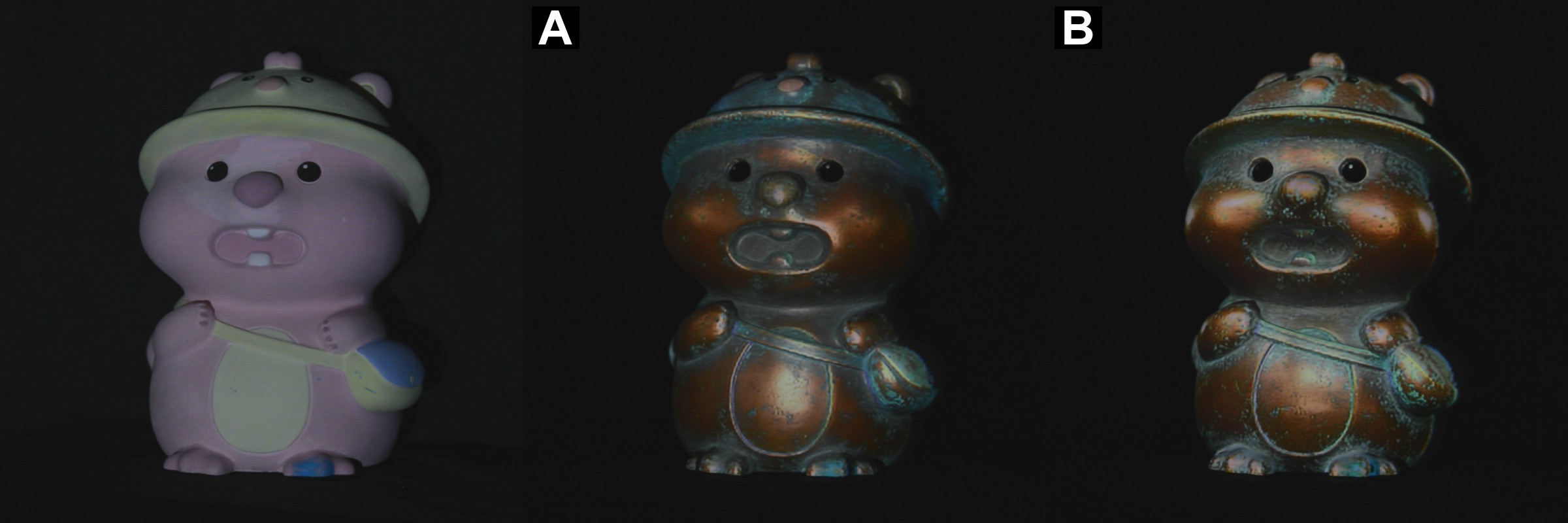}
\sreditcaptionof{figure}{Trial 19 (Qwen): ``apply an aged bronze finish with subtle turquoise patina to the toy's existing surface''}
\end{center}\vspace{2pt}

\begin{center}
\includegraphics[width=\columnwidth]{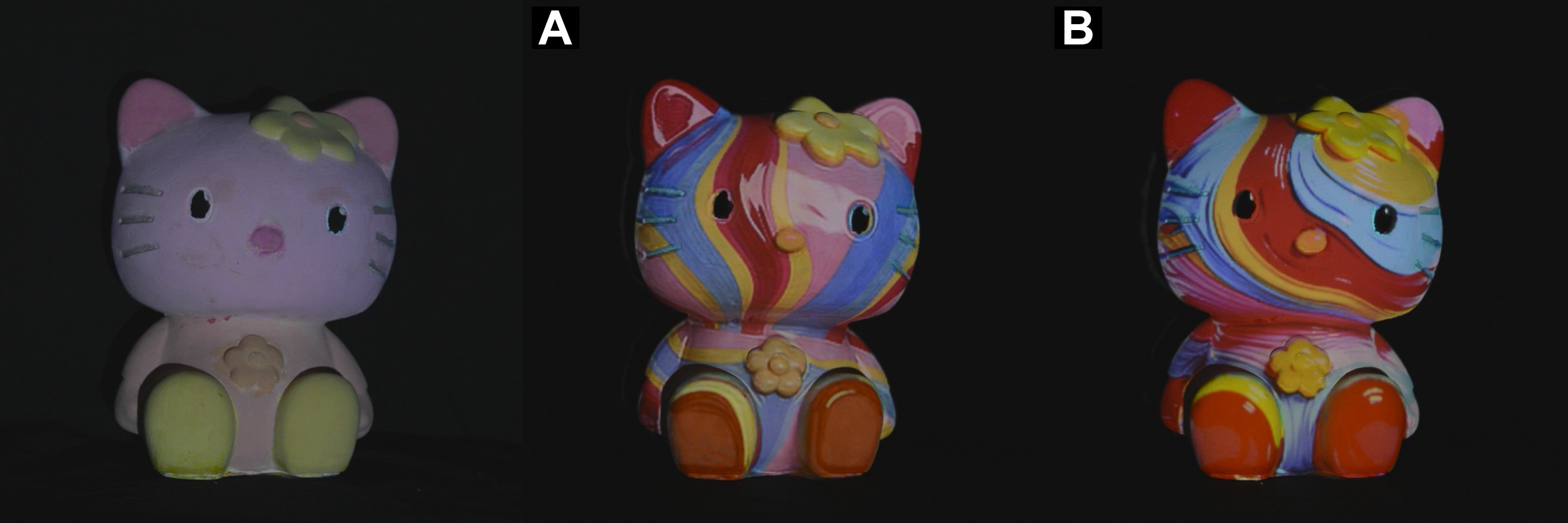}
\sreditcaptionof{figure}{Trial 20 (LongCat): ``Transform the toy’s entire surface into layered colored wax, with broad flowing bands of red, yellow, blue, and pink, glossy hardened textures, visible wax ridges, and rich reflective shading.''}
\end{center}\vspace{2pt}

\begin{center}
\includegraphics[width=\columnwidth]{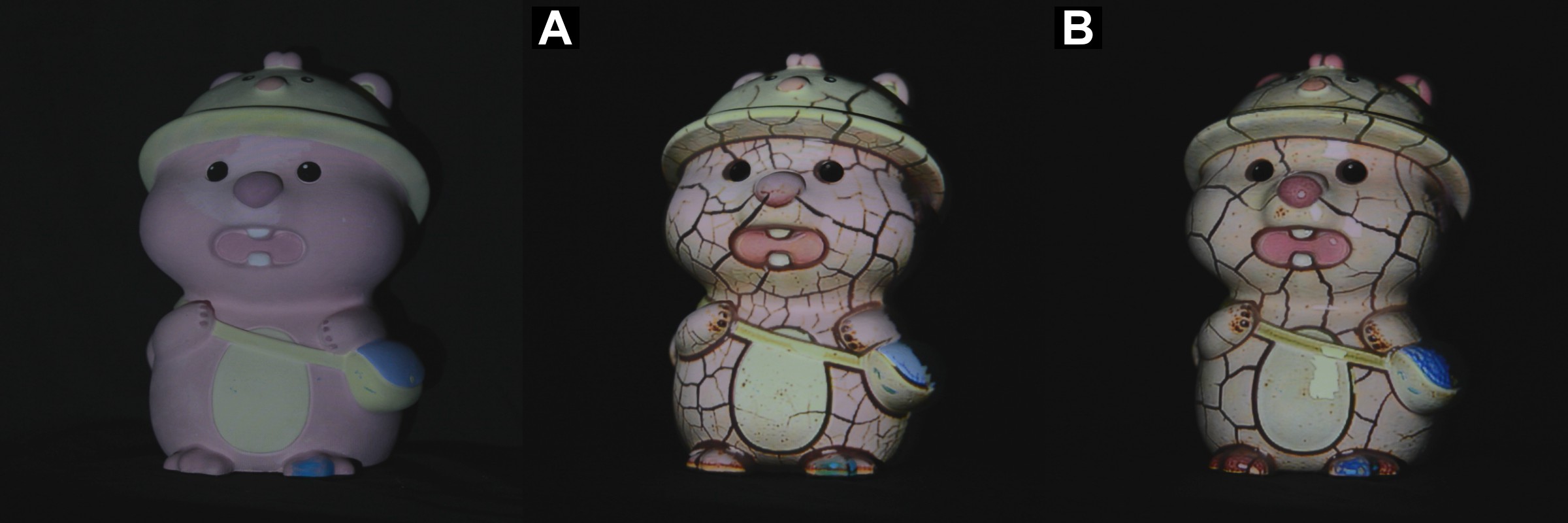}
\sreditcaptionof{figure}{Trial 21 (LongCat): ``Transform the toy’s entire surface into aged cracked ceramic, with an ivory-white glaze, clearly visible dark crackle lines, worn stains, chipped paint textures, and glossy ceramic highlights.''}
\end{center}\vspace{2pt}

\begin{center}
\includegraphics[width=\columnwidth]{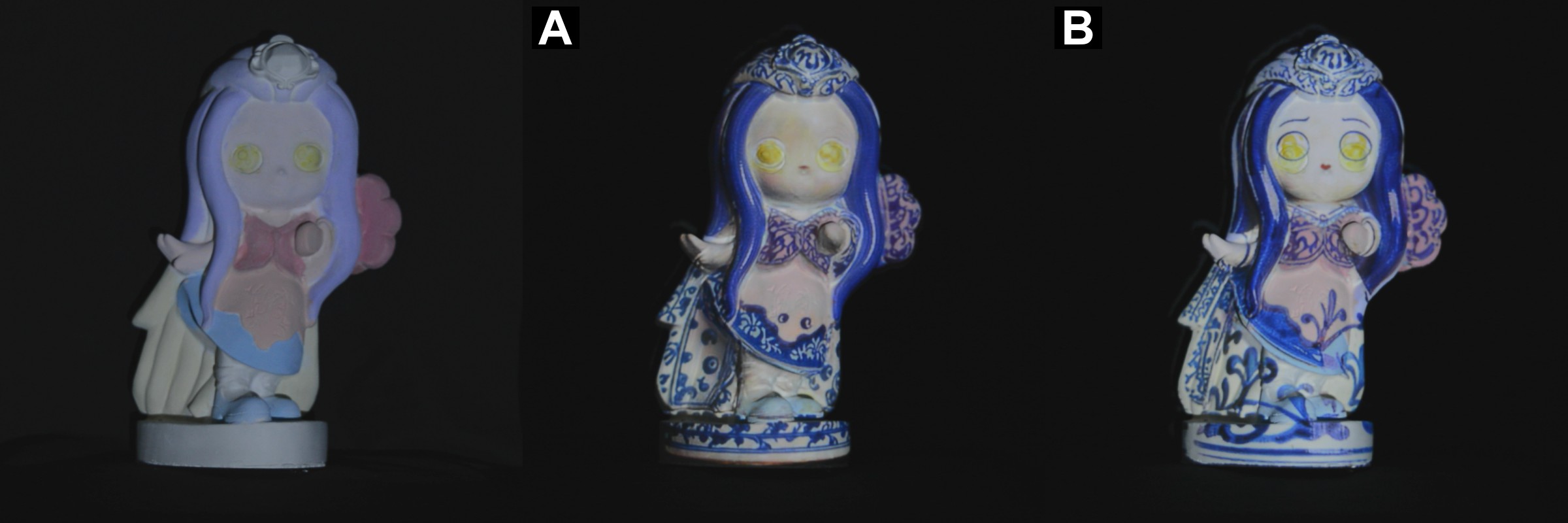}
\sreditcaptionof{figure}{Trial 22 (LongCat): ``Give the toy a blue-and-white porcelain-style doll appearance, with an antique ceramic glaze and classic cobalt-blue patterns, while preserving the original shape and silhouette.''}
\end{center}\vspace{2pt}

\begin{center}
\includegraphics[width=\columnwidth]{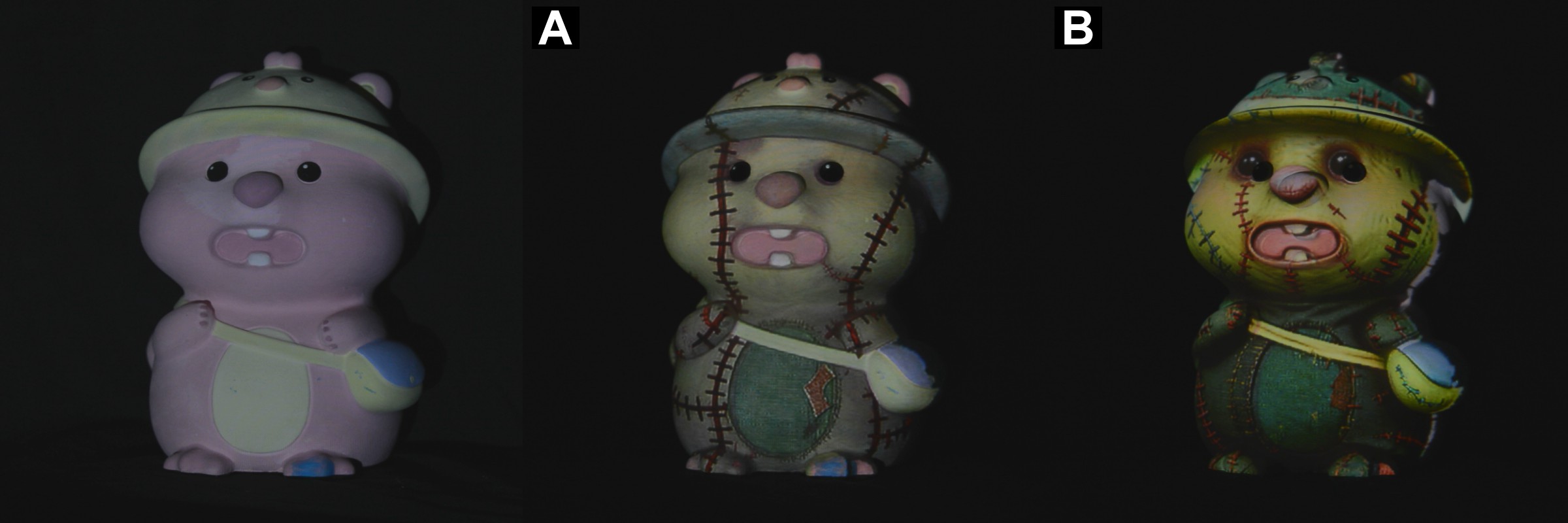}
\sreditcaptionof{figure}{Trial 23 (LongCat): ``Transform the toy’s entire surface into a Frankenstein-style stitched patchwork material, with mismatched green fabric pieces, visible seams, rough stitches, uneven patches, and a handmade monster-like look.''}
\end{center}\vspace{2pt}

\begin{center}
\includegraphics[width=\columnwidth]{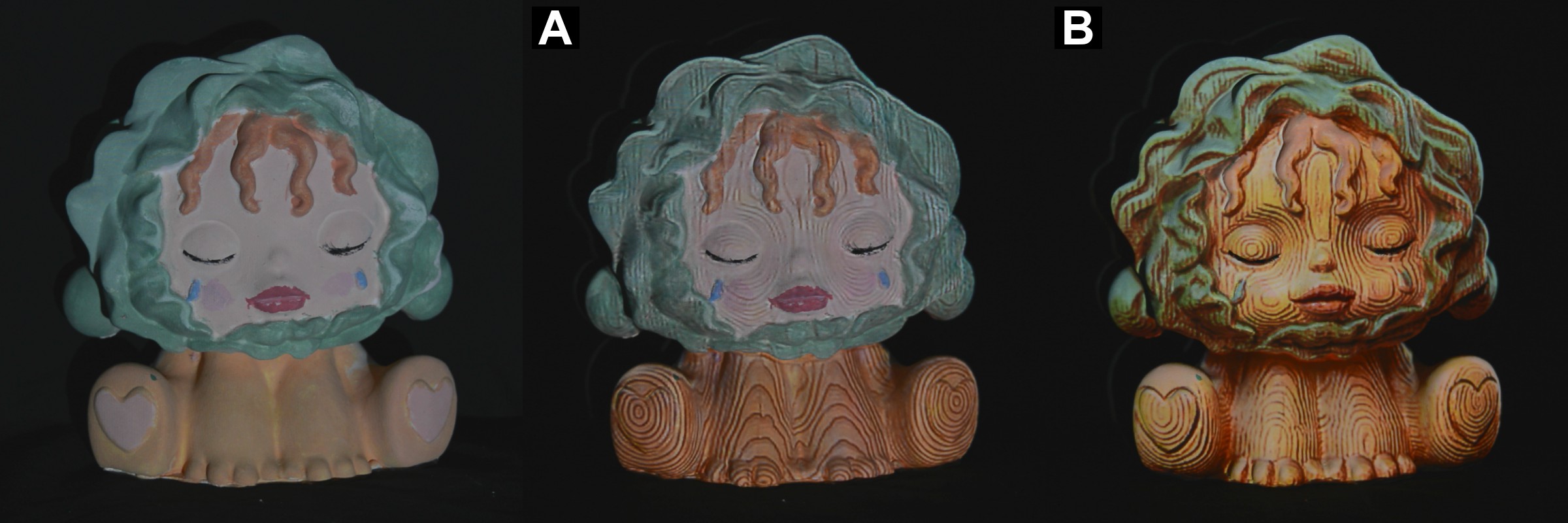}
\sreditcaptionof{figure}{Trial 24 (Qwen): ``Give the toy a woodcarving appearance, with carved wood grain, sculpted chisel marks, and natural wooden textures, while preserving the original shape and silhouette.''}
\end{center}\vspace{2pt}

\begin{center}
\includegraphics[width=\columnwidth]{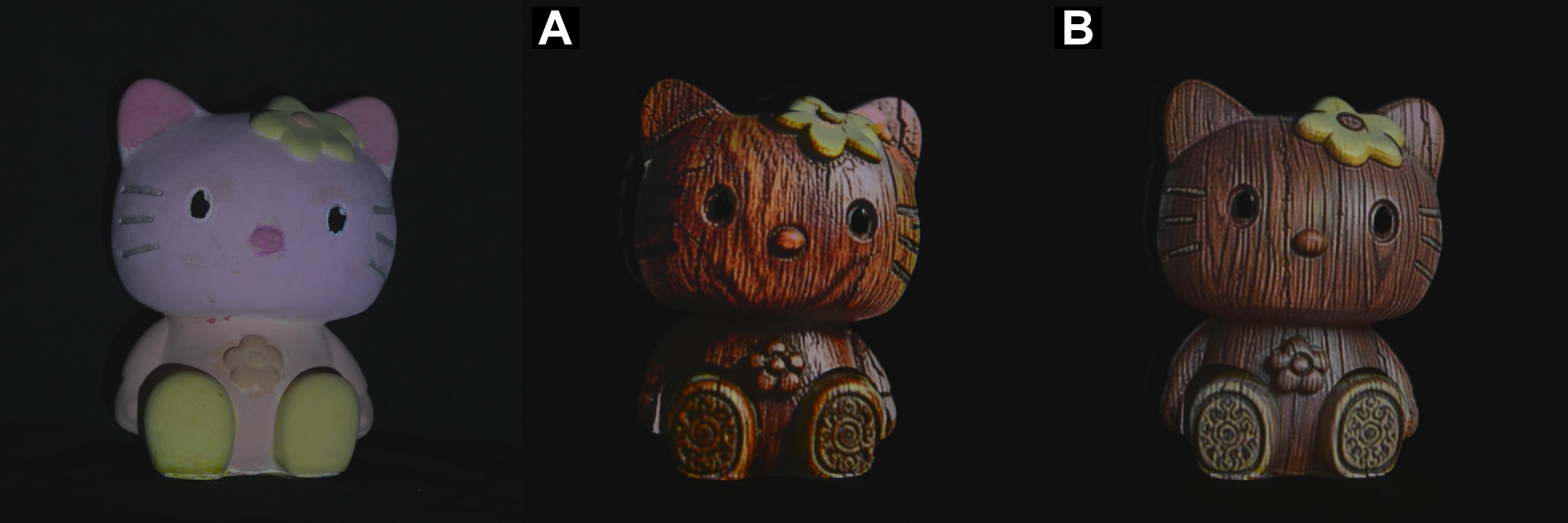}
\sreditcaptionof{figure}{Trial 25 (LongCat): ``Transform the toy’s entire surface into aged carved wood, with rich dark wood grain, visible hand-carved patterns, worn edges, subtle cracks, and a handcrafted sculptural finish.''}
\end{center}\vspace{2pt}

\begin{center}
\includegraphics[width=\columnwidth]{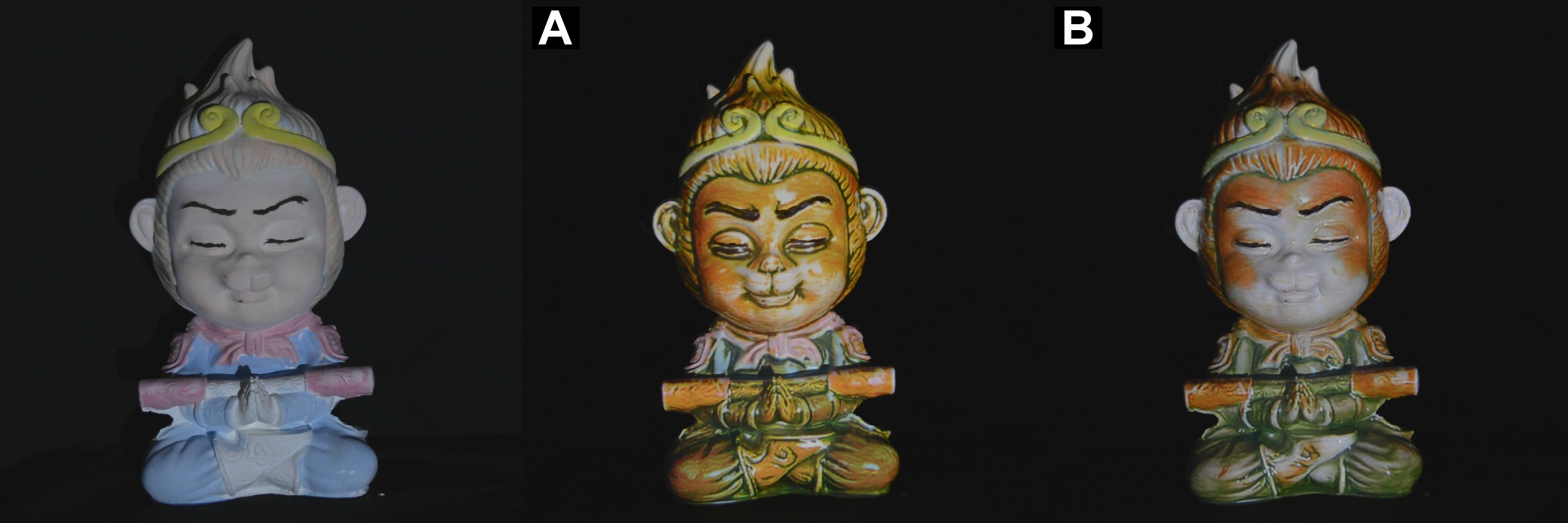}
\sreditcaptionof{figure}{Trial 26 (Qwen): ``Give the toy a Tang-sancai ceramic-figurine-style doll appearance, with amber-green-white glaze colors and flowing ceramic textures, while preserving the original shape and silhouette.''}
\end{center}\vspace{2pt}

\begin{center}
\includegraphics[width=\columnwidth]{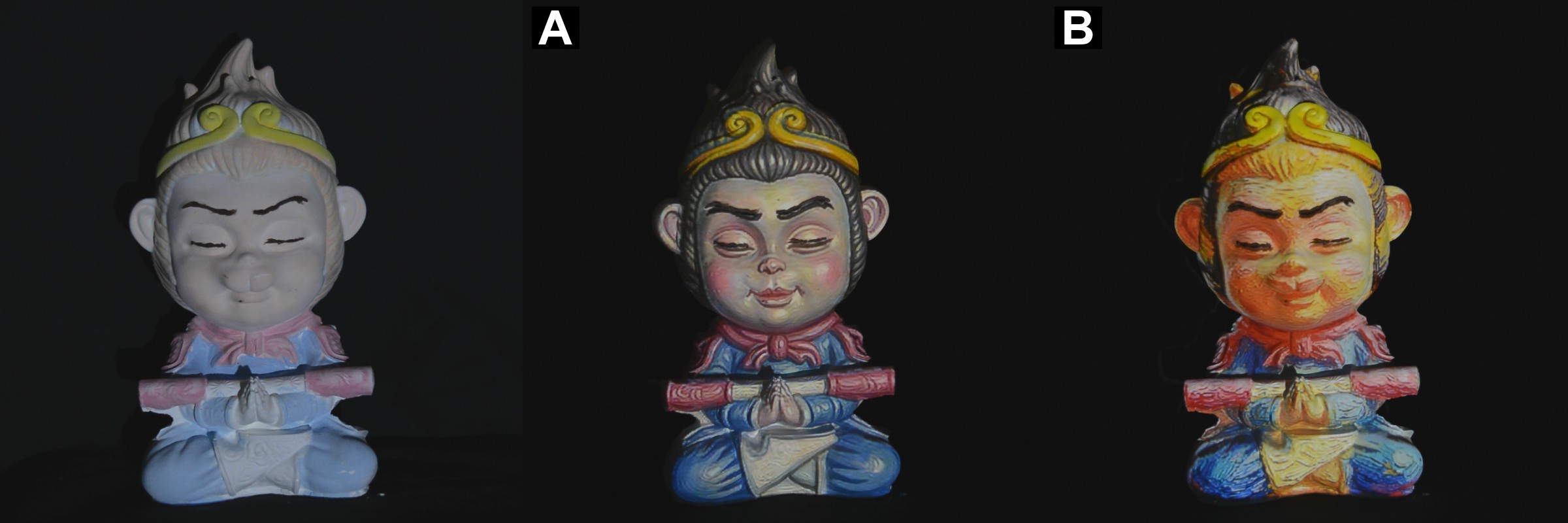}
\sreditcaptionof{figure}{Trial 27 (LongCat): ``Give the toy an oil-painted-statue-style doll appearance, with rich layered colors and visible painterly brush textures, while preserving the original shape and silhouette.''}
\end{center}\vspace{2pt}

\begin{center}
\includegraphics[width=\columnwidth]{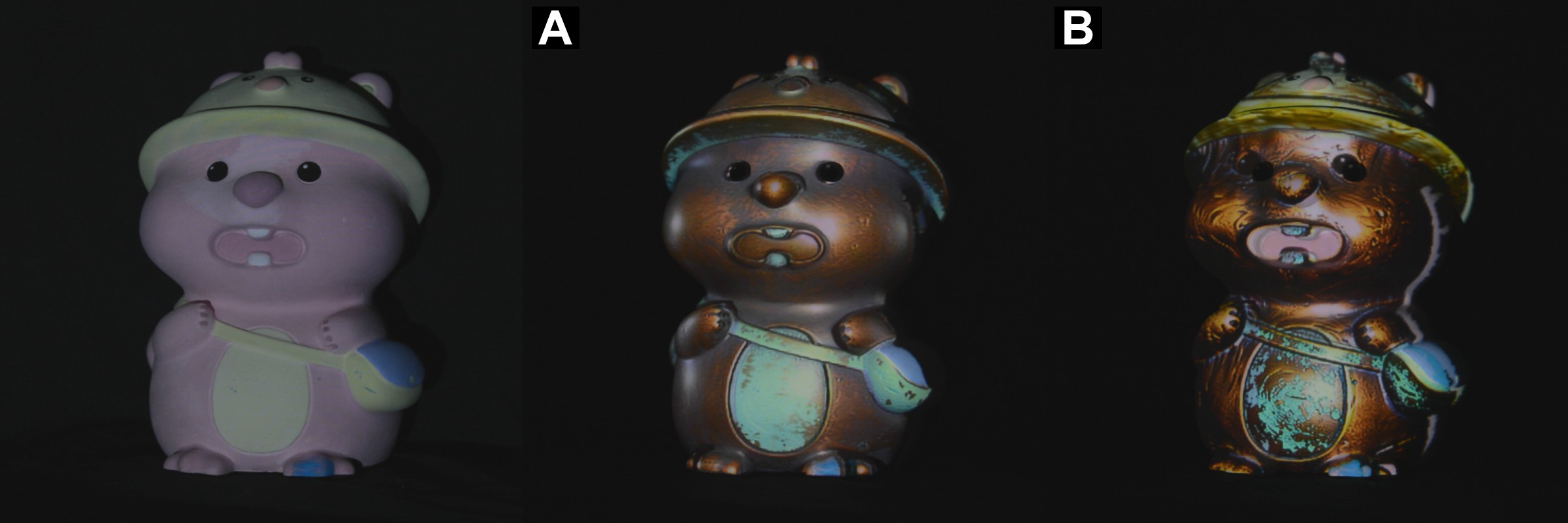}
\sreditcaptionof{figure}{Trial 28 (LongCat): ``apply an aged bronze finish with subtle turquoise patina to the toy's existing surface''}
\end{center}\vspace{2pt}

\begin{center}
\includegraphics[width=\columnwidth]{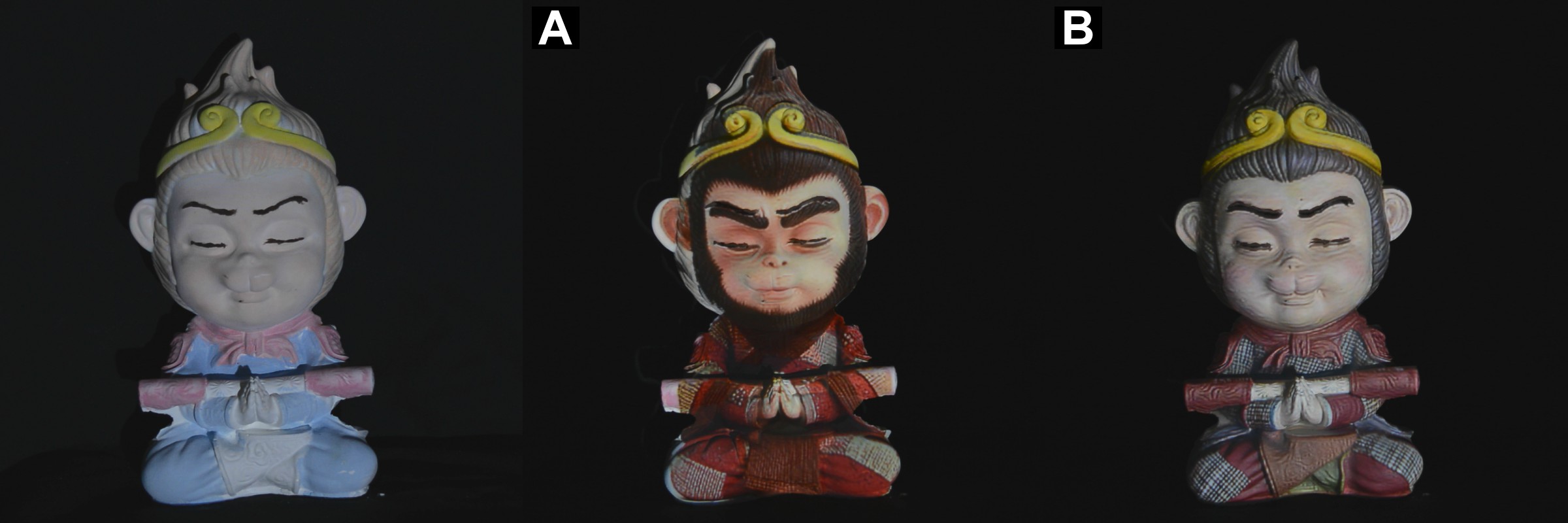}
\sreditcaptionof{figure}{Trial 29 (LongCat): ``Give the toy a Ji-Gong-inspired doll appearance, with rustic brown-red tones and patchwork robe patterns, while preserving the original shape and silhouette.''}
\end{center}\vspace{2pt}

\begin{center}
\includegraphics[width=\columnwidth]{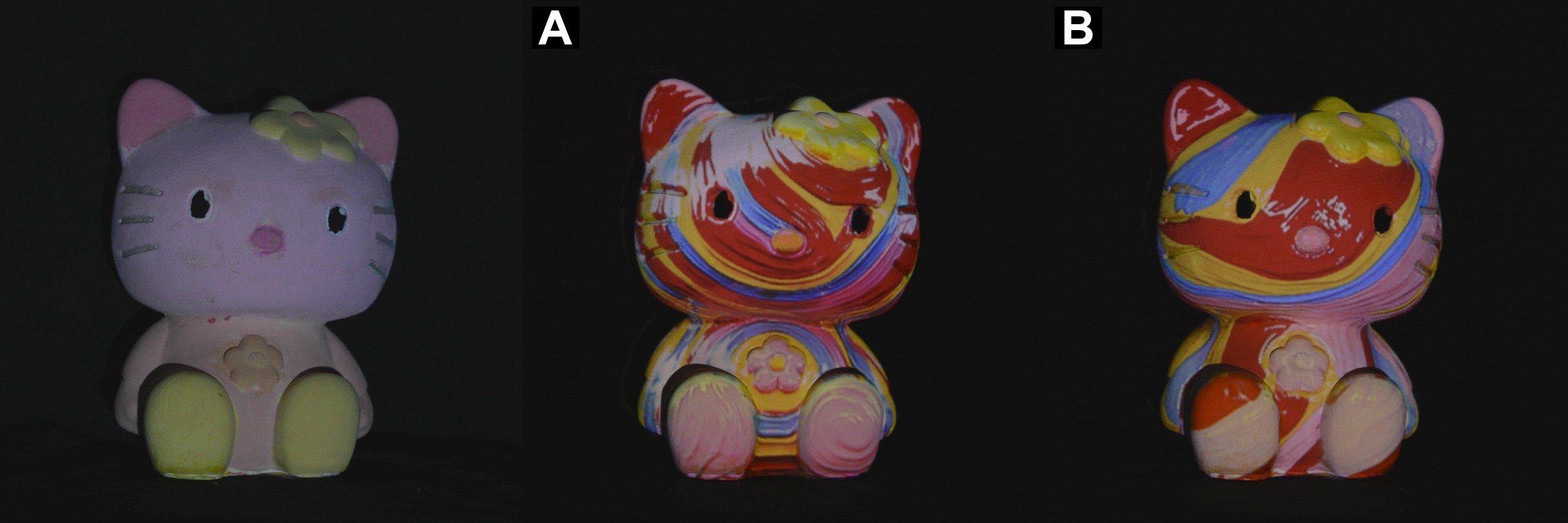}
\sreditcaptionof{figure}{Trial 30 (Qwen): ``Transform the toy’s entire surface into layered colored wax, with broad flowing bands of red, yellow, blue, and pink, glossy hardened textures, visible wax ridges, and rich reflective shading.''}
\end{center}\vspace{2pt}

\begin{center}
\includegraphics[width=\columnwidth]{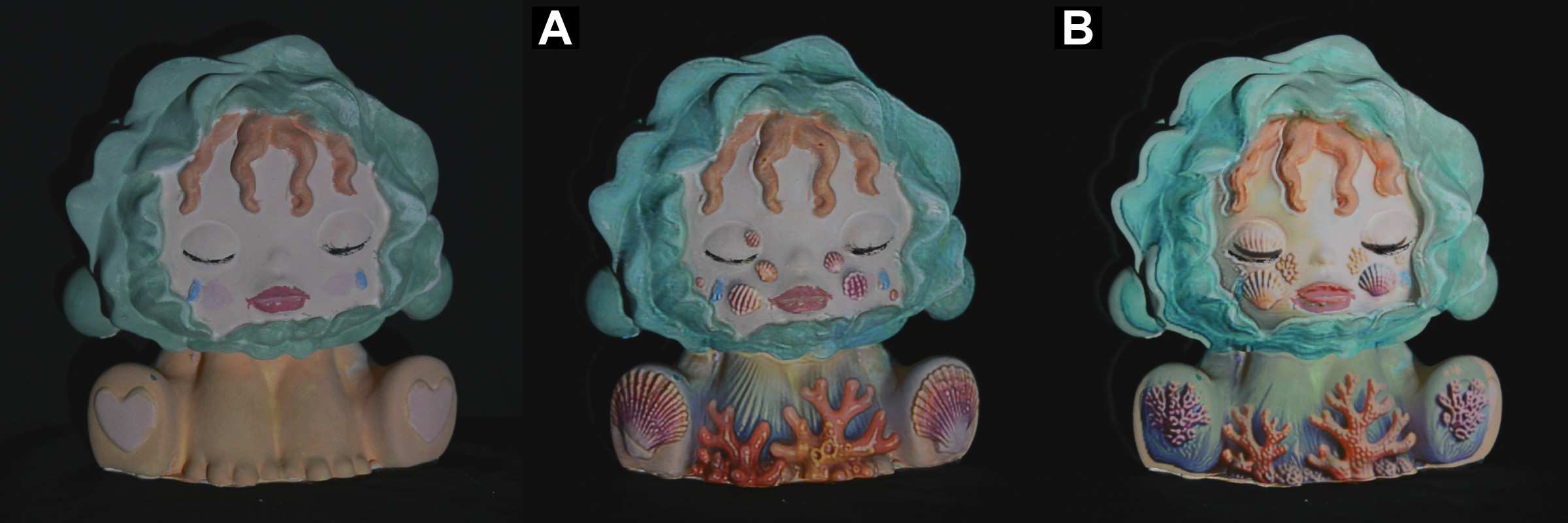}
\sreditcaptionof{figure}{Trial 31 (LongCat): ``Give the toy a coral-reef ocean-style doll appearance, with vivid ocean colors and organic coral patterns, including subtle seashell textures on the face, while preserving the original shape and silhouette.''}
\end{center}\vspace{2pt}

\begin{center}
\includegraphics[width=\columnwidth]{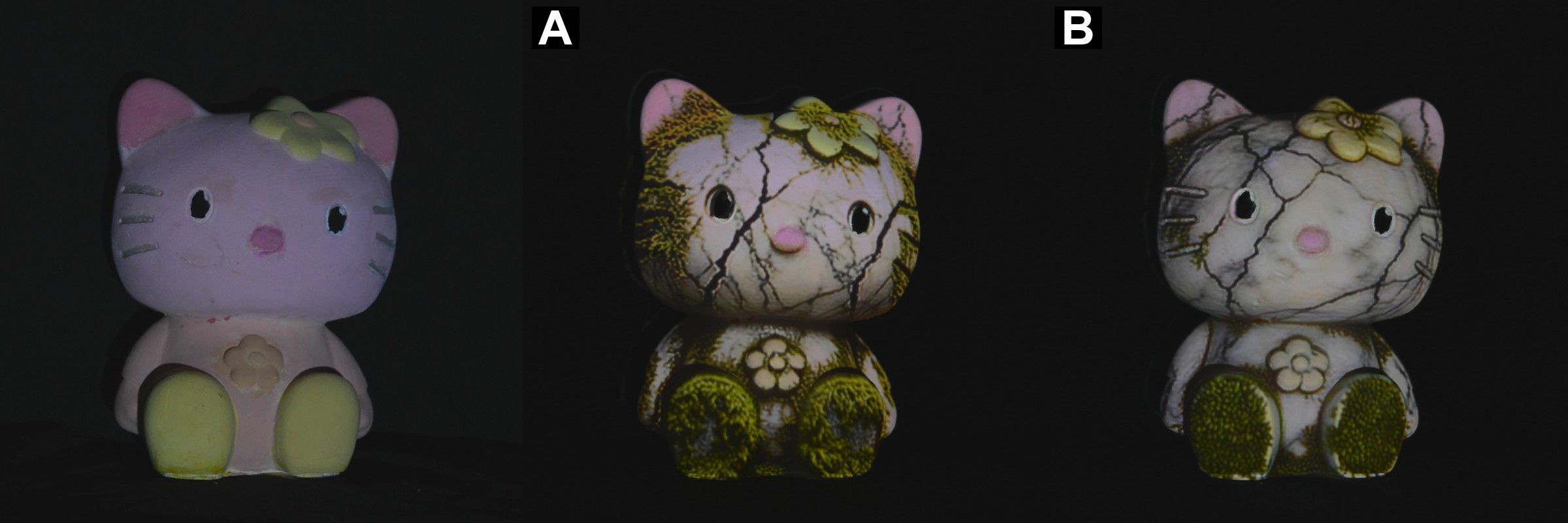}
\sreditcaptionof{figure}{Trial 32 (LongCat): ``Transform the toy’s entire surface into aged marble overgrown with moss, with visible stone veining, weathered cracks, damp stains, soft green moss patches, and a cold sculptural finish.''}
\end{center}\vspace{2pt}

\begin{center}
\includegraphics[width=\columnwidth]{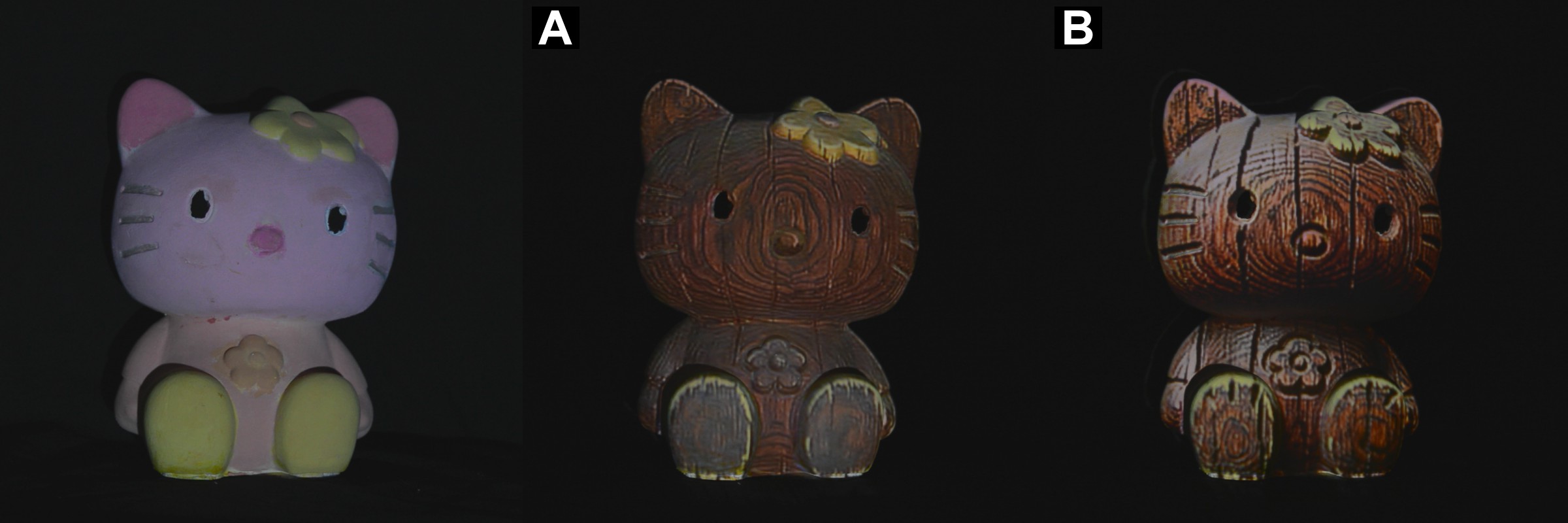}
\sreditcaptionof{figure}{Trial 33 (Qwen): ``Transform the toy’s entire surface into aged carved wood, with rich dark wood grain, visible hand-carved patterns, worn edges, subtle cracks, and a handcrafted sculptural finish.''}
\end{center}\vspace{2pt}

\begin{center}
\includegraphics[width=\columnwidth]{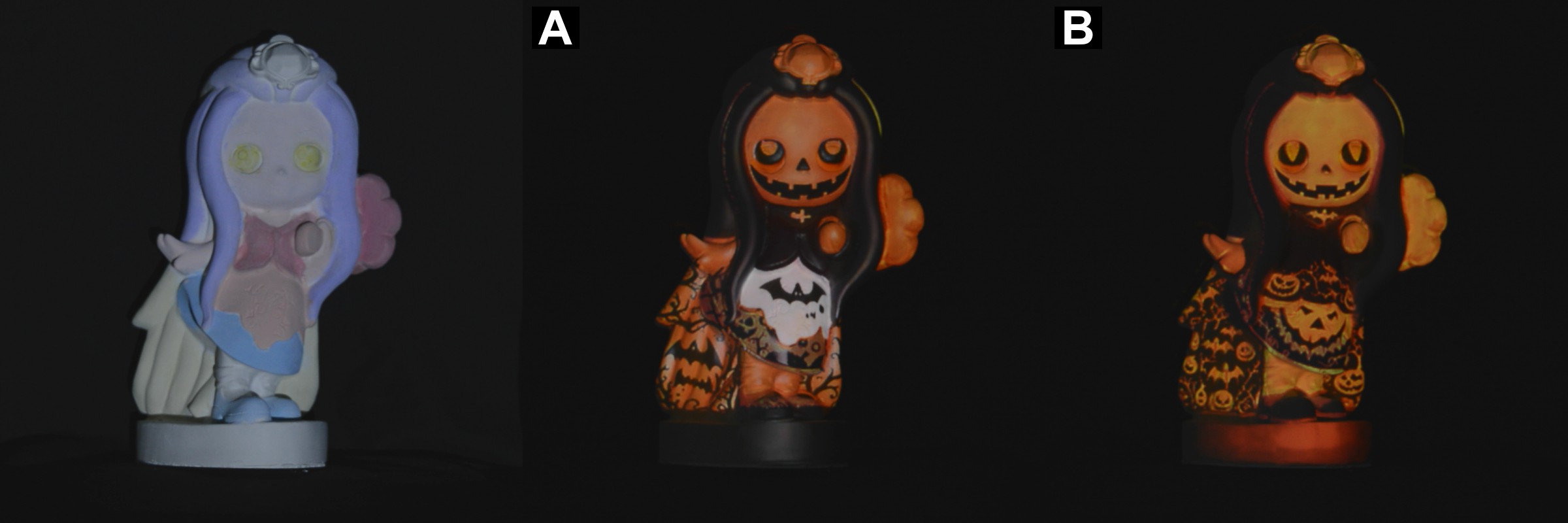}
\sreditcaptionof{figure}{Trial 34 (Qwen): ``Give the toy a Halloween outfit-style doll appearance, with orange-black tones and spooky decorative patterns, while preserving the original shape and silhouette.''}
\end{center}\vspace{2pt}

\begin{center}
\includegraphics[width=\columnwidth]{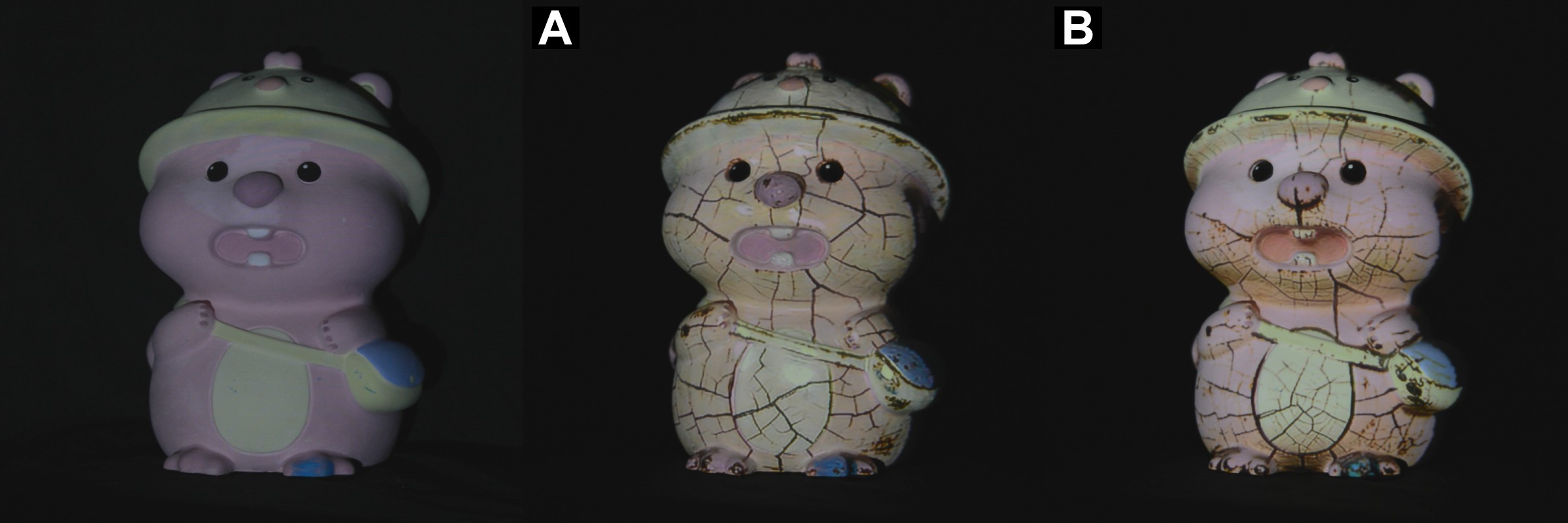}
\sreditcaptionof{figure}{Trial 35 (Qwen): ``Transform the toy’s entire surface into aged cracked ceramic, with an ivory-white glaze, clearly visible dark crackle lines, worn stains, chipped paint textures, and glossy ceramic highlights.''}
\end{center}\vspace{2pt}

\begin{center}
\includegraphics[width=\columnwidth]{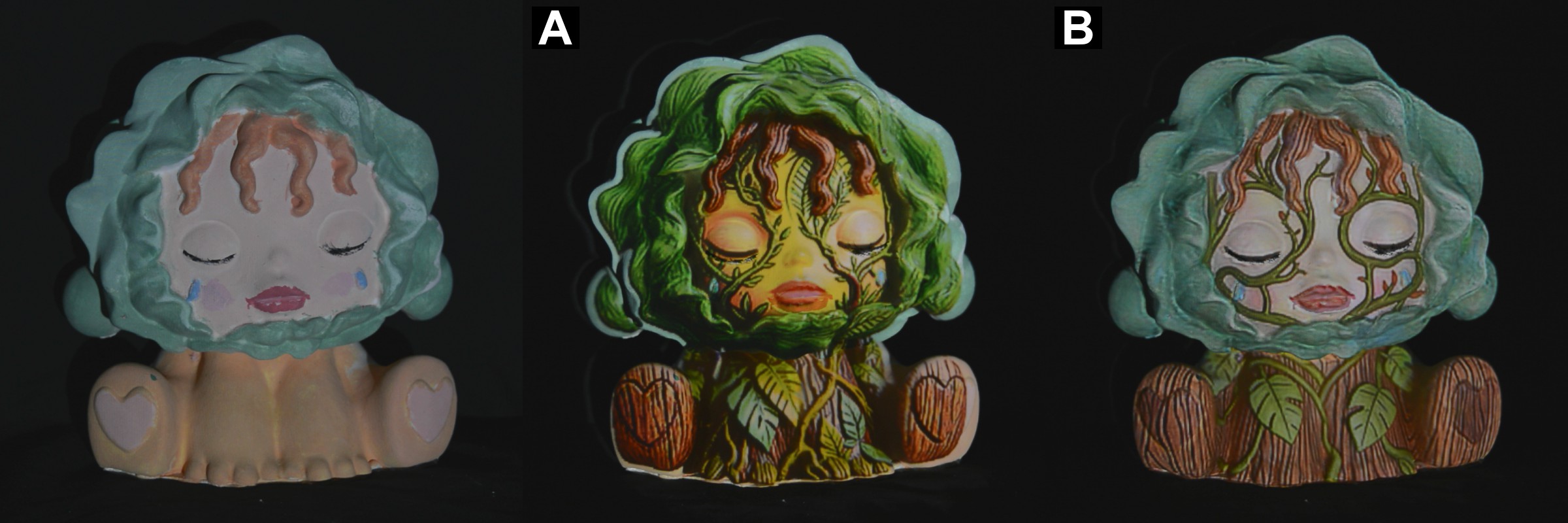}
\sreditcaptionof{figure}{Trial 36 (LongCat): ``Give the toy a jungle-style doll appearance, with lush green tones and organic jungle patterns, including vine and branch textures on the face, and wood, leaf, and vine textures across the surface, while preserving the original shape and silhouette.''}
\end{center}\vspace{2pt}

\clearpage

\end{document}